\documentclass[11pt]{wlscirep}
\usepackage[utf8]{inputenc}
\usepackage[T1]{fontenc}
\usepackage{refcount}
\usepackage{booktabs}
\usepackage[normalem]{ulem}
\useunder{\uline}{\ul}{}
\usepackage{lscape}
\usepackage{longtable}
\usepackage{multirow}
\usepackage{cite}
\usepackage{tablefootnote}
\usepackage[symbol]{footmisc}
\usepackage{setspace}
\usepackage{url}
\usepackage{amsmath}
\usepackage{adjustbox}
\usepackage{array}
\usepackage{tabularx}
\usepackage{longtable}
\usepackage{colortbl}
\usepackage{booktabs}
\usepackage{ulem}
\usepackage{graphicx}
\usepackage{appendix}
\usepackage{subcaption}
\usepackage{hyperref}
\usepackage{nameref}
\usepackage[most]{tcolorbox}
\usepackage{array}
\usepackage{hyperref}
\usepackage{microtype}
\usepackage{jabbrv}

\title{Retrieval augmented generation based dynamic prompting for few-shot biomedical named entity recognition using large language models}
\vspace{-1cm} 
\author[1,2,3,*]{Yao Ge}
\author[1]{Sudeshna Das}
\author[1,2]{Yuting Guo}
\author[1,4]{Abeed Sarker}

\affil[1]{Department of Biomedical Informatics, School of Medicine, Emory University, Atlanta, GA, USA}
\affil[2]{Department of Computer Science, Emory University, Atlanta, GA, USA}
\affil[3]{National Library of Medicine, National Institutes of Health, Bethesda, MD}
\affil[4]{Department of Biomedical Engineering, Georgia Institute of Technology and Emory University, Atlanta, GA, USA}

\affil[*]{Corresponding author: Yao Ge, Ph.D, Department of Biomedical Informatics, School of Medicine, Emory University, Atlanta, GA, USA, \tt yao.ge@emory.edu}

\enlargethispage{\baselineskip}
\begin{abstract}
\vspace{-0.6cm}

Biomedical named entity recognition (NER) is a high-utility natural language processing (NLP) task, and large language models (LLMs) show promise particularly in few-shot settings (i.e., limited training data). In this article, we address the performance challenges of LLMs for few-shot biomedical NER by investigating a dynamic prompting strategy involving retrieval-augmented generation (RAG). In our approach, the annotated in-context learning examples are selected based on their similarities with the input texts, and the prompt is dynamically updated for each instance during inference. We implemented and optimized static and dynamic prompt engineering techniques and evaluated them on five biomedical NER datasets. Static prompting with structured components increased average F$_1$-scores by 12\% for GPT-4, and 11\% for GPT-3.5 and LLaMA 3-70B, relative to basic static prompting. Dynamic prompting further improved performance, with TF-IDF and SBERT retrieval methods yielding the best results, improving average F$_1$-scores by 7.3\% and 5.6\% in 5-shot and 10-shot settings, respectively. These findings highlight the utility of contextually adaptive prompts via RAG for biomedical NER.

\end{abstract}

\newpage

\begin{document}
\flushbottom
\maketitle
%
%
\thispagestyle{empty}
\newpage
\section*{Introduction}

Named entity recognition (NER) is a fundamental natural language processing (NLP) task, involving the extraction of predefined entities from free text, with a wide range of application in digital medicine. Within the broader biomedical domain, NER methods can be applied to detect entities such as diseases, adverse drug reactions, treatments, and symptoms. Biomedical NER often presents unique challenges due to the sparsity of certain medical concepts (\textit{e.g.}, rare health conditions), and the specialized language used in clinical and social health contexts. While some deep neural network based NER methods have achieved human-like performance, they require large amounts of training data, which is often not available for specialized biomedical problems. Furthermore, NER systems trained on open domain (as opposed to restricted domain) data are typically not transferrable to biomedical domain-specific tasks without additional training or manual annotation. Manually annotating data for targeted biomedical NLP problems is expensive (\textit{e.g.}, when they involve medical experts) or infeasible (\textit{e.g.}, for sparsely-occurring medical conditions). Even when studies invest time in annotating biomedical data, the datasets cannot be publicly shared (\textit{e.g.}, electronic health record data for privacy compliance). Thus, there is a need to develop few-shot learning (FSL) based NER solutions that can generalize effectively from a small number of examples~\cite{ge2023few}. In this paper, we describe the development and evaluation of an NER approach that leverages a large language model (LLM) and retrieval augmented generation (RAG) for improving the state of the art in few-shot biomedical NER.  

The emergence of generative LLMs such as the GPT and LLaMA series has enabled the research community to achieve large performance gains in FSL-based NLP methods. Generic LLMs are adaptable to biomedical domain NLP tasks in zero-shot and few-shot settings~\cite{labrak-etal-2024-zero}, and their ability to adapt to new tasks with minimal examples has been transformative, particularly for restricted domains such as biomedical. By leveraging in-context learning (\textit{i.e.}, training examples are provided as part of the prompt), current state-of-the-art LLMs can readily adapt to diverse text structures and vocabularies~\cite{brown2020language}. A major focus of recent biomedical NLP research has been prompt engineering methods~\cite{zaghir2024prompt,yeh-etal-2022-decorate}, as carefully designed prompts can help align an LLM's understanding and outputs with task-specific requirements.

The typical approach in prompt-driven methods is to optimize predefined,\textit{static} prompts for a target task. Static, in this context, refers to the use of the same, consistent prompt for every instance in a dataset during inference. Thus, regardless of the content of the input text, the model applies the same fixed prompt and in-context examples. This lack of flexibility to adjust to specific input data potentially leads to sub-optimal performance~\cite{ye2023prompt}. Their fixed format restricts performance potential, as they do not adjust based on context, even when more suitable annotated examples are present in the training data. Consequently, approaches employing static prompts exhibit high variance depending on the relevance of the in-context examples to the unlabeled input texts~\cite{chang-jia-2023-data}. 

More sophisticated architectures include approaches such as retrieval-augmented generation (RAG) and chain-of-thought (CoT) prompting~\cite{li2024rt}. CoT is a reasoning paradigm that enables models to generate intermediate reasoning steps, mimicking human-like problem-solving processes~\cite{wei2022chain}. It enhances performance on complex tasks by decomposing problems into sequential, logical steps, improving both accuracy and interpretability in multi-step reasoning scenarios. RAG involves retrieving query-relevant information, and enriching the context of the LLMs with retrieved contents prior to generation~\cite{lewis2020retrieval}. The retrieval process is typically guided by similarity measures~\cite{liao1998similarity}, such as cosine similarity between embeddings, which helps the model access contextually relevant examples or documents tailored to the input query. Once these relevant texts are retrieved, they are integrated into the prompt or used as additional context to aid the model’s response generation, resulting in a more contextually informed output.

The motivation behind RAG is to address the limitations of LLMs in handling tasks that require specialized or up-to-date information~\cite{gao2023retrieval}. Even with extensive training data, LLMs may struggle with domain-specific concepts or recent developments due to knowledge cutoffs and lack of domain specificity in training corpora~\cite{jin2024genegpt}. By introducing contextually relevant information at inference time, RAG can significantly improve performance in specialized applications~\cite{xiong2024benchmarking}, such as biomedical text analysis, where precision and relevance are critical. In biomedical NER, RAG can improve the adaptability of a model by retrieving examples or contexts that closely resemble the input text, thereby increasing the identification accuracy of entities~\cite{xiong2024benchmarking}. In FSL settings, RAG architectures have the potential to reduce reliance on large annotated datasets by dynamically selecting relevant data~\cite{jeong2024adaptive}, making it particularly useful for domains where annotated data is limited. Additionally, RAG complements prompting techniques, like CoT prompting, by enabling stepwise reasoning based on retrieved information~\cite{sahoo2024systematic}, which may lead to better precision and recall for complex, sparse entities.

With the aim of addressing the inherent limitations of static prompts and improving biomedical NER performance in FSL settings, we explore dynamic prompting techniques, which involve automatically retrieving suitable training examples and adjusting prompts based on contextual similarity. Following the optimization of prompts, we evaluate the effectiveness of the two types of prompting---static and dynamic---using three LLMs: GPT-3.5, GPT-4, and LLaMA 3 (open source), on five datasets. We investigate the effectiveness of both static and dynamic prompt engineering strategies, integrated with retrieval mechanisms, to improve few-shot NER. We systematically evaluate multiple retrieval mechanisms and investigate how prompt design choices affect model performance. Our results demonstrate the potential of these techniques to enhance entity recall and precision in biomedical NER, offering insights into the optimization of LLMs for biomedical applications.

The primary contributions of this work are as follows:
\begin{itemize}
\item Development of a structured static prompt optimization framework, incorporating task-relevant instructions, entity definitions, and dataset contextualization to improve few-shot biomedical NER.
\item Integration and evaluation of RAG, assessing how distinct retrieval mechanisms (TF-IDF, SBERT, ColBERT, and DPR) enhance dynamic prompting by selecting contextually relevant examples.
\item Comparative analysis of static and dynamic prompting strategies, benchmarking their effectiveness in few-shot biomedical NER, and offering insights into their strengths across different datasets.
\end{itemize}

\section*{Results}
\subsection*{Task-specific Static Prompting}

The results in Table \ref{tab:static_prompt} demonstrate consistent performance improvements across the five biomedical datasets when multiple components of the static prompting strategy are combined for all three LLMs. Compared to the baseline prompt, the addition of task-specific components, such as dataset descriptions, high-frequency instances, error analysis, and few-shot examples, led to significant improvements in precision, recall, and F$_1$-score across all datasets. GPT-4 showed the largest improvements when the full structured prompt was used. For GPT-4, the average F$_1$-score increased by 12.0\% across datasets, ranging from 6.95\% for MIMIC III to 23.7\% for Med-Mentions. GPT-3.5 obtained an average F$_1$-score increase of 11.4\%, with gains ranging from 7.1\% for BC5CDR to 22.9\% for Med-Mentions. LLaMA 3-70B, which started with the lowest baseline performance, showed an average F$_1$-score increase of 11.1\%, with its largest improvement also observed in the Med-Mentions dataset (21.44\%).

GPT-4 consistently outperformed GPT-3.5 and LLaMA 3-70B in all configurations, benefiting more from the integration of task-specific components, particularly in datasets such as BC5CDR and Med-Mentions, where it achieved the highest F$_1$-score. GPT-3.5 and LLaMA 3, while achieving slightly lower overall performance still exhibited performance improvements relative to the baseline. This is evident in datasets such as \textsc{Reddit-Impacts}, where its F$_1$-score exhibited significant improvement with the integration of additional components. 
\begin{table}[]
\centering
\caption[Performance comparison of various prompting strategies]{Performance comparison of various prompting strategies across different datasets in terms of F$_1$-score (F$_1$), Precision (P), and Recall (R). The row "BP + All components" represents the combination of all strategies, with the best performance across datasets highlighted in bold. The red bold font indicates the best F$_1$ score achieved by an individual component, while black bold font highlights the highest Precision, and underlined text denotes the best Recall for a single component. Additionally, green bold font is used to mark F$_1$ scores that are lower than the baseline performance (BP).}
\resizebox{\columnwidth}{!}{%
\begin{tabular}{lrrrrrrrrrrrrrrr}
\hline
 &
  \multicolumn{3}{c|}{\cellcolor[HTML]{D9EAD3}\textit{\textbf{Reddit\_Impacts}}} &
  \multicolumn{3}{c|}{\cellcolor[HTML]{D9EAD3}\textit{\textbf{BC5CDR}}} &
  \multicolumn{3}{c|}{\cellcolor[HTML]{D9EAD3}\textit{\textbf{MIMIC III}}} &
  \multicolumn{3}{c|}{\cellcolor[HTML]{D9EAD3}\textit{\textbf{NCBI}}} &
  \multicolumn{3}{c}{\cellcolor[HTML]{D9EAD3}\textit{\textbf{Med-Mentions}}} \\ \hline
 &
  \multicolumn{1}{l}{\textbf{P}} &
  \multicolumn{1}{l}{\textbf{R}} &
  \multicolumn{1}{l|}{\textbf{F$_1$}} &
  \multicolumn{1}{l}{\textbf{P}} &
  \multicolumn{1}{l}{\textbf{R}} &
  \multicolumn{1}{l|}{\textbf{F$_1$}} &
  \multicolumn{1}{l}{\textbf{P}} &
  \multicolumn{1}{l}{\textbf{R}} &
  \multicolumn{1}{l|}{\textbf{F$_1$}} &
  \multicolumn{1}{l}{\textbf{P}} &
  \multicolumn{1}{l}{\textbf{R}} &
  \multicolumn{1}{l|}{\textbf{F$_1$}} &
  \multicolumn{1}{l}{\textbf{P}} &
  \multicolumn{1}{l}{\textbf{R}} &
  \multicolumn{1}{l}{\textbf{F$_1$}} \\ \hline
\rowcolor[HTML]{E6F1FA} 
\multicolumn{16}{l}{\cellcolor[HTML]{E6F1FA}\textbf{GPT-3.5}} \\ \hline
\rowcolor[HTML]{FCEBEB} 
\cellcolor[HTML]{FCEBEB}\textbf{Basic Prompt (BP)} &
  10.37 &
  43.26 &
  \multicolumn{1}{l|}{16.73} &
  55.64 &
  76.88 &
  \multicolumn{1}{l|}{64.56} &
  55.31 &
  54.11 &
  \multicolumn{1}{l|}{54.70} &
  18.28 &
  51.33 &
  \multicolumn{1}{l|}{26.96} &
  8.55 &
  10.12 &
  9.27 \\ \hline
\textbf{BP + Description of datasets} &
  13.25 &
  52.38 &
  \multicolumn{1}{l|}{21.15} &
  59.25 &
  81.47 &
  \multicolumn{1}{l|}{68.61} &
  59.54 &
  54.18 &
  \multicolumn{1}{l|}{56.73} &
  26.24 &
  50.25 &
  \multicolumn{1}{l|}{34.48} &
  16.52 &
  10.33 &
  12.71 \\ \hline
\textbf{BP + High-frequency instances} &
  13.40 &
  50.13 &
  \multicolumn{1}{l|}{21.15} &
  59.08 &
  82.96 &
  \multicolumn{1}{l|}{69.01} &
  59.93 &
  55.66 &
  \multicolumn{1}{l|}{57.72} &
  26.54 &
  {\ul 55.68} &
  \multicolumn{1}{l|}{35.95} &
  20.16 &
  15.03 &
  17.22 \\ \hline
\textbf{BP + UMLS knowledge} &
  {10.17} &
  42.86 &
  \multicolumn{1}{l|}{{\color[HTML]{6AA84F} \textbf{16.44}}} &
  54.24 &
  80.57 &
  \multicolumn{1}{l|}{{64.83}} &
  {47.16} &
  54.50 &
  \multicolumn{1}{l|}{{\color[HTML]{6AA84F} \textbf{50.57}}} &
  23.93 &
  43.02 &
  \multicolumn{1}{l|}{{30.75}} &
  {\color[HTML]{000000} 10.26} &
  11.58 &
  {10.88} \\ \hline
\textbf{BP + Error analysis} &
  12.02 &
  48.21 &
  \multicolumn{1}{l|}{19.24} &
  57.36 &
  82.49 &
  \multicolumn{1}{l|}{67.67} &
  64.63 &
  55.16 &
  \multicolumn{1}{l|}{59.52} &
  25.23 &
  48.31 &
  \multicolumn{1}{l|}{33.15} &
  {18.74} &
  13.24 &
  {15.52} \\ \hline
\textbf{BP + 5-shot learning with sentences} &
  12.33 &
  44.42 &
  \multicolumn{1}{l|}{19.30} &
  59.30 &
  82.04 &
  \multicolumn{1}{l|}{68.84} &
  53.09 &
  53.37 &
  \multicolumn{1}{l|}{57.03} &
  37.09 &
  43.78 &
  \multicolumn{1}{l|}{40.16} &
  17.28 &
  25.54 &
  20.61 \\ \hline
\textbf{BP + 5-shot learning with tokens} &
  {\textbf{13.63}} &
  {\ul 53.11} &
  \multicolumn{1}{l|}{{\color[HTML]{FE0000} \textbf{21.69}}} &
  {\textbf{62.27}} &
  {\ul 82.02} &
  \multicolumn{1}{l|}{{\color[HTML]{FE0000} \textbf{70.79}}} &
  {\textbf{67.50}} &
  {\ul 55.97} &
  \multicolumn{1}{l|}{{\color[HTML]{FE0000} \textbf{61.21}}} &
  \textbf{40.15} &
  46.32 &
  \multicolumn{1}{l|}{{\color[HTML]{FE0000} \textbf{43.01}}} &
  {\textbf{20.54}} &
  {\ul 30.57} &
  {\color[HTML]{FE0000} \textbf{24.57}} \\ \hline
\textbf{BP + All above} &
  \cellcolor[HTML]{FCE5CD}\textbf{15.36} &
  \cellcolor[HTML]{FCE5CD}\textbf{53.92} &
  \multicolumn{1}{l|}{\cellcolor[HTML]{FCE5CD}\textbf{23.91}} &
  \cellcolor[HTML]{FCE5CD}\textbf{63.64} &
  \cellcolor[HTML]{FCE5CD}\textbf{84.86} &
  \multicolumn{1}{l|}{\cellcolor[HTML]{FCE5CD}\textbf{72.73}} &
  \cellcolor[HTML]{FCE5CD}\textbf{67.77} &
  \cellcolor[HTML]{FCE5CD}\textbf{57.10} &
  \multicolumn{1}{l|}{\cellcolor[HTML]{FCE5CD}\textbf{61.99}} &
  \cellcolor[HTML]{FCE5CD}\textbf{42.73} &
  \cellcolor[HTML]{FCE5CD}\textbf{48.07} &
  \multicolumn{1}{l|}{\cellcolor[HTML]{FCE5CD}\textbf{45.24}} &
  \cellcolor[HTML]{FCE5CD}\textbf{22.15} &
  \cellcolor[HTML]{FCE5CD}\textbf{55.32} &
  \cellcolor[HTML]{FCE5CD}\textbf{31.63} \\ \hline
\rowcolor[HTML]{E6F1FA} 
\multicolumn{16}{l}{\cellcolor[HTML]{E6F1FA}\textbf{GPT-4}} \\ \hline
\rowcolor[HTML]{FCEBEB} 
\textbf{Basic Prompt (BP)} &
  12.75 &
  48.15 &
  \multicolumn{1}{l|}{20.16} &
  59.56 &
  83.22 &
  \multicolumn{1}{l|}{69.43} &
  57.57 &
  55.72 &
  \multicolumn{1}{l|}{56.63} &
  25.13 &
  50.48 &
  \multicolumn{1}{l|}{33.56} &
  18.27 &
  11.12 &
  13.83 \\ \hline
\textbf{BP + Description of datasets} &
  15.12 &
  52.94 &
  \multicolumn{1}{l|}{23.52} &
  60.66 &
  84.58 &
  \multicolumn{1}{l|}{70.65} &
  63.35 &
  56.42 &
  \multicolumn{1}{l|}{59.68} &
  26.43 &
  {\ul 55.22} &
  \multicolumn{1}{l|}{35.75} &
  21.23 &
  11.96 &
  15.30 \\ \hline
\textbf{BP + High-frequency instances} &
  15.98 &
  {\ul 53.75} &
  \multicolumn{1}{l|}{24.64} &
  63.89 &
  84.06 &
  \multicolumn{1}{l|}{72.60} &
  \textbf{64.61} &
  56.14 &
  \multicolumn{1}{l|}{60.08} &
  35.02 &
  41.44 &
  \multicolumn{1}{l|}{37.96} &
  21.72 &
  17.69 &
  19.50 \\ \hline
\textbf{BP + UMLS knowledge} &
  12.85 &
  50.14 &
  \multicolumn{1}{l|}{20.46} &
  59.48 &
  84.63 &
  \multicolumn{1}{l|}{69.86} &
  {55.37} &
  54.90 &
  \multicolumn{1}{l|}{{\color[HTML]{6AA84F} \textbf{55.13}}} &
  {22.80} &
  47.92 &
  \multicolumn{1}{l|}{{\color[HTML]{6AA84F} \textbf{30.90}}} &
  18.72 &
  11.83 &
  {14.50} \\ \hline
\textbf{BP + Error analysis} &
  14.87 &
  52.04 &
  \multicolumn{1}{l|}{23.13} &
  {67.92} &
  82.75 &
  \multicolumn{1}{l|}{{74.61}} &
  63.93 &
  56.72 &
  \multicolumn{1}{l|}{{60.11}} &
  34.86 &
  41.38 &
  \multicolumn{1}{l|}{{37.84}} &
  20.28 &
  16.59 &
  {18.25} \\ \hline
\textbf{BP + 5-shot learning with sentences} &
  {14.71} &
  51.48 &
  \multicolumn{1}{l|}{{22.88}} &
  {65.04} &
  83.18 &
  \multicolumn{1}{l|}{{73.00}} &
  58.49 &
  55.03 &
  \multicolumn{1}{l|}{{58.25}} &
  36.96 &
  45.67 &
  \multicolumn{1}{l|}{{40.86}} &
  27.42 &
  30.33 &
  {28.80} \\ \hline
\textbf{BP + 5-shot learning with tokens} &
  {\textbf{17.23}} &
  52.57 &
  \multicolumn{1}{l|}{{\color[HTML]{FE0000} \textbf{25.95}}} &
  {\textbf{68.10}} &
  {\ul 87.66} &
  \multicolumn{1}{l|}{{\color[HTML]{FE0000} \textbf{76.65}}} &
  {63.40} &
  {\ul 62.49} &
  \multicolumn{1}{l|}{{\color[HTML]{FE0000} \textbf{62.94}}} &
  {\textbf{40.72}} &
  48.42 &
  \multicolumn{1}{l|}{{\color[HTML]{FE0000} \textbf{44.24}}} &
  {\textbf{27.71}} &
  {\ul 41.41} &
  {\color[HTML]{FE0000} \textbf{33.20}} \\ \hline
\textbf{BP + All above} &
  \cellcolor[HTML]{FCE5CD}\textbf{18.87} &
  \cellcolor[HTML]{FCE5CD}\textbf{52.01} &
  \multicolumn{1}{l|}{\cellcolor[HTML]{FCE5CD}\textbf{27.60}} &
  \cellcolor[HTML]{FCE5CD}\textbf{68.62} &
  \cellcolor[HTML]{FCE5CD}\textbf{90.32} &
  \multicolumn{1}{l|}{\cellcolor[HTML]{FCE5CD}\textbf{78.03}} &
  \cellcolor[HTML]{FCE5CD}63.06 &
  \cellcolor[HTML]{FCE5CD}\textbf{64.12} &
  \multicolumn{1}{l|}{\cellcolor[HTML]{FCE5CD}\textbf{63.58}} &
  \cellcolor[HTML]{FCE5CD}\textbf{45.02} &
  \cellcolor[HTML]{FCE5CD}49.02 &
  \multicolumn{1}{l|}{\cellcolor[HTML]{FCE5CD}\textbf{46.93}} &
  \cellcolor[HTML]{FCE5CD}27.26 &
  \cellcolor[HTML]{FCE5CD}\textbf{60.06} &
  \cellcolor[HTML]{FCE5CD}\textbf{37.49} \\ \hline
\rowcolor[HTML]{E6F1FA} 
\multicolumn{16}{l}{\cellcolor[HTML]{E6F1FA}\textbf{Llama3-70B}} \\ \hline
\rowcolor[HTML]{FCEBEB} 
\textbf{Basic Prompt (BP)} &
  9.93 &
  36.42 &
  \multicolumn{1}{l|}{15.61} &
  52.52 &
  76.04 &
  \multicolumn{1}{l|}{62.13} &
  46.57 &
  55.64 &
  \multicolumn{1}{l|}{50.70} &
  15.63 &
  24.37 &
  \multicolumn{1}{l|}{19.15} &
  19.59 &
  23.17 &
  21.23 \\ \hline
\textbf{BP + Description of datasets} &
  13.27 &
  35.26 &
  \multicolumn{1}{l|}{19.28} &
  58.53 &
  80.22 &
  \multicolumn{1}{l|}{67.68} &
  56.64 &
  55.81 &
  \multicolumn{1}{l|}{56.22} &
  17.30 &
  36.43 &
  \multicolumn{1}{l|}{21.44} &
  23.59 &
  19.87 &
  21.57 \\ \hline
\textbf{BP + High-frequency instances} &
  {\textbf{14.52}} &
  34.53 &
  \multicolumn{1}{l|}{\color[HTML]{FE0000} \textbf{20.44}} &
  60.85 &
  78.05 &
  \multicolumn{1}{l|}{{68.39}} &
  55.65 &
  56.47 &
  \multicolumn{1}{l|}{{56.06}} &
  21.11 &
  42.97 &
  \multicolumn{1}{l|}{{26.62}} &
  \textbf{23.99} &
  31.20 &
  {27.12} \\ \hline
\textbf{BP + UMLS knowledge} &
  {7.87} &
  35.88 &
  \multicolumn{1}{l|}{{\color[HTML]{6AA84F} \textbf{12.91}}} &
  57.11 &
  74.63 &
  \multicolumn{1}{l|}{64.71} &
  {46.78} &
  51.63 &
  \multicolumn{1}{l|}{{\color[HTML]{6AA84F} \textbf{48.92}}} &
  14.95 &
  34.72 &
  \multicolumn{1}{l|}{20.91} &
  22.09 &
  25.51 &
  23.68 \\ \hline
\textbf{BP + Error analysis} &
  12.46 &
  38.86 &
  \multicolumn{1}{l|}{18.87} &
  58.96 &
  {\ul 80.52} &
  \multicolumn{1}{l|}{68.07} &
  63.18 &
  55.20 &
  \multicolumn{1}{l|}{58.92} &
  16.84 &
  44.65 &
  \multicolumn{1}{l|}{24.46} &
  22.64 &
  29.92 &
  25.78 \\ \hline
\textbf{BP + 5-shot learning with sentences} &
  11.35 &
  39.71 &
  \multicolumn{1}{l|}{17.65} &
  65.16 &
  77.28 &
  \multicolumn{1}{l|}{70.70} &
  62.90 &
  51.61 &
  \multicolumn{1}{l|}{56.85} &
  21.97 &
  {\ul 49.94} &
  \multicolumn{1}{l|}{30.52} &
  23.87 &
  64.66 &
  34.87 \\ \hline
\textbf{BP + 5-shot learning with tokens} &
  13.33 &
  {\ul 40.32} &
  \multicolumn{1}{l|}{20.04} &
  {\textbf{66.03}} &
  78.57 &
  \multicolumn{1}{l|}{{\color[HTML]{FE0000} \textbf{71.76}}} &
  \textbf{63.89} &
  {\ul 60.19} &
  \multicolumn{1}{l|}{{\color[HTML]{FE0000} \textbf{61.98}}} &
  {\textbf{34.49}} &
  32.41 &
  \multicolumn{1}{l|}{{\color[HTML]{FE0000} \textbf{33.42}}} &
  {23.72} &
  {\ul 68.45} &
  {\color[HTML]{FE0000} \textbf{35.23}} \\ \hline
\textbf{BP + All above} &
  \cellcolor[HTML]{FCE5CD}13.16 &
  \cellcolor[HTML]{FCE5CD}\textbf{57.86} &
  \multicolumn{1}{l|}{\cellcolor[HTML]{FCE5CD}\textbf{21.43}} &
  \cellcolor[HTML]{FCE5CD}\textbf{68.97} &
  \cellcolor[HTML]{FCE5CD}78.36 &
  \multicolumn{1}{l|}{\cellcolor[HTML]{FCE5CD}\textbf{73.32}} &
  \cellcolor[HTML]{FCE5CD}59.30 &
  \cellcolor[HTML]{FCE5CD}\textbf{67.27} &
  \multicolumn{1}{l|}{\cellcolor[HTML]{FCE5CD}\textbf{62.94}} &
  \cellcolor[HTML]{FCE5CD}\textbf{35.81} &
  \cellcolor[HTML]{FCE5CD}\textbf{34.71} &
  \multicolumn{1}{l|}{\cellcolor[HTML]{FCE5CD}\textbf{34.80}} &
  \cellcolor[HTML]{FCE5CD}{\textbf{25.89}} &
  \cellcolor[HTML]{FCE5CD}\textbf{67.05} &
  \cellcolor[HTML]{FCE5CD}{\textbf{37.26}} \\ \hline
\end{tabular}%
}
\label{tab:static_prompt}
\end{table}

As illustrated in Table \ref{tab:static_prompt}, high-frequency instances (see Section~\hyperref[sec:static-prompt]{Static Prompt Engineering}), and dataset descriptions had the most notable impact on recall. For example, in the Med-Mentions dataset, adding high-frequency instances improved recall for GPT-4 by 6.57\%. Few-shot examples at the token level provided the most significant increase in precision across models. For instance, precision in the NCBI dataset increased by 21.9\% for GPT-3.5 and by 15.6\% for GPT-4.

\begin{figure}[htbp]
    \centering
    \includegraphics[width=0.8\textwidth]{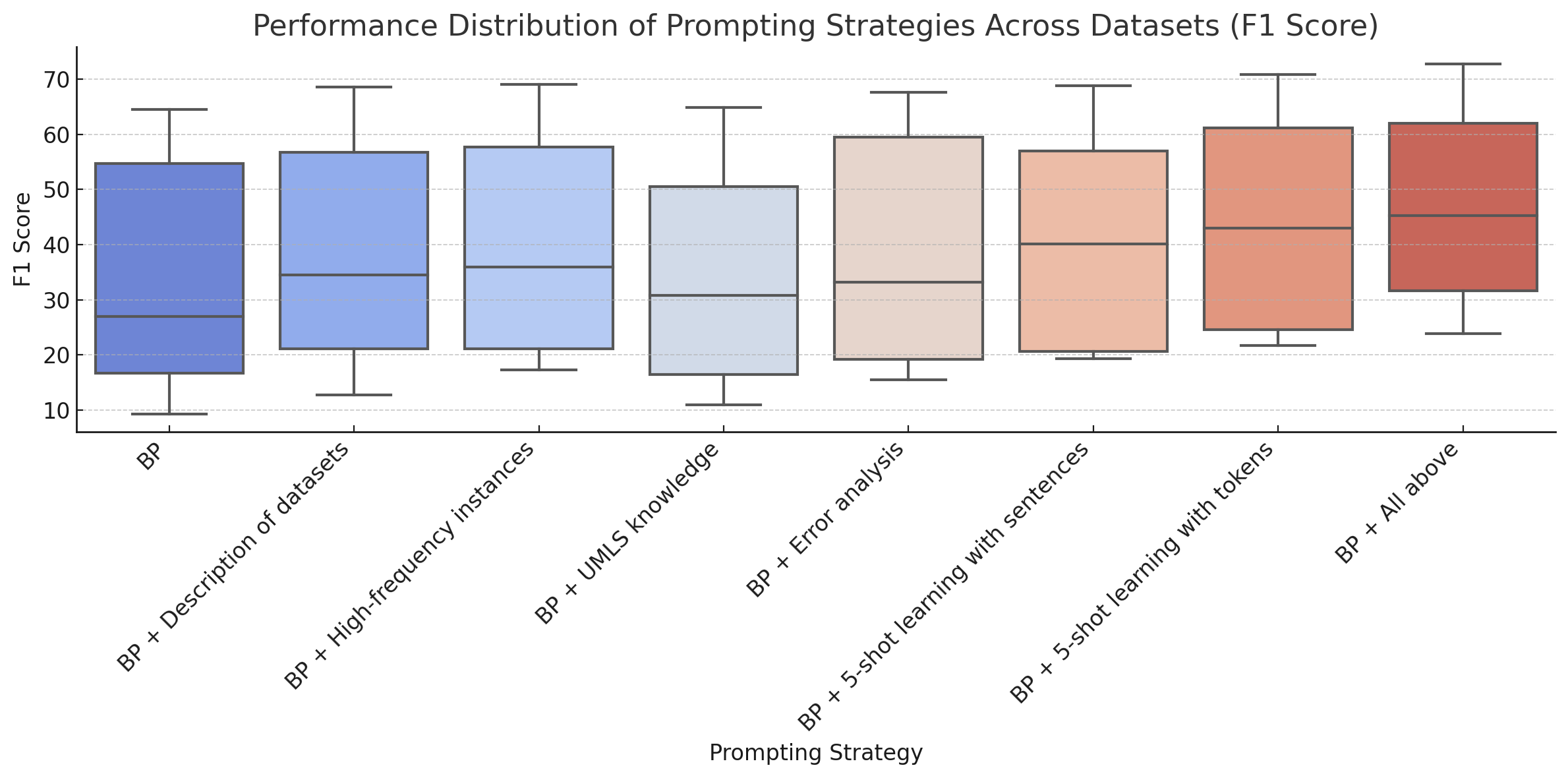}
    \caption[Performance Distribution of Prompting Strategies Across Datasets]{Performance distribution of prompting strategies across datasets (F$_1$-score). The box plots depict the performance of various prompting strategies applied to five biomedical datasets, highlighting the range, median, and distribution of F$_1$-scores for each strategy.}
    \label{fig:prompting_distribution}
\end{figure}

Figure \ref{fig:prompting_distribution} highlights the performance changes associated with different prompting strategies across datasets. The incorporation of knowledge from the unified medical language system (UMLS) improved recall in certain datasets such as BC5CDR but underperformed compared to the baseline prompt for datasets like \textsc{Reddit-Impacts} and NCBI. This component aimed to provide foundational biomedical knowledge by introducing descriptions and context derived from UMLS. However, this approach may have introduced noise, particularly in datasets that are not strongly aligned with UMLS's predefined biomedical concepts. For example, in the \textsc{Reddit-Impacts} dataset, GPT-3.5's F$_1$-score decreased slightly from 16.7 to 16.4, suggesting that the background information from UMLS diluted the model's ability to capture task-specific cues. 95\% confidence intervals (CIs) for each metric are provided in Table 6 in Supplementary Materials.

\subsection*{Dynamic Prompting with RAG}

\begin{table}[]
\centering
\caption[Performance of dynamic prompting strategies]{Evaluation of dynamic prompting strategies (5-shot, 10-shot, and 20-shot) using GPT-4 and Llama 3 across five biomedical datasets. The table presents F$_1$-score, precision, and recall for each retrieval method: Base Prompt, TF-IDF, SBERT, ColBERT, and DPR. The row "Base" represents using static prompts we proposed in the former section.}
\resizebox{\columnwidth}{!}{%
\begin{tabular}{llrrrrrrrrrrrrrrr}
\hline
 &
   &
  \multicolumn{3}{c|}{\cellcolor[HTML]{D9EAD3}\textit{\textbf{Reddit\_Impacts}}} &
  \multicolumn{3}{c|}{\cellcolor[HTML]{D9EAD3}\textit{\textbf{BC5CDR}}} &
  \multicolumn{3}{c|}{\cellcolor[HTML]{D9EAD3}\textit{\textbf{MIMIC III}}} &
  \multicolumn{3}{c|}{\cellcolor[HTML]{D9EAD3}\textit{\textbf{NCBI}}} &
  \multicolumn{3}{c}{\cellcolor[HTML]{D9EAD3}\textit{\textbf{Med-Mentions}}} \\ \hline
 &
   &
  \multicolumn{1}{l}{\textbf{P}} &
  \multicolumn{1}{l}{\textbf{R}} &
  \multicolumn{1}{l|}{\textbf{F$_1$}} &
  \multicolumn{1}{l}{\textbf{P}} &
  \multicolumn{1}{l}{\textbf{R}} &
  \multicolumn{1}{l|}{\textbf{F$_1$}} &
  \multicolumn{1}{l}{\textbf{P}} &
  \multicolumn{1}{l}{\textbf{R}} &
  \multicolumn{1}{l|}{\textbf{F$_1$}} &
  \multicolumn{1}{l}{\textbf{P}} &
  \multicolumn{1}{l}{\textbf{R}} &
  \multicolumn{1}{l|}{\textbf{F$_1$}} &
  \multicolumn{1}{l}{\textbf{P}} &
  \multicolumn{1}{l}{\textbf{R}} &
  \multicolumn{1}{l}{\textbf{F$_1$}} \\ \hline
\multicolumn{17}{l}{\cellcolor[HTML]{E6F1FA}\textbf{GPT-4}} \\ \hline
 &
  \cellcolor[HTML]{FCEBEB}\textbf{Base} &
  \cellcolor[HTML]{FCEBEB}18.87 &
  \cellcolor[HTML]{FCEBEB}52.01 &
  \multicolumn{1}{l|}{\cellcolor[HTML]{FCEBEB}27.60} &
  \cellcolor[HTML]{FCEBEB}68.62 &
  \cellcolor[HTML]{FCEBEB}90.32 &
  \multicolumn{1}{l|}{\cellcolor[HTML]{FCEBEB}78.03} &
  \cellcolor[HTML]{FCEBEB}63.06 &
  \cellcolor[HTML]{FCEBEB}64.12 &
  \multicolumn{1}{l|}{\cellcolor[HTML]{FCEBEB}63.58} &
  \cellcolor[HTML]{FCEBEB}45.02 &
  \cellcolor[HTML]{FCEBEB}49.02 &
  \multicolumn{1}{l|}{\cellcolor[HTML]{FCEBEB}46.93} &
  \cellcolor[HTML]{FCEBEB}27.26 &
  \cellcolor[HTML]{FCEBEB}60.06 &
  \cellcolor[HTML]{FCEBEB}37.49 \\ \cline{2-17} 
 &
  \textbf{TF-IDF} &
  19.71 &
  51.25 &
  \multicolumn{1}{l|}{28.47} &
  {\color[HTML]{000000} \textbf{82.31}} &
  89.76 &
  \multicolumn{1}{l|}{\color[HTML]{CC0000} \textbf{85.88}} &
  {\color[HTML]{000000} \textbf{74.43}} &
  {\ul 78.14} &
  \multicolumn{1}{l|}{{\color[HTML]{CC0000} \textbf{76.24}}} &
  {\color[HTML]{000000} \textbf{56.86}} &
  {\ul 63.68} &
  \multicolumn{1}{l|}{{\color[HTML]{CC0000} \textbf{60.08}}} &
  27.22 &
  62.68 &
  37.96 \\ \cline{2-17} 
 &
  \textbf{SBERT} &
  {\color[HTML]{000000} \textbf{24.31}} &
  55.00 &
  \multicolumn{1}{l|}{{\color[HTML]{CC0000} \textbf{33.72}}} &
  76.63 &
  {\ul 91.41} &
  \multicolumn{1}{l|}{83.37} &
  72.63 &
  74.27 &
  \multicolumn{1}{l|}{73.44} &
  55.05 &
  60.30 &
  \multicolumn{1}{l|}{57.56} &
  28.05 &
  64.65 &
  39.12 \\ \cline{2-17} 
 &
  \textbf{ColBERT} &
  22.66 &
  56.79 &
  \multicolumn{1}{l|}{32.39} &
  78.64 &
  81.03 &
  \multicolumn{1}{l|}{{\color[HTML]{000000} 79.82}} &
  74.14 &
  77.02 &
  \multicolumn{1}{l|}{75.56} &
  50.43 &
  54.48 &
  \multicolumn{1}{l|}{52.38} &
  {\color[HTML]{000000} \textbf{28.14}} &
  {\ul 68.69} &
  {\color[HTML]{CC0000} \textbf{39.93}} \\ \cline{2-17} 
\multirow{-5}{*}{5-shot} &
  \textbf{DPR} &
  22.60 &
  {\ul 58.79} &
  \multicolumn{1}{l|}{32.64} &
  79.39 &
  88.24 &
  \multicolumn{1}{l|}{83.58} &
  69.77 &
  70.00 &
  \multicolumn{1}{l|}{69.89} &
  46.67 &
  52.39 &
  \multicolumn{1}{l|}{49.37} &
  {\color[HTML]{333333} 27.90} &
  65.49 &
  {\color[HTML]{333333} 39.13} \\ \hline
 &
  \cellcolor[HTML]{FCEBEB}\textbf{Base} &
  \cellcolor[HTML]{FCEBEB}22.25 &
  \cellcolor[HTML]{FCEBEB}56.66 &
  \multicolumn{1}{l|}{\cellcolor[HTML]{FCEBEB}31.92} &
  \cellcolor[HTML]{FCEBEB}75.33 &
  \cellcolor[HTML]{FCEBEB}88.31 &
  \multicolumn{1}{l|}{\cellcolor[HTML]{FCEBEB}81.27} &
  \cellcolor[HTML]{FCEBEB}66.38 &
  \cellcolor[HTML]{FCEBEB}74.24 &
  \multicolumn{1}{l|}{\cellcolor[HTML]{FCEBEB}70.09} &
  \cellcolor[HTML]{FCEBEB}53.23 &
  \cellcolor[HTML]{FCEBEB}52.13 &
  \multicolumn{1}{l|}{\cellcolor[HTML]{FCEBEB}52.67} &
  \cellcolor[HTML]{FCEBEB}26.67 &
  \cellcolor[HTML]{FCEBEB}59.20 &
  \cellcolor[HTML]{FCEBEB}36.74 \\ \cline{2-17} 
 &
  \textbf{TF-IDF} &
  21.53 &
  56.25 &
  \multicolumn{1}{l|}{31.14} &
  {\color[HTML]{000000} 83.81} &
  {\ul 89.67} &
  \multicolumn{1}{l|}{\color[HTML]{CC0000} \textbf{86.64}} &
  73.85 &
  77.29 &
  \multicolumn{1}{l|}{75.53} &
  {\color[HTML]{000000} \textbf{58.81}} &
  {\ul 65.66} &
  \multicolumn{1}{l|}{{\color[HTML]{CC0000} \textbf{62.05}}} &
  28.14 &
  71.42 &
  40.37 \\ \cline{2-17} 
 &
  \textbf{SBERT} &
  {\color[HTML]{000000} \textbf{25.41}} &
  {\ul 58.75} &
  \multicolumn{1}{l|}{{\color[HTML]{CC0000} \textbf{35.47}}} &
  83.94 &
  87.99 &
  \multicolumn{1}{l|}{85.92} &
  72.73 &
  75.08 &
  \multicolumn{1}{l|}{73.89} &
  58.79 &
  63.02 &
  \multicolumn{1}{l|}{60.83} &
  \textbf{28.32} &
  70.26 &
  40.37 \\ \cline{2-17} 
 &
  \textbf{ColBERT} &
  23.86 &
  58.02 &
  \multicolumn{1}{l|}{33.81} &
  83.49 &
  88.05 &
  \multicolumn{1}{l|}{{\color[HTML]{000000} 85.71}} &
  {\color[HTML]{000000} \textbf{74.69}} &
  {\ul 78.06} &
  \multicolumn{1}{l|}{{\color[HTML]{CC0000} \textbf{76.34}}} &
  55.12 &
  59.56 &
  \multicolumn{1}{l|}{57.25} &
  {\color[HTML]{000000} 28.15} &
  {\ul 71.99} &
  {\color[HTML]{CC0000} \textbf{40.48}} \\ \cline{2-17} 
\multirow{-5}{*}{10-shot} &
  \textbf{DPR} &
  22.96 &
  56.25 &
  \multicolumn{1}{l|}{32.61} &
  \textbf{85.16} &
  84.42 &
  \multicolumn{1}{l|}{84.79} &
  71.84 &
  72.42 &
  \multicolumn{1}{l|}{72.13} &
  56.82 &
  60.72 &
  \multicolumn{1}{l|}{58.70} &
  28.25 &
  70.04 &
  40.25 \\ \hline
 &
  \cellcolor[HTML]{FCEBEB}\textbf{Base} &
  \cellcolor[HTML]{FCEBEB}27.74 &
  \cellcolor[HTML]{FCEBEB}58.75 &
  \multicolumn{1}{l|}{\cellcolor[HTML]{FCEBEB}37.67} &
  \cellcolor[HTML]{FCEBEB}74.57 &
  \cellcolor[HTML]{FCEBEB}89.18 &
  \multicolumn{1}{l|}{\cellcolor[HTML]{FCEBEB}81.15} &
  \cellcolor[HTML]{FCEBEB}70.65 &
  \cellcolor[HTML]{FCEBEB}71.32 &
  \multicolumn{1}{l|}{\cellcolor[HTML]{FCEBEB}70.98} &
  \cellcolor[HTML]{FCEBEB}51.68 &
  \cellcolor[HTML]{FCEBEB}52.29 &
  \multicolumn{1}{l|}{\cellcolor[HTML]{FCEBEB}51.98} &
  \cellcolor[HTML]{FCEBEB}28.10 &
  \cellcolor[HTML]{FCEBEB}60.78 &
  \cellcolor[HTML]{FCEBEB}38.39 \\ \cline{2-17} 
 &
  \textbf{TF-IDF} &
  27.72 &
  62.20 &
  \multicolumn{1}{l|}{38.35} &
  85.41 &
  88.98 &
  \multicolumn{1}{l|}{87.16} &
  {\color[HTML]{000000} \textbf{75.81}} &
  {\ul 79.61} &
  \multicolumn{1}{l|}{{\color[HTML]{CC0000} \textbf{77.66}}} &
  {\color[HTML]{000000} \textbf{61.80}} &
  {\ul 67.13} &
  \multicolumn{1}{l|}{{\color[HTML]{CC0000} \textbf{64.36}}} &
  {\color[HTML]{000000} \textbf{28.20}} &
  {\ul 77.30} &
  {\color[HTML]{CC0000} \textbf{41.32}} \\ \cline{2-17} 
 &
  \textbf{SBERT} &
  28.44 &
  59.50 &
  \multicolumn{1}{l|}{38.22} &
  {\color[HTML]{000000} 85.37} &
  {\ul 89.57} &
  \multicolumn{1}{l|}{{\color[HTML]{CC0000} \textbf{87.42}}} &
  73.79 &
  76.54 &
  \multicolumn{1}{l|}{75.14} &
  60.89 &
  63.59 &
  \multicolumn{1}{l|}{62.21} &
  26.81 &
  74.09 &
  39.37 \\ \cline{2-17} 
 &
  \textbf{ColBERT} &
  {\color[HTML]{000000} \textbf{31.19}} &
  {\ul 66.67} &
  \multicolumn{1}{l|}{{\color[HTML]{CC0000} \textbf{42.49}}} &
  82.09 &
  83.94 &
  \multicolumn{1}{l|}{83.00} &
  75.27 &
  78.19 &
  \multicolumn{1}{l|}{76.70} &
  56.13 &
  59.35 &
  \multicolumn{1}{l|}{57.69} &
  27.70 &
  75.47 &
  40.53 \\ \cline{2-17} 
\multirow{-5}{*}{20-shot} &
  \textbf{DPR} &
  28.55 &
  60.75 &
  \multicolumn{1}{l|}{38.84} &
  \textbf{85.81} &
  85.40 &
  \multicolumn{1}{l|}{85.60} &
  71.82 &
  72.74 &
  \multicolumn{1}{l|}{72.28} &
  59.00 &
  61.74 &
  \multicolumn{1}{l|}{60.34} &
  27.16 &
  69.37 &
  39.23 \\ \hline
\multicolumn{17}{l}{\cellcolor[HTML]{E6F1FA}\textbf{Llama3-70B}} \\ \hline
 &
  \cellcolor[HTML]{FCEBEB}\textbf{Base} &
  \cellcolor[HTML]{FCEBEB}13.16 &
  \cellcolor[HTML]{FCEBEB}57.86 &
  \multicolumn{1}{l|}{\cellcolor[HTML]{FCEBEB}21.43} &
  \cellcolor[HTML]{FCEBEB}68.97 &
  \cellcolor[HTML]{FCEBEB}78.36 &
  \multicolumn{1}{l|}{\cellcolor[HTML]{FCEBEB}73.32} &
  \cellcolor[HTML]{FCEBEB}59.30 &
  \cellcolor[HTML]{FCEBEB}67.27 &
  \multicolumn{1}{l|}{\cellcolor[HTML]{FCEBEB}62.94} &
  \cellcolor[HTML]{FCEBEB}35.81 &
  \cellcolor[HTML]{FCEBEB}34.71 &
  \multicolumn{1}{l|}{\cellcolor[HTML]{FCEBEB}34.80} &
  \cellcolor[HTML]{FCEBEB}25.89 &
  \cellcolor[HTML]{FCEBEB}67.05 &
  \cellcolor[HTML]{FCEBEB}37.26 \\ \cline{2-17} 
 &
  \textbf{TF-IDF} &
  18.89 &
  58.62 &
  \multicolumn{1}{l|}{28.57} &
  \textbf{78.49} &
  81.78 &
  \multicolumn{1}{l|}{80.11} &
  66.48 &
  74.84 &
  \multicolumn{1}{l|}{70.41} &
  48.93 &
  {\ul 50.70} &
  \multicolumn{1}{l|}{49.80} &
  26.46 &
  72.06 &
  38.68 \\ \cline{2-17} 
 &
  \textbf{SBERT} &
  {\color[HTML]{000000} \textbf{23.20}} &
  {\ul 66.67} &
  \multicolumn{1}{l|}{{\color[HTML]{CC0000} \textbf{34.42}}} &
  {\color[HTML]{000000} 77.26} &
  {\ul 83.79} &
  \multicolumn{1}{l|}{{\color[HTML]{CC0000} \textbf{80.39}}} &
  64.04 &
  72.21 &
  \multicolumn{1}{l|}{67.88} &
  {\color[HTML]{000000} \textbf{50.66}} &
  49.59 &
  \multicolumn{1}{l|}{{\color[HTML]{CC0000} \textbf{50.12}}} &
  26.15 &
  68.92 &
  37.91 \\ \cline{2-17} 
 &
  \textbf{ColBERT} &
  22.05 &
  65.12 &
  \multicolumn{1}{l|}{32.94} &
  71.21 &
  72.33 &
  \multicolumn{1}{l|}{71.76} &
  {\color[HTML]{000000} \textbf{68.37}} &
  {\ul 75.32} &
  \multicolumn{1}{l|}{{\color[HTML]{CC0000} \textbf{71.68}}} &
  44.93 &
  46.08 &
  \multicolumn{1}{l|}{45.50} &
  {\color[HTML]{000000} \textbf{26.68}} &
  {\ul 72.38} &
  {\color[HTML]{CC0000} \textbf{38.99}} \\ \cline{2-17} 
\multirow{-5}{*}{5-shot} &
  \textbf{DPR} &
  19.20 &
  59.26 &
  \multicolumn{1}{l|}{29.00} &
  74.47 &
  76.91 &
  \multicolumn{1}{l|}{75.67} &
  65.74 &
  72.54 &
  \multicolumn{1}{l|}{68.97} &
  41.06 &
  48.66 &
  \multicolumn{1}{l|}{44.54} &
  26.51 &
  71.38 &
  38.66 \\ \hline
 &
  \cellcolor[HTML]{FCEBEB}\textbf{Base} &
  \cellcolor[HTML]{FCEBEB}22.37 &
  \cellcolor[HTML]{FCEBEB}59.94 &
  \multicolumn{1}{l|}{\cellcolor[HTML]{FCEBEB}32.50} &
  \cellcolor[HTML]{FCEBEB}72.56 &
  \cellcolor[HTML]{FCEBEB}77.91 &
  \multicolumn{1}{l|}{\cellcolor[HTML]{FCEBEB}75.15} &
  \cellcolor[HTML]{FCEBEB}59.13 &
  \cellcolor[HTML]{FCEBEB}71.63 &
  \multicolumn{1}{l|}{\cellcolor[HTML]{FCEBEB}63.77} &
  \cellcolor[HTML]{FCEBEB}39.67 &
  \cellcolor[HTML]{FCEBEB}31.49 &
  \multicolumn{1}{l|}{\cellcolor[HTML]{FCEBEB}35.60} &
  \cellcolor[HTML]{FCEBEB}25.57 &
  \cellcolor[HTML]{FCEBEB}64.33 &
  \cellcolor[HTML]{FCEBEB}36.50 \\ \cline{2-17} 
 &
  \textbf{TF-IDF} &
  23.53 &
  {\ul 62.65} &
  \multicolumn{1}{l|}{{\ul 34.21}} &
  80.82 &
  80.32 &
  \multicolumn{1}{l|}{80.57} &
  55.79 &
  55.34 &
  \multicolumn{1}{l|}{55.56} &
  49.59 &
  49.41 &
  \multicolumn{1}{l|}{49.50} &
  24.03 &
  68.00 &
  35.51 \\ \cline{2-17} 
 &
  \textbf{SBERT} &
  22.27 &
  59.76 &
  \multicolumn{1}{l|}{32.45} &
  77.72 &
  {\ul 84.94} &
  \multicolumn{1}{l|}{81.17} &
  67.67 &
  76.09 &
  \multicolumn{1}{l|}{71.63} &
  {\color[HTML]{000000} \textbf{52.84}} &
  {\ul 49.94} &
  \multicolumn{1}{l|}{{\color[HTML]{CC0000} \textbf{51.35}}} &
  {\color[HTML]{000000} \textbf{27.61}} &
  66.88 &
  {\color[HTML]{CC0000} \textbf{39.08}} \\ \cline{2-17} 
 &
  \textbf{ColBERT} &
  22.58 &
  60.50 &
  \multicolumn{1}{l|}{32.89} &
  78.40 &
  82.37 &
  \multicolumn{1}{l|}{80.34} &
  {\color[HTML]{000000} \textbf{69.65}} &
  {\ul 76.37} &
  \multicolumn{1}{l|}{{\color[HTML]{CC0000} \textbf{72.85}}} &
  38.72 &
  38.81 &
  \multicolumn{1}{l|}{38.77} &
  26.49 &
  67.58 &
  38.06 \\ \cline{2-17} 
\multirow{-5}{*}{10-shot} &
  \textbf{DPR} &
  {\color[HTML]{000000} \textbf{24.37}} &
  57.83 &
  \multicolumn{1}{l|}{{\color[HTML]{CC0000} \textbf{34.29}}} &
  {\color[HTML]{000000} \textbf{85.16}} &
  84.42 &
  \multicolumn{1}{l|}{{\color[HTML]{CC0000} \textbf{84.79}}} &
  65.85 &
  73.68 &
  \multicolumn{1}{l|}{69.54} &
  47.60 &
  45.04 &
  \multicolumn{1}{l|}{46.28} &
  25.80 &
  {\ul 70.97} &
  {\ul 37.85} \\ \hline
 &
  \cellcolor[HTML]{FCEBEB}\textbf{Base} &
  \cellcolor[HTML]{FCEBEB}24.52 &
  \cellcolor[HTML]{FCEBEB}53.81 &
  \multicolumn{1}{l|}{\cellcolor[HTML]{FCEBEB}33.67} &
  \cellcolor[HTML]{FCEBEB}\textbf{75.42} &
  \cellcolor[HTML]{FCEBEB}75.58 &
  \multicolumn{1}{l|}{\cellcolor[HTML]{FCEBEB}75.50} &
  \cellcolor[HTML]{FCEBEB}62.01 &
  \cellcolor[HTML]{FCEBEB}62.12 &
  \multicolumn{1}{l|}{\cellcolor[HTML]{FCEBEB}62.05} &
  \cellcolor[HTML]{FCEBEB}40.71 &
  \cellcolor[HTML]{FCEBEB}42.58 &
  \multicolumn{1}{l|}{\cellcolor[HTML]{FCEBEB}41.62} &
  \cellcolor[HTML]{FCEBEB}26.57 &
  \cellcolor[HTML]{FCEBEB}64.79 &
  \cellcolor[HTML]{FCEBEB}37.67 \\ \cline{2-17} 
 &
  \textbf{TF-IDF} &
  27.62 &
  66.95 &
  \multicolumn{1}{l|}{39.11} &
  {\color[HTML]{000000} 74.64} &
  {\ul 82.47} &
  \multicolumn{1}{l|}{{\color[HTML]{CC0000} \textbf{78.36}}} &
  55.95 &
  58.51 &
  \multicolumn{1}{l|}{57.66} &
  45.39 &
  49.83 &
  \multicolumn{1}{l|}{47.50} &
  {\color[HTML]{000000} \textbf{27.80}} &
  64.39 &
  {\color[HTML]{CC0000} \textbf{38.83}} \\ \cline{2-17} 
 &
  \textbf{SBERT} &
  {\color[HTML]{000000} \textbf{29.93}} &
  {\ul 68.06} &
  \multicolumn{1}{l|}{{\color[HTML]{CC0000} \textbf{41.43}}} &
  75.04 &
  80.75 &
  \multicolumn{1}{l|}{76.85} &
  \textbf{65.90} &
  64.77 &
  \multicolumn{1}{l|}{65.35} &
  42.09 &
  46.40 &
  \multicolumn{1}{l|}{44.14} &
  25.48 &
  61.36 &
  36.01 \\ \cline{2-17} 
 &
  \textbf{ColBERT} &
  23.57 &
  65.48 &
  \multicolumn{1}{l|}{34.66} &
  73.74 &
  70.70 &
  \multicolumn{1}{l|}{72.19} &
  58.25 &
  57.03 &
  \multicolumn{1}{l|}{57.63} &
  {\color[HTML]{000000} \textbf{47.08}} &
  {\ul 49.88} &
  \multicolumn{1}{l|}{{\color[HTML]{CC0000} \textbf{48.44}}} &
  25.41 &
  {\ul 66.98} &
  {\ul 36.85} \\ \cline{2-17} 
\multirow{-5}{*}{20-shot} &
  \textbf{DPR} &
  26.15 &
  65.04 &
  \multicolumn{1}{l|}{37.30} &
  72.58 &
  77.15 &
  \multicolumn{1}{l|}{74.80} &
  {\color[HTML]{000000} 62.72} &
  {\ul 69.19} &
  \multicolumn{1}{l|}{{\color[HTML]{CC0000} \textbf{65.80}}} &
  37.18 &
  44.13 &
  \multicolumn{1}{l|}{40.36} &
  26.10 &
  62.88 &
  36.89 \\ \hline
\end{tabular}%
}
\label{tab:dynamic_prompt}
\end{table}

The results in Table \ref{tab:dynamic_prompt} demonstrate the effectiveness of dynamic prompting in multiple FSL settings (5-shot, 10-shot, and 20-shot) for GPT-4 and LLaMA 3 across five biomedical datasets\footnote{We excluded GPT-3.5 due to its consistent underperformance relative to GPT-4 in the previous experiments.}. As described in the Section~\hyperref[sec:experimental-setup]{Experimental Setup}, the baseline prompts used randomly selected examples, and the results were averaged over four random runs. Detailed results for each random run, along with the averaged results, are presented in Table 1 to Table 5 in Supplementary Materials. 95\% confidence intervals (CIs) for each metric are provided in Table 7 in Supplementary Materials. We also provided examples of predictions in Supplementary Materials.

Both LLMs benefit significantly from retrieval-augmented methods. TF-IDF and ColBERT frequently produce the highest F$_1$-scores for both models. SBERT also consistently improves over the base method, especially for GPT-4. For GPT-4, TF-IDF retrieval outperforms other methods in most cases. For example, on the BC5CDR dataset, TF-IDF achieves the highest F$_1$-score of 85.9\% in the 5-shot setting, 86.6\% in the 10-shot setting, and 87.2\% in the 20-shot setting. Similarly, for the MIMIC III dataset, TF-IDF achieves the top F$_1$-score of 76.2\% in the 5-shot setting and 77.7\% in the 20-shot setting. In contrast, SBERT exhibits strong performance on the \textsc{Reddit-Impacts} dataset, where it achieves the highest F$_1$-scores of 33.7\% (5-shot) and 35.5\% (10-shot). Moreover, SBERT achieves an F$_1$-score of 41.43\% on \textsc{Reddit-Impacts} in the 20-shot setting, outperforming TF-IDF by a margin of 3.08\%. For LLaMA 3, DPR retrieval achieves competitive results, particularly on BC5CDR, where it achieves the highest F$_1$-scores of 84.8\% (10-shot) and 74.8\% (20-shot). SBERT also performs strongly on the \textsc{Reddit-Impacts} dataset, achieving F$_1$-scores of 34.4\% (5-shot) and 41.4\% (20-shot).

\begin{figure}[htbp]
    \centering
    \begin{subfigure}[b]{\textwidth}
        \centering
        \includegraphics[width=0.7\textwidth]{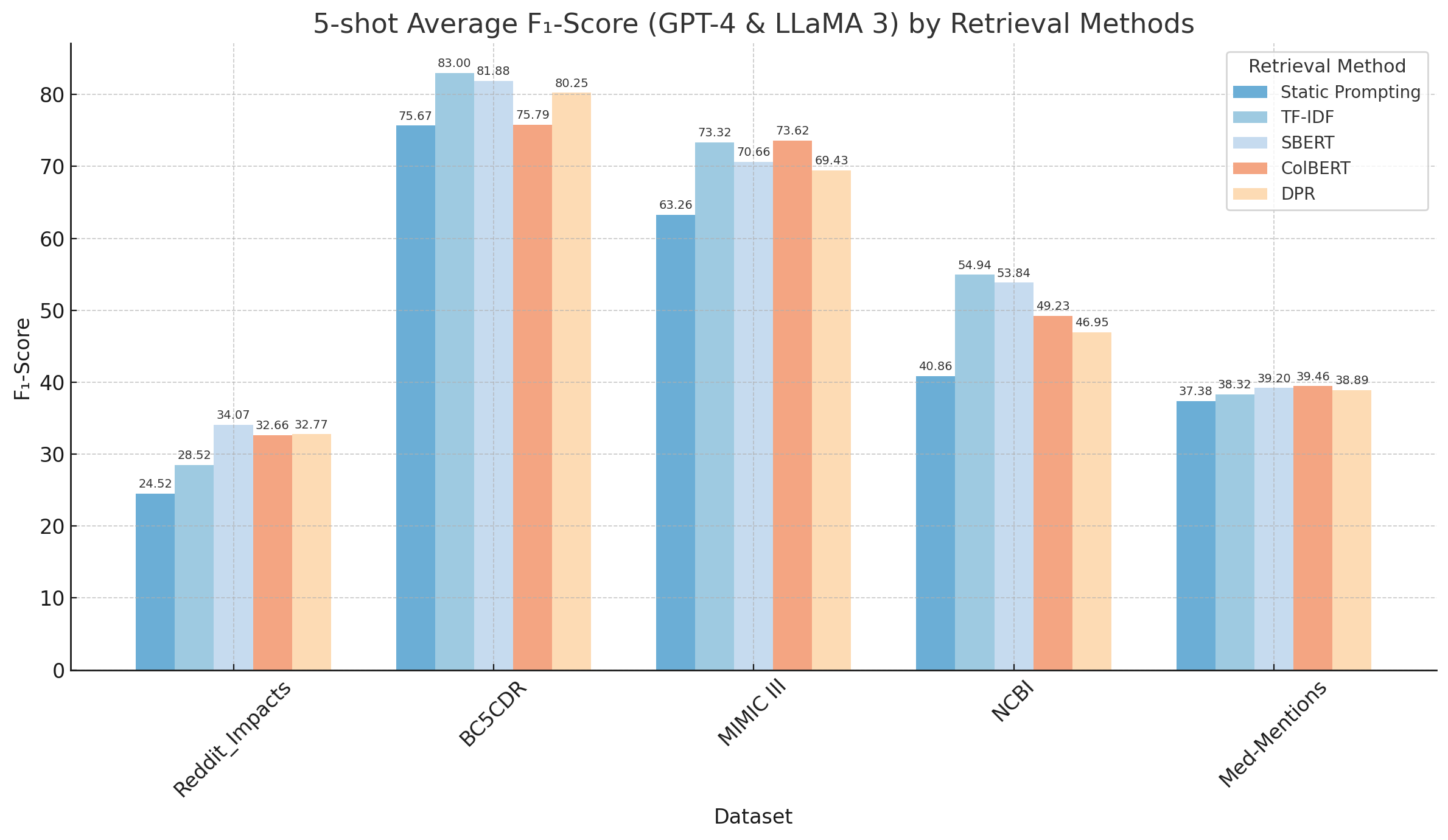} 
    \end{subfigure}
    
    \vspace{1em} 
    
    \begin{subfigure}[b]{\textwidth}
        \centering
        \includegraphics[width=0.7\textwidth]{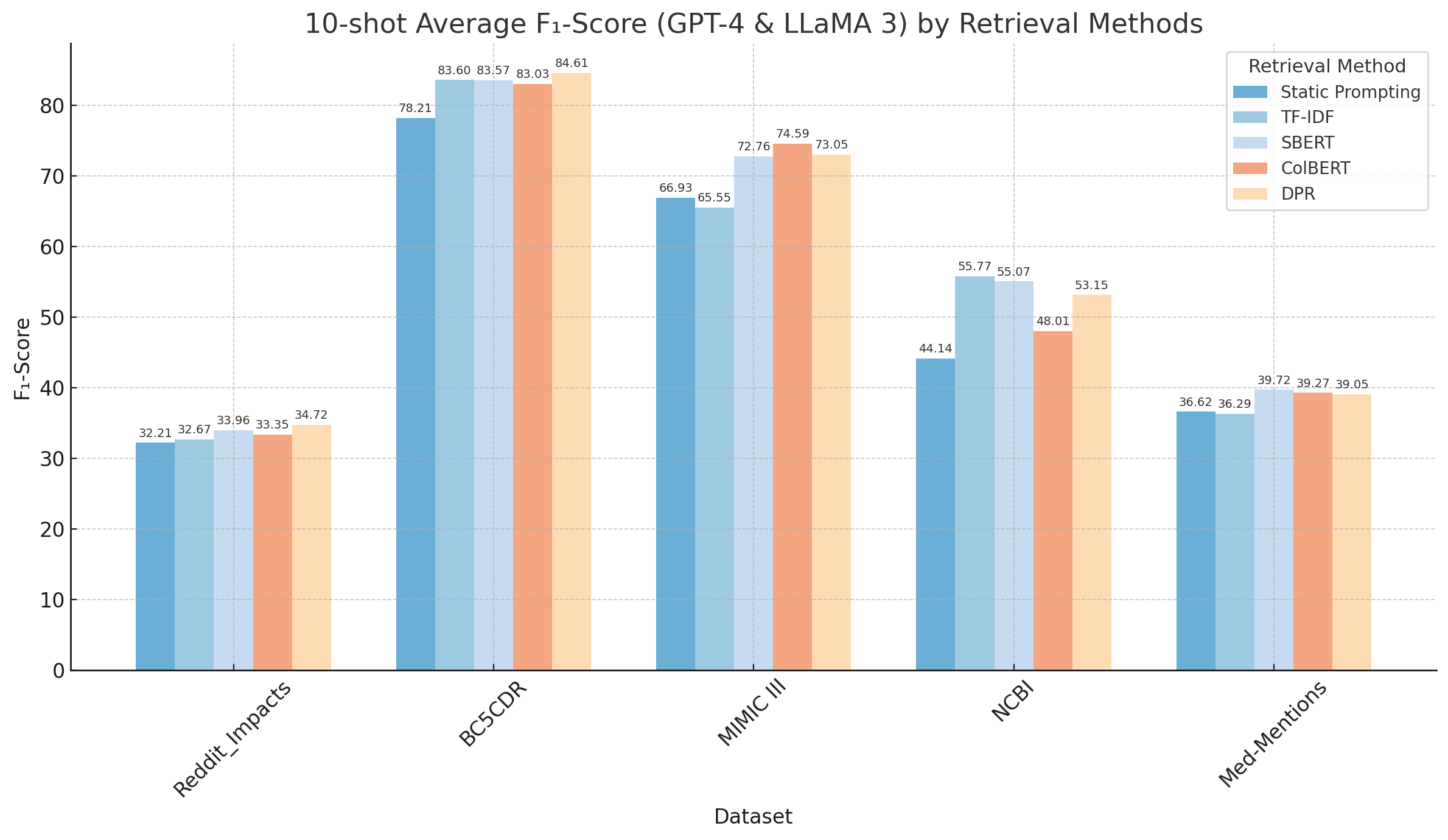} 
    \end{subfigure}
    
    \vspace{1em} 
    
    \begin{subfigure}[b]{\textwidth}
        \centering
        \includegraphics[width=0.7\textwidth]{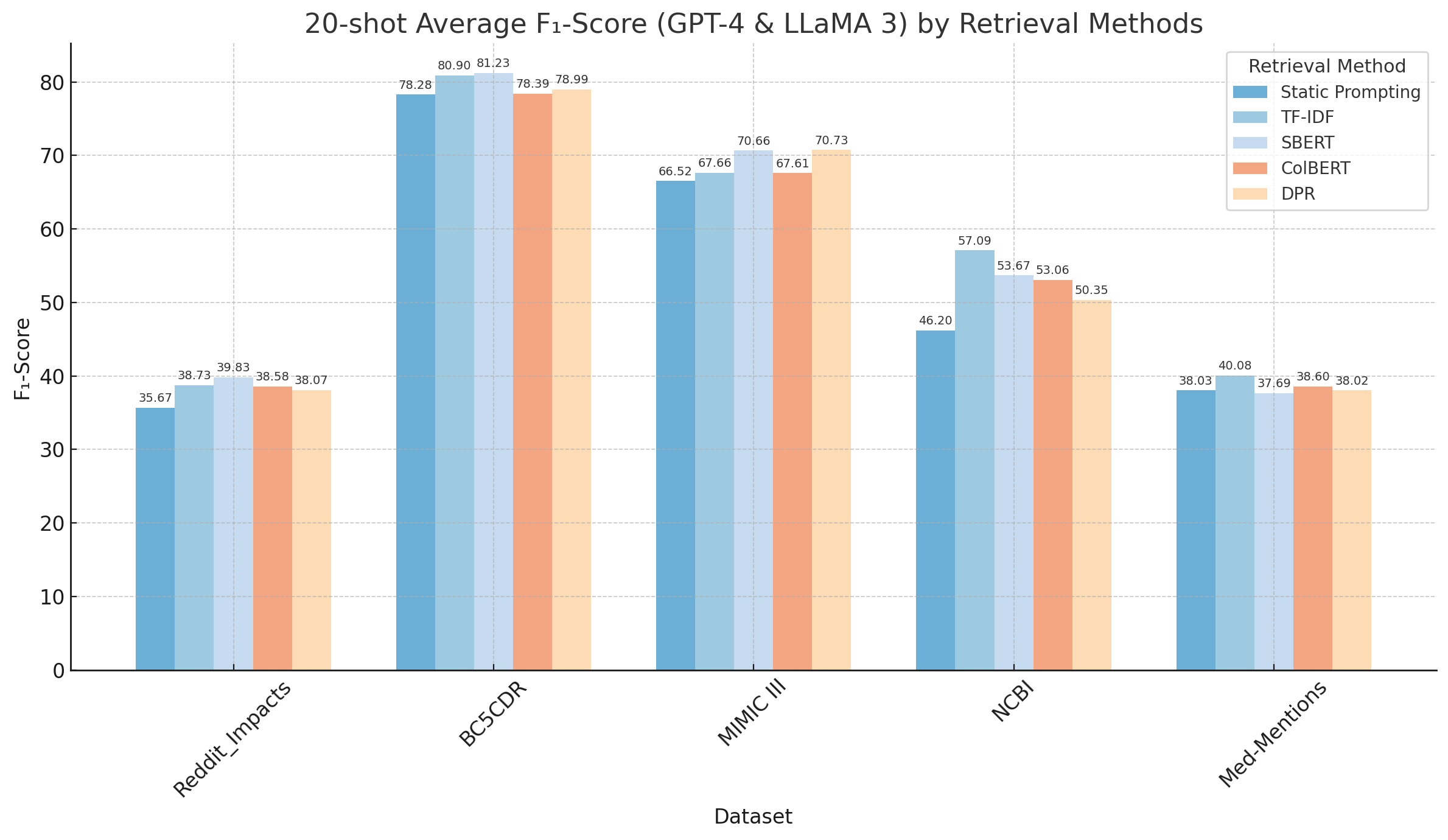} 
    \end{subfigure}

    \caption[Comparison of Average F$_1$-scores]{Comparison of average F$_1$-scores for GPT-4 and LLaMA 3 models across different datasets under varying shot settings.}
    \label{fig:comparison}
\end{figure}

Figure \ref{fig:comparison} presents the F$_1$-scores of five biomedical datasets by using different retrieval methods compared to static prompting for multiple shot settings: 5-shot, 10-shot, and 20-shot. The results are averaged across evaluations conducted using GPT-4 and LLaMA 3 models. Across all settings, all retrieval-based methods show significant improvements, demonstrating the benefit of incorporating retrieval methods into the prompting strategy.

\paragraph{1. 5-shot Analysis:} The SBERT retrieval engine achieves the highest average F$_1$-score for the \textsc{Reddit-Impacts} dataset (34.1\%), while TF-IDF performs best on BC5CDR (83.0\%) and NCBI (54.9\%). For MIMIC III, ColBERT leads with an F$_1$-score of 73.6\%, and on Med-Mentions, ColBERT also stands out with a top score of 39.5\%. TF-IDF achieved an average F$_1$-score improvement of 7.28\% across all datasets in the 5-shot setting, followed closely by SBERT with 7.46\%. ColBERT and DPR showed more modest improvements, with 5.76\% and 4.81\%, respectively. These results highlight the dataset-specific strengths of different retrieval methods, with TF-IDF showing strong performance on entity-rich datasets like BC5CDR and NCBI.

\paragraph{2. 10-shot Analysis:} The SBERT engine again stands out as the best-performing retrieval method overall, achieving the highest F$_1$-scores on three datasets: \textsc{Reddit-Impacts} (34.0\%), Med-Mentions (39.7\%), and NCBI (56.1\%). DPR achieves the top score on BC5CDR (84.8\%), while ColBERT performs best on MIMIC III with an F$_1$-score of 74.6\%. Consistent with these findings, SBERT also demonstrates the largest overall improvement over the base method, with an average gain of 5.59\%. It is followed by ColBERT with 4.87\% and DPR with 4.50\%, while TF-IDF shows the smallest improvement in the 10-shot setting, with an average increase of 3.49\%. These results highlight a departure from the 5-shot setting, where TF-IDF dominated, indicating that SBERT is better suited for slightly larger data scenarios.

\paragraph{3. 20-shot Analysis:} TF-IDF once again demonstrates strong performance, achieving the highest F$_1$-scores on three datasets: BC5CDR (82.8\%), NCBI (55.9\%), and Med-Mentions (40.1\%). SBERT leads on \textsc{Reddit-Impacts} with a top score of 39.8\%, while it also performs best on MIMIC III with an F$_1$-score of 70.2\%. Among the retrieval approaches, in the 20-shot setting, TF-IDF achieves the highest average improvement with 3.96\%, followed closely by SBERT with 3.55\%. DPR shows a moderate improvement with 2.08\%, while ColBERT exhibits the lowest increase of 1.95\%. These results highlight TF-IDF's and SBERT's consistent robustness across multiple datasets as the top-performing retrieval method.

Overall, GPT-4 consistently achieves higher F$_1$-scores compared to LLaMA 3 across most datasets and retrieval methods, particularly in 5-shot setting, with an average F$_1$ score 17.3\% higher than LLaMA 3. In the 10-shot setting, this gap narrows to 5.47\%, but GPT-4 still maintains a clear advantage. Finally, in the 20-shot setting, GPT-4 surpasses LLaMA 3 by an average F$_1$ score of 8.30\%. This improvement becomes even more significant in datasets with sparse or noisy data, where retrieval-augmented methods play a critical role. LLaMA 3 shows comparable performance in 20-shot setting but struggles to close the gap with GPT-4 in scenarios with fewer examples or more noisy data. This highlights GPT-4's robustness in leveraging limited training data.

Across all datasets and shot settings on GPT 4, larger training sizes (20-shot) tend to yield higher F$_1$-scores, precision, and recall. Specifically, from 5-shot to 10-shot, the mean F$_1$-score increases by 2.51\%, while precision and recall improve by 2.22\% and 1.02\%, respectively. However, from 10-shot to 20-shot, the performance gains are notably smaller, with F$_1$-score increasing by 1.64\%, precision by 0.16\%, and recall by only 0.05\%. Overall, comparing 5-shot to 20-shot, the models achieve a cumulative improvement of 4.16\% in F$_1$-score, 2.38\% in precision, and 1.07\% in recall. However, on LLaMA 3, this increase is less consistent, with the best performance observed at the 10-shot setting across all datasets except for the \textsc{Reddit-Impacts} dataset. From 10-shot to 20-shot, mean F$_1$-score decreases by -1.46\%, mean precision drops by -1.36\%, and mean recall declines by -1.75\%. The combination of effective retrieval methods and larger shot sizes (more examples) contributes significantly to the overall improvements observed in model performance across all datasets.

\section*{Discussion}
NER is one of the most commonly applied NLP tasks, and while the emergence of LLMs have led to substantial leaps in few-shot NER performance, innovative strategies are needed to address some of their limitations for real-life application in biomedicine. Improving FSL NER methods involving LLMs has the potential to substantially reduce the time and cost required for manual annotation. The methods proposed in this paper, validated on multiple standardized datasets with differing characteristics, present an important step towards operationalizing automated NER from biomedical texts, including in healthcare settings.

Our extensive empirical explorations revealed findings that will be useful for future research and application in this space. First, GPT-4 consistently outperforms GPT-3.5 and LLaMA 3-70B across datasets and configurations, demonstrating its robustness in understanding nuanced biomedical information. The consistent high performance of GPT-4 may be attributable to several factors. GPT-4 has significantly more parameters compared to GPT-3.5 and LLaMA 3-70B, enabling it to capture finer-grained contextual nuances, especially in complex and domain-specific tasks. Furthermore, in datasets with sparse or ambiguous annotations, such as \textsc{Reddit-Impacts} or Med-Mentions, GPT-4 achieves higher recall, indicating its ability to identify relevant entities and relationships more comprehensively. Further, retrieval engines improve performance by providing task-relevant context that enhances the model's understanding of the input, effectively bridging the gap between the model's general pretraining knowledge and the specific requirements of the task. 
Our results broadly show that TF-IDF based retrieval works well for datasets that have low noise and limited out-of-vocabulary expressions, despite its simplicity. In contrast, engines like SBERT perform better on linguistically diverse datasets, especially on \textsc{Reddit-Impacts} dataset, by leveraging semantic embeddings, which capture nuanced relationships between words and phrases. Advanced retrieval methods like ColBERT and DPR generally underperformed compared to TF-IDF and SBERT. This may be due to several reasons. ColBERT and DPR rely on dense representations, which, while powerful for general-purpose semantic matching, may fail to capture the precise, domain-specific distinctions critical in biomedical datasets. Furthermore, their reliance on dense embeddings can sometimes overfit to irrelevant semantic similarities, retrieving documents that are semantically related but not contextually relevant to the query. 

Given these findings, our results suggest that TF-IDF is the most efficient option for retrieval in datasets with low noise, while SBERT is better suited for handling linguistically diverse data. ColBERT and DPR, despite their strengths in general-purpose retrieval, do not provide substantial advantages in this domain and may introduce unnecessary computational overhead. Thus, for biomedical applications requiring high precision and efficiency, TF-IDF and SBERT offer the best balance of performance and efficiency.
The effect of shot size on performance is not uniform, as observed in the results across datasets. While increasing the shot size from 5 to 20 generally improves F$_1$-scores, the extent of improvement is dataset-dependent. Datasets with formal texts, like BC5CDR, which already benefit from the inclusion of the retrieval engine, exhibit marginal gains with additional examples. In contrast, noisy datasets like \textsc{Reddit-Impacts} are more sensitive to shot size, as more examples help the model adapt to diverse linguistic patterns and reduce misclassifications. 20-shot does not always yield the best results. One reason is diminishing returns: as the number of examples increases, redundancy or noise may be introduced, especially in datasets where retrieval engines already provide strong task-specific context. Another potential reason arises from the inherent constraints of LLMs, such as input token limits. As the shot size grows, the available space for processing task-specific context diminishes, potentially diluting the effectiveness of the prompt or truncating important information. 
Our findings suggest that for NER tasks involving sparsely-occurring entities, RAG-based dynamic prompting is likely to obtain better performance compared to optimized, static prompts. For retrieval, performances of the different engines were mostly comparable, and TF-IDF and SBERT consistently performed well. As the number of training instances increased, the impact of dynamic prompting over static prompting (Base), became less visible. This is expected since in high-shot settings, random draws of training instances contain considerable diversity to enable model generalization. It is also possible that as the number of examples provided for in-context learning increases, the overall increase in the length of the prompt diminishes the performance of the LLMs. The influence of input text length and LLM performance is an area of active research~\cite{levy2024same}.

\section*{Methods}

\phantomsection 
\addcontentsline{toc}{subsection}{Static Prompt Engineering}
\subsection*{Static Prompt Engineering}
\label{sec:static-prompt}

\begin{figure}[ht]
  \includegraphics[width=1.0\textwidth]{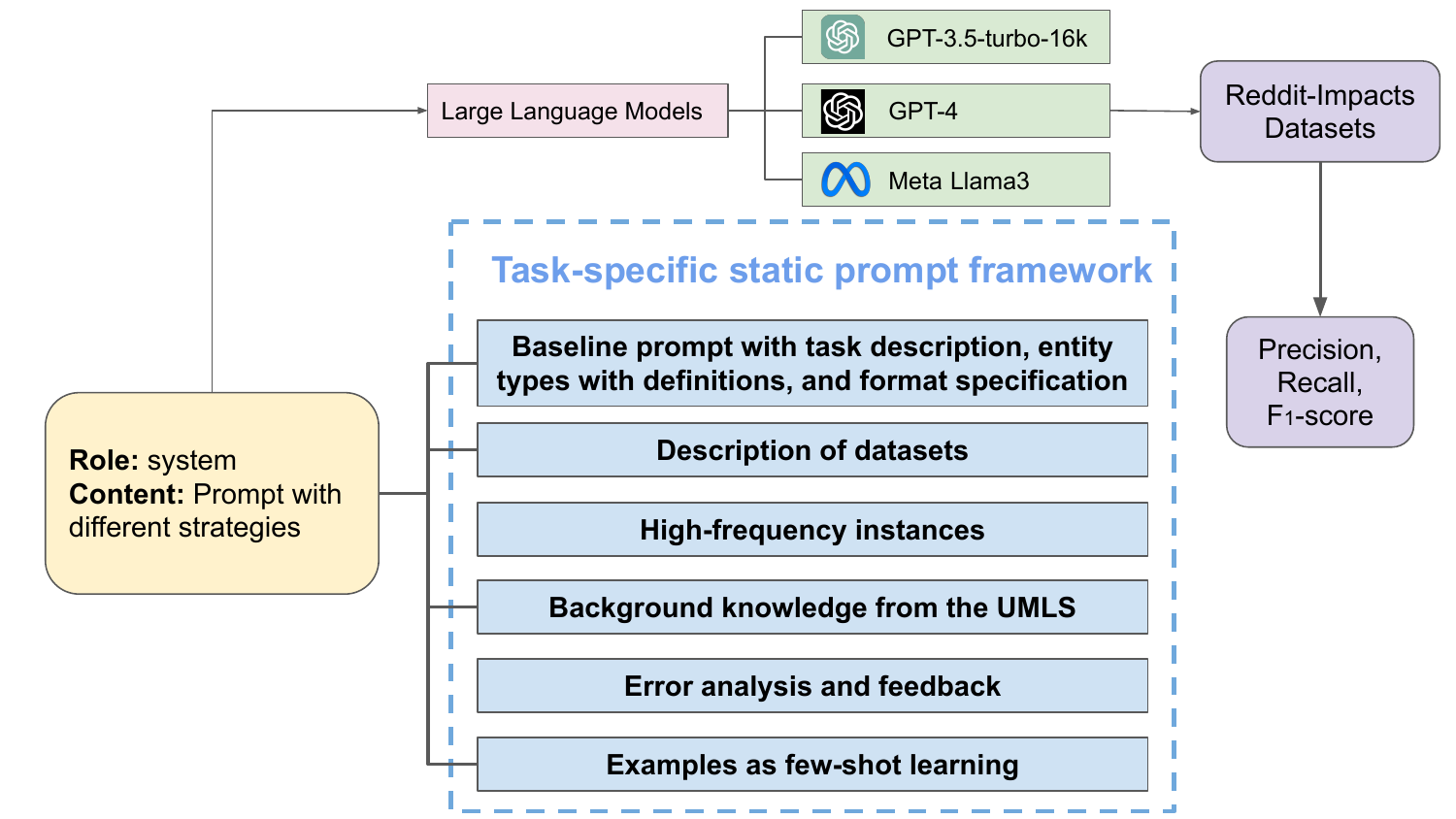}
  \caption[Overview of Task-specific Static Prompting]{An overview of the NER strategy based on static prompting on three LLMs. Static prompts containing different information are provided to the LLMs, which, in turn, generate predictions for evaluation.}
  \label{fig:overflow}
\end{figure}

Figure \ref{fig:overflow} presents the components of the static prompt we optimized for the LLMs. We systematically designed task-specific static prompts comprising the following components:

\paragraph{1. Baseline prompt with task description, entity types with definitions, and format specification:} The baseline component provides the LLM with essential information regarding the primary aims of the task, which is extracting and classifying entities. The categories of labels present in the dataset, along with their definitions. Entity definitions provide detailed and unequivocal explanations of an entity in the context of a specific task, crucially guiding the LLM toward accurately pinpointing entities within texts. Also, we provided the input, and instructions regarding the output format in the base prompt. For generative LLMs, NER presents greater challenges, relative to classification, as it is essentially a sequence-to-sequence problem, where each token is assigned a corresponding label. However, when a prompt includes a sentence as is, we found that LLMs may struggle to accurately assign labels to each token, resulting in mismatches in the number of input tokens (as annotated in the dataset) and output tokens. This issue is exacerbated by the fact that LLMs have their own tokenization mechanisms, which may differ from the tokenization in the annotated data. If the input and labels are provided in the BIO format instead, it often results in degraded performance due to the LLM's inability to fully understand the text.

One input approach is to provide a text and indicate the entities within it~\cite{xie2023empirical}. For example, in the sentence 'I was a codeine addict,' the phrase 'codeine addict' is identified as an entity and is annotated as `Clinical Impacts'. However, this format can become ambiguous when faced with long sentences that contain the same word or phrase multiple times, each with different contextual meanings, not all of which may be labeled as the relevant entity. Another input method involves providing spans corresponding to the entities~\cite{ocad259}, but this also causes mismatches between spans and entities frequently when generative LLMs are used.

To address these challenges, we adopt a new format for constructing the input and output for the LLMs. We provide LLMs with a list of tokens that have already been tokenized. For the output, we instruct the model to return each token ,concatenated with its corresponding label. This method allows us to easily extract labels for evaluation, and it ensures a one-to-one correspondence between the predicted labels and tokens, with the number of labels always consistent with the number of tokens in the input sentence.

For example:\\
\begin{quote}
    
Input: [`I', `was', `a', `codeine', `addict.']\\
Output: [`I-O', `was-O', `a-O', `codeine-B-Clinical\_Impacts', `addict.I-Clinical\_Impacts']
\end{quote}

To minimize the potential loss of sentence context caused using only tokens, we also explored the effectiveness of using the untokenized sentences as input, and tagged tokens as output: \\
\begin{quote}
    
Input: [`I was a codeine addict.']\\
Output: [`I-O', `was-O', `a-O', `codeine-B-Clinical\_Impacts', `addict.I-Clinical\_Impacts']
\end{quote}

\paragraph{2. Description of datasets:} By describing a dataset's origin, content, and themes, we aim to provide LLMs with a basic understanding of the dataset. For example, for the \textsc{Reddit-Impacts} dataset, we described that it focuses on individuals who use opioids, and we are interested in the impact of opioid use on their health and lives. 

\paragraph{3. High-frequency instances:} Some entities do not have clear definitions, and the determination is more ambiguous. Therefore, we provide the most frequently occurring words or phrases in each entity type within the training dataset to assist LLMs in understanding the potential distribution of entities and the theme of the text for this task. Specifically, we selected high-frequency instances for each class by computing word frequencies from lexicon-annotated data, ranking them by occurrence, and choosing the top 6 high-frequency words as label words for each class. This approach ensures that the selected label words effectively reflect the data distribution and help the model predict appropriate class labels at entity positions. By adding high-frequency instances, we tried to provide a LLM with a lexicon of the concepts of interest.

\paragraph{4. Incorporation of background knowledge from the UMLS:} We provide LLMs with comprehensive and structured information we obtained from the UMLS. Our intuition, based on the findings reported in prior work, was that this knowledge could enhance the understanding and interpretation of biomedical concepts, relationships, and terminologies.

\paragraph{5. Error analysis and feedback:} To improve the model's accuracy and address prediction errors, we provide an error analysis and feedback mechanism. After an initial set of predictions was made by LLMs on unseen training set instances, we manually reviewed the errors by comparing the model's predictions with the gold standard annotations. For each incorrect prediction, we analyze the type and cause of the error, such as misclassification, missed entities, or spurious entities. Based on this analysis, we provide a summarization of feedback to the model. This feedback includes only general descriptions of errors without any examples. While this element of the prompt requires preliminary explorations of the dataset, common possible errors can be identified easily using a small set of training examples (\textit{e.g.,} 5-shot), and this enables a mechanism of incorporating expert feedback into the process. 

\paragraph{6. Annotated samples:} We provide \textit{k} annotated instances within the prompt for in-context learning. Samples are randomly selected and formatted according to the task description and entity markup guide.

We compared the effectiveness of different components of static prompting by incrementally incorporating descriptions of datasets, high-frequency instances, background knowledge from the UMLS, error analysis and feedback, and varying k-shot annotated samples. Detailed prompts used for each dataset are provided in Table 8 to Table 12 in Supplementary Materials. 

\subsection*{Dynamic Prompt Engineering}
In prompt-based strategies using LLMs for in-context learning, the common approach has been to provide the model with a static prompt to guide its predictions. These prompts often include example instances, and CoT prompting. However, a significant limitation of this approach is that the provided examples may differ substantially from the texts from which the model is expected to extract named entities. Note that even in the presence of additional annotated samples, the LLMs context window size may limit the number of instances that can be embedded in a prompt for in-context learning. A static prompt, thus, does not generalize well, leading to high variance in performance.

\begin{figure}[h!]
\centering
\includegraphics[width=1.0
\columnwidth]{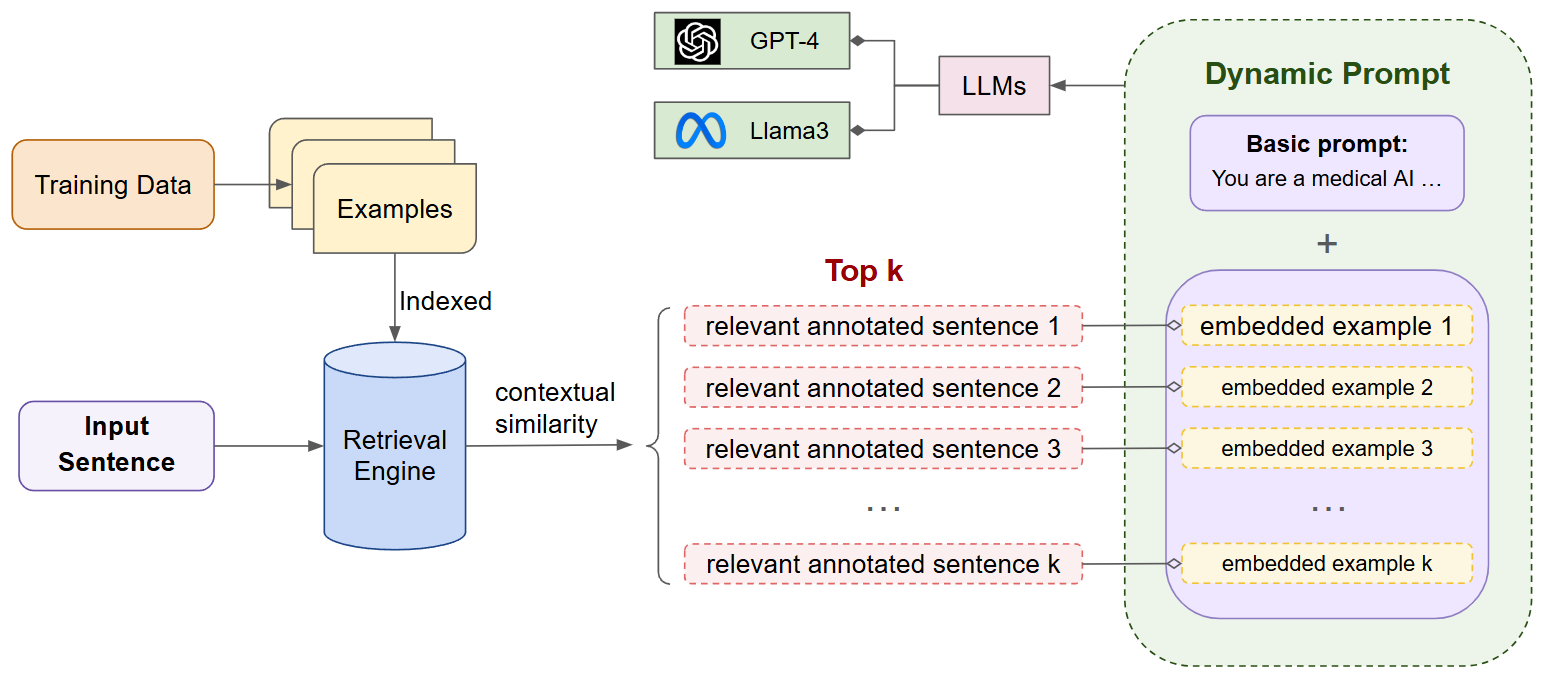}  
\caption[Overview of Retrieval-based Dynamic Prompting]{Overview of Retrieval-based Dynamic Prompting model. First the training data are provided to the retrieval engine for indexing. During inference, the system first ranks all training examples based on contextual similarity with the input text. Finally, the top \textit{n} retrieved instances are embedded in the prompt, which is passed to the LLM (e.g., GPT-4, LLaMA 3).}  
\label{fig:dynamic_prompt}
\end{figure}

To address this issue, we attempted to improve upon static prompting and adopted a dynamic approach involving RAG. In our proposed approach, a retrieval engine is first indexed with the annotated examples from the training set. Upon receiving an input sentence, the system first retrieves the top \textit{n} annotated examples using the retrieval engine. The retrieved examples are then embedded into the prompt, which is then passed to the LLM along with the input text. Figure \ref{fig:dynamic_prompt} presents an overview of the system architecture.

\subsubsection*{Retrieval Engines}
Selecting an effective retrieval engine is crucial since the examples embedded in the prompt influence the model's performance. We considered several retrieval methods, each chosen for its unique strengths in handling diverse biomedical texts, and applicability in FSL settings. The engines we selected are: TF-IDF~\cite{sparck1972statistical}, Sentence-BERT (SBERT)~\cite{reimers2019sentence}, ColBERT~\cite{khattab2020colbert}, and Dense Passage Retrieval (DPR)~\cite{karpukhin2020dense}. These search mechanisms offer a range of capabilities, from efficient keyword matching to advanced deep-learning-based retrieval. We provide further details below.

    \paragraph{1. TF-IDF:} Term Frequency-Inverse Document Frequency (TF-IDF) scores the relevance of documents based on the frequency of terms. We included TF-IDF due to its efficiency and simplicity, which allows for rapid retrieval of relevant examples based on keyword overlap. While it lacks semantic understanding, it serves as a strong baseline, particularly when the input contains well-defined biomedical terminologies.

    \paragraph{2. Sentence-BERT (SBERT):} SBERT leverages a pre-trained BERT model fine-tuned for semantic similarity tasks. By encoding input sentences into dense embeddings, SBERT can capture the semantic relationships between sentences, making it well-suited for identifying contextually similar examples even when the input phrasing differs from the training data. This capability is particularly advantageous in the biomedical domain, where synonymous terms and varied expressions are common.

    \paragraph{3. ColBERT:} ColBERT (Contextualized Late Interaction over BERT) enhances retrieval performance by focusing on contextualized token representations. It uses a late-interaction mechanism that allows for more nuanced matching of query and document tokens. We selected ColBERT for its ability to capture fine-grained semantic details, which is essential for handling complex biomedical texts with diverse and context-dependent entity mentions.

    \paragraph{4. Dense Passage Retrieval (DPR):} DPR employs a dual-encoder architecture, where separate encoders are used for queries and documents. It uses deep neural networks to learn dense embeddings, optimizing for maximum similarity between relevant query-document pairs. DPR’s strength lies in its ability to handle open-domain retrieval tasks effectively, making it a powerful choice for dynamically selecting annotated examples that are highly relevant to the input text, thus improving the contextual adaptability of our dynamic prompts.

In our experiments, we evaluated the performance of each retrieval method, assessing their impact on few-shot NER across multiple biomedical datasets.

\phantomsection 
\addcontentsline{toc}{subsection}{Experimental Setup}
\subsection*{Experimental Setup}
\label{sec:experimental-setup}
Below, we report our experimental setup for the two prompting strategies---Static Prompting and Dynamic Prompting.

\paragraph{1. Static Prompting:} For static prompting, we evaluated three language models: GPT-3.5, GPT-4, and LLaMA 3. We used prompts containing five examples per label to provide context and guide the models' predictions. For GPT-3.5, we used the OpenAI API version "2023-07-01-preview", and for GPT-4, we used the version "2024-02-15-preview". Both models were configured with the following settings: temperature = 0.2, top\_p = 0.1, frequency\_penalty = 0, presence\_penalty = 0, and no stop tokens specified. 

For LLaMA 3, we used the Meta-Llama-3-70B-Instruct model, with a temperature of 0.5 and top\_p of 0.95. Preliminary experiments (reported later in this chapter) revealed that GPT-3.5 consistently performed significantly worse compared to GPT-4. Hence, we excluded GPT-3.5 from further experiments in the dynamic prompting phase to limit API usage costs. To ensure robustness in the static prompting phase, the few-shot examples were randomly selected four times, and the reported results are the average of these four random selections.

\paragraph{2. Dynamic Prompting:} In the dynamic prompting phase, we focused on evaluating GPT-4 and LLaMA 3 on multiple datasets. We conducted experiments using three different in-context settings: 5-shot, 10-shot, and 20-shot, to assess the impact of increasing the number of examples on the model’s performance. The baseline prompts in this phase also used randomly selected examples, with the results averaged over four random runs. 

The evaluations were conducted on five biomedical datasets: MIMIC-III (clinical notes dataset), BC5CDR (disease and chemical entity recognition), NCBI-Disease (disease annotations from PubMed abstracts), Med-Mentions (large-scale UMLS concepts dataset), and our \textsc{Reddit-Impacts} dataset (annotated for clinical and social impacts entity extraction). Further details about these datasets are provided in Chapter 3. We used precision (P), recall (R), and F$_1$-score (F$_1$) as evaluation metrics to comprehensively asses the models’ performance across different datasets. In addition, to account for the variability in performance across different experimental runs, we include 95\% confidence intervals (CIs)~\cite{morey2008confidence} for each metric, providing a measure of the statistical robustness of the results. The confidence intervals were computed via bootstrap resampling~\cite{efron1992bootstrap} with 1000 samples with replacement.

\section*{Data availability}
We utilized five distinct medical text datasets as benchmarks to evaluate the performance of our models and to support the development of new approaches. These datasets provide a diverse range of clinical narratives and biomedical information, allowing for a comprehensive assessment of our methods.

\paragraph{MIMIC III}\cite{johnson2016}. The MIMIC III dataset is a large, publicly available database with patient data from critical care units, including medications, lab results, clinical notes, diagnostic codes, imaging reports, and survival data. It is widely used for few-shot classification and NER tasks.

\paragraph{BC5CDR}\cite{li2016biocreative}. This resource extracts relationships between chemicals and diseases from annotated biomedical articles, aimed at developing systems to automatically identify these interactions for applications like drug discovery, toxicology, and understanding disease mechanisms.

\paragraph{Med-Mentions}\cite{mohan2019medmentions}. Med-Mentions is a large biomedical corpus annotated with UMLS concepts, containing PubMed articles linked to entities like diseases, chemicals, genes, and anatomical terms. It supports tasks such as information extraction, literature mining, and knowledge base construction.

\paragraph{NCBI-Disease}\cite{dougan2014ncbi}. This dataset contains PubMed abstracts annotated with disease names, linked to standardized concepts in MeSH and OMIM databases. It is used to train and evaluate models for recognizing and normalizing disease names in biomedical texts. 

\paragraph{\textsc{Reddit-Impacts}}\cite{ge2024reddit}. a challenging NER dataset curated from subreddits dedicated to discussions on prescription and illicit opioids, as well as medications for opioid use disorder. This dataset includes posts from 14 opioid-related subreddits, and specifically focuses on the clinical and social impacts of nonmedical substance use.

\renewcommand{\arraystretch}{1.5}

\begin{table*}[h!]
\centering
\caption[Statistics of publicly available datasets]{Statistics of the eight standardized biomedical datasets we used, including the source and aim of their tasks, training and test sizes (number of tokens), the number of entity types and the number of entities in each dataset.}
\resizebox{\textwidth}{!}{%
\begin{tabular}{lllll}
\hline
\textbf{Datasets}                                                          & \textbf{Training Size} & \textbf{Test Size} & \textbf{Entity Types} & \textbf{Entities} \\ \hline
\textbf{MIMIC III (information relating to patients)}   & 36.4k  & 6.4k   & 12 & 8.7k   \\ \hline
\textbf{BC5CDR (extracting relationships between chemicals and diseases)}  & 228.8k                 & 122.2k             & 2                     & 28.8k             \\ \hline
\textbf{Med-Mentions (annotated with UMLS concepts)}    & 847.9k & 593.6k & 1  & 340.9k \\ \hline
\textbf{NCBI Disease (PubMed abstracts annotated with disease names)}      & 134.0k                 & 20.5k              & 4                     & 6.3k              \\ \hline
\textbf{\textsc{Reddit-Impacts} (clinical impacts and social impacts collected from Reddit)}      & 30.0k                 & 6.0k              & 2                     & 0.2k              \\ \hline
\end{tabular}%
}
\label{tab:statistics-publicly-available-datasets}
\end{table*}
Table \ref{tab:statistics-publicly-available-datasets} presents relevant statistics for all publicly available datasets we used in this study, including the source and aim of each dataset, training and test set sizes, the number of entity types and the number of entities in each dataset.

\section*{Code availability}
All data used in this study was publicly available. The code used for the experiments can be accessed at \href{https://github.com/Yao-Ge-1218AM/RAG-based-Dynamic-Prompting/}{https://github.com/Yao-Ge-1218AM/RAG-based-Dynamic-Prompting/}.

\section*{Acknowledgments}
Research reported in this publication was supported by the National Institute on Drug Abuse of the National Institutes of Health (NIH) under award number R01DA057599. The content is solely the responsibility of the authors and does not necessarily represent the official views of the NIH.

\section*{Author contributions}
YGe led model design and implementation, analysis, evaluation, visualization, and original draft preparation. YGuo contributed to the software development and implementation. SD assisted with analysis, evaluation, and manuscript preparation. AS conceptualized the study and supervised the study. All authors contributed to the writing of the final article.

\section*{Competing Interests}
All authors declare no competing interests.

\newpage
\bibliography{sample}

\begin{thebibliography}{10}
\urlstyle{rm}
\expandafter\ifx\csname url\endcsname\relax
  \def\url#1{\texttt{#1}}\fi
\expandafter\ifx\csname urlprefix\endcsname\relax\def\urlprefix{URL }\fi
\expandafter\ifx\csname doiprefix\endcsname\relax\def\doiprefix{DOI: }\fi
\providecommand{\bibinfo}[2]{#2}
\providecommand{\eprint}[2][]{\url{#2}}

\bibitem{ge2023few}
\bibinfo{author}{Ge, Y.}, \bibinfo{author}{Guo, Y.}, \bibinfo{author}{Das, S.}, \bibinfo{author}{Al-Garadi, M.~A.} \& \bibinfo{author}{Sarker, A.}
\newblock \bibinfo{journal}{\bibinfo{title}{Few-shot learning for medical text: A review of advances, trends, and opportunities}}.
\newblock {\emph{\JournalTitle{Journal of Biomedical Informatics}}} \bibinfo{pages}{104458} (\bibinfo{year}{2023}).

\bibitem{labrak-etal-2024-zero}
\bibinfo{author}{Labrak, Y.}, \bibinfo{author}{Rouvier, M.} \& \bibinfo{author}{Dufour, R.}
\newblock \bibinfo{title}{A zero-shot and few-shot study of instruction-finetuned large language models applied to clinical and biomedical tasks}.
\newblock In \bibinfo{editor}{Calzolari, N.} \emph{et~al.} (eds.) \emph{\bibinfo{booktitle}{Proceedings of the 2024 Joint International Conference on Computational Linguistics, Language Resources and Evaluation (LREC-COLING 2024)}}, \bibinfo{pages}{2049--2066} (\bibinfo{publisher}{ELRA and ICCL}, \bibinfo{address}{Torino, Italia}, \bibinfo{year}{2024}).

\bibitem{brown2020language}
\bibinfo{author}{Brown, T.} \emph{et~al.}
\newblock \bibinfo{journal}{\bibinfo{title}{Language models are few-shot learners}}.
\newblock {\emph{\JournalTitle{Advances in neural information processing systems}}} \textbf{\bibinfo{volume}{33}}, \bibinfo{pages}{1877--1901} (\bibinfo{year}{2020}).

\bibitem{zaghir2024prompt}
\bibinfo{author}{Zaghir, J.} \emph{et~al.}
\newblock \bibinfo{journal}{\bibinfo{title}{Prompt engineering paradigms for medical applications: Scoping review}}.
\newblock {\emph{\JournalTitle{Journal of Medical Internet Research}}} \textbf{\bibinfo{volume}{26}}, \bibinfo{pages}{e60501} (\bibinfo{year}{2024}).

\bibitem{yeh-etal-2022-decorate}
\bibinfo{author}{Yeh, H.-S.}, \bibinfo{author}{Lavergne, T.} \& \bibinfo{author}{Zweigenbaum, P.}
\newblock \bibinfo{title}{Decorate the examples: A simple method of prompt design for biomedical relation extraction}.
\newblock In \bibinfo{editor}{Calzolari, N.} \emph{et~al.} (eds.) \emph{\bibinfo{booktitle}{Proceedings of the Thirteenth Language Resources and Evaluation Conference}}, \bibinfo{pages}{3780--3787} (\bibinfo{publisher}{European Language Resources Association}, \bibinfo{address}{Marseille, France}, \bibinfo{year}{2022}).

\bibitem{ye2023prompt}
\bibinfo{author}{Ye, Q.}, \bibinfo{author}{Axmed, M.}, \bibinfo{author}{Pryzant, R.} \& \bibinfo{author}{Khani, F.}
\newblock \bibinfo{journal}{\bibinfo{title}{Prompt engineering a prompt engineer}}.
\newblock {\emph{\JournalTitle{arXiv preprint arXiv:2311.05661}}}  (\bibinfo{year}{2023}).

\bibitem{chang-jia-2023-data}
\bibinfo{author}{Chang, T.-Y.} \& \bibinfo{author}{Jia, R.}
\newblock \bibinfo{title}{Data curation alone can stabilize in-context learning}.
\newblock In \bibinfo{editor}{Rogers, A.}, \bibinfo{editor}{Boyd-Graber, J.} \& \bibinfo{editor}{Okazaki, N.} (eds.) \emph{\bibinfo{booktitle}{Proceedings of the 61st Annual Meeting of the Association for Computational Linguistics (Volume 1: Long Papers)}}, \bibinfo{pages}{8123--8144}, \doiprefix\url{10.18653/v1/2023.acl-long.452} (\bibinfo{publisher}{Association for Computational Linguistics}, \bibinfo{address}{Toronto, Canada}, \bibinfo{year}{2023}).

\bibitem{li2024rt}
\bibinfo{author}{Li, M.}, \bibinfo{author}{Zhou, H.}, \bibinfo{author}{Yang, H.} \& \bibinfo{author}{Zhang, R.}
\newblock \bibinfo{journal}{\bibinfo{title}{Rt: a retrieving and chain-of-thought framework for few-shot medical named entity recognition}}.
\newblock {\emph{\JournalTitle{Journal of the American Medical Informatics Association}}} \bibinfo{pages}{ocae095} (\bibinfo{year}{2024}).

\bibitem{wei2022chain}
\bibinfo{author}{Wei, J.} \emph{et~al.}
\newblock \bibinfo{journal}{\bibinfo{title}{Chain-of-thought prompting elicits reasoning in large language models}}.
\newblock {\emph{\JournalTitle{Advances in neural information processing systems}}} \textbf{\bibinfo{volume}{35}}, \bibinfo{pages}{24824--24837} (\bibinfo{year}{2022}).

\bibitem{lewis2020retrieval}
\bibinfo{author}{Lewis, P.} \emph{et~al.}
\newblock \bibinfo{journal}{\bibinfo{title}{Retrieval-augmented generation for knowledge-intensive nlp tasks}}.
\newblock {\emph{\JournalTitle{Advances in Neural Information Processing Systems}}} \textbf{\bibinfo{volume}{33}}, \bibinfo{pages}{9459--9474} (\bibinfo{year}{2020}).

\bibitem{liao1998similarity}
\bibinfo{author}{Liao, T.~W.}, \bibinfo{author}{Zhang, Z.} \& \bibinfo{author}{Mount, C.~R.}
\newblock \bibinfo{journal}{\bibinfo{title}{Similarity measures for retrieval in case-based reasoning systems}}.
\newblock {\emph{\JournalTitle{Applied Artificial Intelligence}}} \textbf{\bibinfo{volume}{12}}, \bibinfo{pages}{267--288} (\bibinfo{year}{1998}).

\bibitem{gao2023retrieval}
\bibinfo{author}{Gao, Y.} \emph{et~al.}
\newblock \bibinfo{journal}{\bibinfo{title}{Retrieval-augmented generation for large language models: A survey}}.
\newblock {\emph{\JournalTitle{arXiv preprint arXiv:2312.10997}}}  (\bibinfo{year}{2023}).

\bibitem{jin2024genegpt}
\bibinfo{author}{Jin, Q.}, \bibinfo{author}{Yang, Y.}, \bibinfo{author}{Chen, Q.} \& \bibinfo{author}{Lu, Z.}
\newblock \bibinfo{journal}{\bibinfo{title}{Genegpt: Augmenting large language models with domain tools for improved access to biomedical information}}.
\newblock {\emph{\JournalTitle{Bioinformatics}}} \textbf{\bibinfo{volume}{40}}, \bibinfo{pages}{btae075} (\bibinfo{year}{2024}).

\bibitem{xiong2024benchmarking}
\bibinfo{author}{Xiong, G.}, \bibinfo{author}{Jin, Q.}, \bibinfo{author}{Lu, Z.} \& \bibinfo{author}{Zhang, A.}
\newblock \bibinfo{journal}{\bibinfo{title}{Benchmarking retrieval-augmented generation for medicine}}.
\newblock {\emph{\JournalTitle{arXiv preprint arXiv:2402.13178}}}  (\bibinfo{year}{2024}).

\bibitem{jeong2024adaptive}
\bibinfo{author}{Jeong, S.}, \bibinfo{author}{Baek, J.}, \bibinfo{author}{Cho, S.}, \bibinfo{author}{Hwang, S.~J.} \& \bibinfo{author}{Park, J.~C.}
\newblock \bibinfo{journal}{\bibinfo{title}{Adaptive-rag: Learning to adapt retrieval-augmented large language models through question complexity}}.
\newblock {\emph{\JournalTitle{arXiv preprint arXiv:2403.14403}}}  (\bibinfo{year}{2024}).

\bibitem{sahoo2024systematic}
\bibinfo{author}{Sahoo, P.} \emph{et~al.}
\newblock \bibinfo{journal}{\bibinfo{title}{A systematic survey of prompt engineering in large language models: Techniques and applications}}.
\newblock {\emph{\JournalTitle{arXiv preprint arXiv:2402.07927}}}  (\bibinfo{year}{2024}).

\bibitem{levy2024same}
\bibinfo{author}{Levy, M.}, \bibinfo{author}{Jacoby, A.} \& \bibinfo{author}{Goldberg, Y.}
\newblock \bibinfo{journal}{\bibinfo{title}{Same task, more tokens: the impact of input length on the reasoning performance of large language models}}.
\newblock {\emph{\JournalTitle{arXiv preprint arXiv:2402.14848}}}  (\bibinfo{year}{2024}).

\bibitem{xie2023empirical}
\bibinfo{author}{Xie, T.} \emph{et~al.}
\newblock \bibinfo{title}{Empirical study of zero-shot {NER} with {C}hat{GPT}}.
\newblock In \bibinfo{editor}{Bouamor, H.}, \bibinfo{editor}{Pino, J.} \& \bibinfo{editor}{Bali, K.} (eds.) \emph{\bibinfo{booktitle}{Proceedings of the 2023 Conference on Empirical Methods in Natural Language Processing}}, \bibinfo{pages}{7935--7956}, \doiprefix\url{10.18653/v1/2023.emnlp-main.493} (\bibinfo{publisher}{Association for Computational Linguistics}, \bibinfo{address}{Singapore}, \bibinfo{year}{2023}).

\bibitem{ocad259}
\bibinfo{author}{Hu, Y.} \emph{et~al.}
\newblock \bibinfo{journal}{\bibinfo{title}{{Improving large language models for clinical named entity recognition via prompt engineering}}}.
\newblock {\emph{\JournalTitle{Journal of the American Medical Informatics Association}}} \bibinfo{pages}{ocad259}, \doiprefix\url{10.1093/jamia/ocad259} (\bibinfo{year}{2024}).
\newblock \eprint{https://academic.oup.com/jamia/advance-article-pdf/doi/10.1093/jamia/ocad259/56437671/ocad259.pdf}.

\bibitem{sparck1972statistical}
\bibinfo{author}{Sparck~Jones, K.}
\newblock \bibinfo{journal}{\bibinfo{title}{A statistical interpretation of term specificity and its application in retrieval}}.
\newblock {\emph{\JournalTitle{Journal of documentation}}} \textbf{\bibinfo{volume}{28}}, \bibinfo{pages}{11--21} (\bibinfo{year}{1972}).

\bibitem{reimers2019sentence}
\bibinfo{author}{Reimers, N.}
\newblock \bibinfo{journal}{\bibinfo{title}{Sentence-bert: Sentence embeddings using siamese bert-networks}}.
\newblock {\emph{\JournalTitle{arXiv preprint arXiv:1908.10084}}}  (\bibinfo{year}{2019}).

\bibitem{khattab2020colbert}
\bibinfo{author}{Khattab, O.} \& \bibinfo{author}{Zaharia, M.}
\newblock \bibinfo{title}{Colbert: Efficient and effective passage search via contextualized late interaction over bert}.
\newblock In \emph{\bibinfo{booktitle}{Proceedings of the 43rd International ACM SIGIR conference on research and development in Information Retrieval}}, \bibinfo{pages}{39--48} (\bibinfo{year}{2020}).

\bibitem{karpukhin2020dense}
\bibinfo{author}{Karpukhin, V.} \emph{et~al.}
\newblock \bibinfo{journal}{\bibinfo{title}{Dense passage retrieval for open-domain question answering}}.
\newblock {\emph{\JournalTitle{arXiv preprint arXiv:2004.04906}}}  (\bibinfo{year}{2020}).

\bibitem{morey2008confidence}
\bibinfo{author}{Morey, R.~D.} \emph{et~al.}
\newblock \bibinfo{journal}{\bibinfo{title}{Confidence intervals from normalized data: A correction to cousineau (2005)}}.
\newblock {\emph{\JournalTitle{Tutorials in quantitative methods for psychology}}} \textbf{\bibinfo{volume}{4}}, \bibinfo{pages}{61--64} (\bibinfo{year}{2008}).

\bibitem{efron1992bootstrap}
\bibinfo{author}{Efron, B.}
\newblock \bibinfo{title}{Bootstrap methods: another look at the jackknife}.
\newblock In \emph{\bibinfo{booktitle}{Breakthroughs in statistics: Methodology and distribution}}, \bibinfo{pages}{569--593} (\bibinfo{publisher}{Springer}, \bibinfo{year}{1992}).

\bibitem{johnson2016}
\bibinfo{author}{Johnson, A.~E.} \emph{et~al.}
\newblock \bibinfo{journal}{\bibinfo{title}{Mimic-iii, a freely accessible critical care database}}.
\newblock {\emph{\JournalTitle{Scientific data}}} \textbf{\bibinfo{volume}{3}}, \bibinfo{pages}{1--9}, \doiprefix\url{10.1038/sdata.2016.35} (\bibinfo{year}{2016}).

\bibitem{li2016biocreative}
\bibinfo{author}{Li, J.} \emph{et~al.}
\newblock \bibinfo{journal}{\bibinfo{title}{{BioCreative V CDR task corpus: a resource for chemical disease relation extraction}}}.
\newblock {\emph{\JournalTitle{Database}}} \textbf{\bibinfo{volume}{2016}}, \bibinfo{pages}{baw068}, \doiprefix\url{10.1093/database/baw068} (\bibinfo{year}{2016}).
\newblock \eprint{https://academic.oup.com/database/article-pdf/doi/10.1093/database/baw068/8224483/baw068.pdf}.

\bibitem{mohan2019medmentions}
\bibinfo{author}{Mohan, S.} \& \bibinfo{author}{Li, D.}
\newblock \bibinfo{title}{Medmentions: A large biomedical corpus annotated with {\{}umls{\}} concepts}.
\newblock In \emph{\bibinfo{booktitle}{Automated Knowledge Base Construction (AKBC)}} (\bibinfo{year}{2019}).

\bibitem{dougan2014ncbi}
\bibinfo{author}{Doğan, R.~I.}, \bibinfo{author}{Leaman, R.} \& \bibinfo{author}{Lu, Z.}
\newblock \bibinfo{journal}{\bibinfo{title}{Ncbi disease corpus: A resource for disease name recognition and concept normalization}}.
\newblock {\emph{\JournalTitle{Journal of Biomedical Informatics}}} \textbf{\bibinfo{volume}{47}}, \bibinfo{pages}{1--10}, \doiprefix\url{https://doi.org/10.1016/j.jbi.2013.12.006} (\bibinfo{year}{2014}).

\bibitem{ge2024reddit}
\bibinfo{author}{Ge, Y.} \emph{et~al.}
\newblock \bibinfo{journal}{\bibinfo{title}{Reddit-impacts: A named entity recognition dataset for analyzing clinical and social effects of substance use derived from social media}}.
\newblock {\emph{\JournalTitle{arXiv preprint arXiv:2405.06145}}}  (\bibinfo{year}{2024}).

\end{thebibliography}

\newpage
\begin{appendices}
\clearpage
\section{Averaged Performance of the Baseline Dynamic Prompt Model}\label{app:A}

\begin{table}[h]
\centering
\footnotesize
\renewcommand{\arraystretch}{1.2}
\resizebox{\columnwidth}{!}{%
\begin{tabular}{llrrrllrrr}
\hline
 &
   &
  \multicolumn{1}{r}{\textbf{Precision}} &
  \multicolumn{1}{r}{\textbf{Recall}} &
  \multicolumn{1}{r}{\textbf{F$_1$-score}} &
   &
   &
  \multicolumn{1}{r}{\textbf{Precision}} &
  \multicolumn{1}{r}{\textbf{Recall}} &
  \multicolumn{1}{r}{\textbf{F$_1$-score}} \\ \hline
\multicolumn{1}{|l|}{} &
  \multicolumn{1}{l|}{} &
  16.40 &
  50.00 &
  \multicolumn{1}{r|}{24.70} &
  \multicolumn{1}{l|}{} &
  \multicolumn{1}{l|}{} &
  12.56 &
  56.32 &
  20.55 \\ \cline{3-5} \cline{8-10} 
\multicolumn{1}{|l|}{} &
  \multicolumn{1}{l|}{} &
  17.67 &
  50.62 &
  \multicolumn{1}{r|}{26.20} &
  \multicolumn{1}{l|}{} &
  \multicolumn{1}{l|}{} &
  13.68 &
  55.81 &
  21.97 \\ \cline{3-5} \cline{8-10} 
\multicolumn{1}{|l|}{} &
  \multicolumn{1}{l|}{} &
  17.48 &
  53.09 &
  \multicolumn{1}{r|}{26.30} &
  \multicolumn{1}{l|}{} &
  \multicolumn{1}{l|}{} &
  12.14 &
  59.52 &
  20.16 \\ \cline{3-5} \cline{8-10} 
\multicolumn{1}{|l|}{} &
  \multicolumn{1}{l|}{\multirow{-4}{*}{\textbf{5-shot}}} &
  23.91 &
  54.32 &
  \multicolumn{1}{r|}{33.20} &
  \multicolumn{1}{l|}{} &
  \multicolumn{1}{l|}{\multirow{-4}{*}{\textbf{5-shot}}} &
  14.25 &
  59.79 &
  23.02 \\ \cline{2-5} \cline{7-10} 
\multicolumn{1}{|l|}{} &
  \multicolumn{1}{l|}{\textbf{AVG}} &
  \cellcolor[HTML]{ECF4FF}18.865 &
  \cellcolor[HTML]{ECF4FF}52.0075 &
  \multicolumn{1}{r|}{\cellcolor[HTML]{ECF4FF}27.60} &
  \multicolumn{1}{l|}{} &
  \multicolumn{1}{l|}{\textbf{AVG}} &
  \cellcolor[HTML]{ECF4FF}13.1575 &
  \cellcolor[HTML]{ECF4FF}57.86 &
  \cellcolor[HTML]{ECF4FF}21.425 \\ \cline{2-5} \cline{7-10} 
\multicolumn{1}{|l|}{} &
  \multicolumn{1}{l|}{} &
  25.41 &
  58.75 &
  \multicolumn{1}{r|}{35.47} &
  \multicolumn{1}{l|}{} &
  \multicolumn{1}{l|}{} &
  24.51 &
  59.52 &
  34.72 \\ \cline{3-5} \cline{8-10} 
\multicolumn{1}{|l|}{} &
  \multicolumn{1}{l|}{} &
  21.67 &
  54.32 &
  \multicolumn{1}{r|}{30.99} &
  \multicolumn{1}{l|}{} &
  \multicolumn{1}{l|}{} &
  19.63 &
  62.35 &
  29.86 \\ \cline{3-5} \cline{8-10} 
\multicolumn{1}{|l|}{} &
  \multicolumn{1}{l|}{} &
  21.43 &
  59.26 &
  \multicolumn{1}{r|}{31.48} &
  \multicolumn{1}{l|}{} &
  \multicolumn{1}{l|}{} &
  23.96 &
  57.44 &
  33.81 \\ \cline{3-5} \cline{8-10} 
\multicolumn{1}{|l|}{} &
  \multicolumn{1}{l|}{\multirow{-4}{*}{\textbf{10-shot}}} &
  20.47 &
  54.32 &
  \multicolumn{1}{r|}{29.73} &
  \multicolumn{1}{l|}{} &
  \multicolumn{1}{l|}{\multirow{-4}{*}{\textbf{10-shot}}} &
  21.38 &
  60.43 &
  31.59 \\ \cline{2-5} \cline{7-10} 
\multicolumn{1}{|l|}{} &
  \multicolumn{1}{l|}{\textbf{AVG}} &
  \cellcolor[HTML]{ECF4FF}22.245 &
  \cellcolor[HTML]{ECF4FF}56.6625 &
  \multicolumn{1}{r|}{\cellcolor[HTML]{ECF4FF}31.9175} &
  \multicolumn{1}{l|}{} &
  \multicolumn{1}{l|}{\textbf{AVG}} &
  \cellcolor[HTML]{ECF4FF}22.37 &
  \cellcolor[HTML]{ECF4FF}59.935 &
  \cellcolor[HTML]{ECF4FF}32.495 \\ \cline{2-5} \cline{7-10} 
\multicolumn{1}{|l|}{} &
  \multicolumn{1}{l|}{} &
  28.74 &
  58.63 &
  \multicolumn{1}{r|}{38.57} &
  \multicolumn{1}{l|}{} &
  \multicolumn{1}{l|}{} &
  27.22 &
  57.65 &
  36.98 \\ \cline{3-5} \cline{8-10} 
\multicolumn{1}{|l|}{} &
  \multicolumn{1}{l|}{} &
  27.03 &
  57.54 &
  \multicolumn{1}{r|}{36.78} &
  \multicolumn{1}{l|}{} &
  \multicolumn{1}{l|}{} &
  23.24 &
  52.44 &
  32.21 \\ \cline{3-5} \cline{8-10} 
\multicolumn{1}{|l|}{} &
  \multicolumn{1}{l|}{} &
  26.52 &
  59.79 &
  \multicolumn{1}{r|}{36.74} &
  \multicolumn{1}{l|}{} &
  \multicolumn{1}{l|}{} &
  22.12 &
  52.08 &
  31.06 \\ \cline{3-5} \cline{8-10} 
\multicolumn{1}{|l|}{} &
  \multicolumn{1}{l|}{\multirow{-4}{*}{\textbf{20-shot}}} &
  28.65 &
  59.02 &
  \multicolumn{1}{r|}{38.57} &
  \multicolumn{1}{l|}{} &
  \multicolumn{1}{l|}{\multirow{-4}{*}{\textbf{20-shot}}} &
  25.49 &
  53.06 &
  34.44 \\ \cline{2-5} \cline{7-10} 
\multicolumn{1}{|l|}{\multirow{-15}{*}{\textbf{GPT 4}}} &
  \multicolumn{1}{l|}{\textbf{AVG}} &
  \cellcolor[HTML]{ECF4FF}27.735 &
  \cellcolor[HTML]{ECF4FF}58.745 &
  \multicolumn{1}{r|}{\cellcolor[HTML]{ECF4FF}37.665} &
  \multicolumn{1}{l|}{\multirow{-15}{*}{\textbf{Llama3}}} &
  \multicolumn{1}{l|}{\textbf{AVG}} &
  \cellcolor[HTML]{ECF4FF}24.5175 &
  \cellcolor[HTML]{ECF4FF}53.8075 &
  \cellcolor[HTML]{ECF4FF}33.6725 \\ \hline
\end{tabular}%
}
\caption{Averaged performance of the baseline dynamic prompt model on the \textsc{Reddit-Impacts} dataset across different shot settings.}
\label{tab:reddit_impacts_avg}
\end{table}
\begin{table}[h]
\centering
\footnotesize
\renewcommand{\arraystretch}{1.2}
\resizebox{\columnwidth}{!}{%
\begin{tabular}{llrrrllrrr}
\hline
 &
   &
  \multicolumn{1}{r}{\textbf{Precision}} &
  \multicolumn{1}{r}{\textbf{Recall}} &
  \multicolumn{1}{r}{\textbf{F$_1$-score}} &
   &
   &
  \multicolumn{1}{r}{\textbf{Precision}} &
  \multicolumn{1}{r}{\textbf{Recall}} &
  \multicolumn{1}{r}{\textbf{F$_1$-score}} \\ \hline
\multicolumn{1}{|l|}{} &
  \multicolumn{1}{l|}{} &
  69.97 &
  91.43 &
  \multicolumn{1}{r|}{79.27} &
  \multicolumn{1}{l|}{} &
  \multicolumn{1}{l|}{} &
  68.09 &
  84.35 &
  75.35 \\ \cline{3-5} \cline{8-10} 
\multicolumn{1}{|l|}{} &
  \multicolumn{1}{l|}{} &
  68.32 &
  88.3 &
  \multicolumn{1}{r|}{77.29} &
  \multicolumn{1}{l|}{} &
  \multicolumn{1}{l|}{} &
  69.70 &
  76.67 &
  73.02 \\ \cline{3-5} \cline{8-10} 
\multicolumn{1}{|l|}{} &
  \multicolumn{1}{l|}{} &
  64.58 &
  90.15 &
  \multicolumn{1}{r|}{75.25} &
  \multicolumn{1}{l|}{} &
  \multicolumn{1}{l|}{} &
  71.00 &
  79.15 &
  74.86 \\ \cline{3-5} \cline{8-10} 
\multicolumn{1}{|l|}{} &
  \multicolumn{1}{l|}{\multirow{-4}{*}{\textbf{5-shot}}} &
  71.59 &
  91.39 &
  \multicolumn{1}{r|}{80.29} &
  \multicolumn{1}{l|}{} &
  \multicolumn{1}{l|}{\multirow{-4}{*}{\textbf{5-shot}}} &
  67.10 &
  73.25 &
  70.04 \\ \cline{2-5} \cline{7-10} 
\multicolumn{1}{|l|}{} &
  \multicolumn{1}{l|}{\textbf{AVG}} &
  \cellcolor[HTML]{ECF4FF}68.615 &
  \cellcolor[HTML]{ECF4FF}90.3175 &
  \multicolumn{1}{r|}{\cellcolor[HTML]{ECF4FF}78.025} &
  \multicolumn{1}{l|}{} &
  \multicolumn{1}{l|}{\textbf{AVG}} &
  \cellcolor[HTML]{ECF4FF}68.9725 &
  \cellcolor[HTML]{ECF4FF}78.355 &
  \cellcolor[HTML]{ECF4FF}73.3175 \\ \cline{2-5} \cline{7-10} 
\multicolumn{1}{|l|}{} &
  \multicolumn{1}{l|}{} &
  75.11 &
  90.84 &
  \multicolumn{1}{r|}{82.23} &
  \multicolumn{1}{l|}{} &
  \multicolumn{1}{l|}{} &
  72.38 &
  79.18 &
  75.63 \\ \cline{3-5} \cline{8-10} 
\multicolumn{1}{|l|}{} &
  \multicolumn{1}{l|}{} &
  74.41 &
  90.32 &
  \multicolumn{1}{r|}{81.60} &
  \multicolumn{1}{l|}{} &
  \multicolumn{1}{l|}{} &
  73.31 &
  74.24 &
  73.77 \\ \cline{3-5} \cline{8-10} 
\multicolumn{1}{|l|}{} &
  \multicolumn{1}{l|}{} &
  74.13 &
  85.71 &
  \multicolumn{1}{r|}{79.50} &
  \multicolumn{1}{l|}{} &
  \multicolumn{1}{l|}{} &
  71.15 &
  77.05 &
  73.98 \\ \cline{3-5} \cline{8-10} 
\multicolumn{1}{|l|}{} &
  \multicolumn{1}{l|}{\multirow{-4}{*}{\textbf{10-shot}}} &
  77.65 &
  86.35 &
  \multicolumn{1}{r|}{81.73} &
  \multicolumn{1}{l|}{} &
  \multicolumn{1}{l|}{\multirow{-4}{*}{\textbf{10-shot}}} &
  73.40 &
  81.17 &
  77.2 \\ \cline{2-5} \cline{7-10} 
\multicolumn{1}{|l|}{} &
  \multicolumn{1}{l|}{\textbf{AVG}} &
  \cellcolor[HTML]{ECF4FF}75.325 &
  \cellcolor[HTML]{ECF4FF}88.305 &
  \multicolumn{1}{r|}{\cellcolor[HTML]{ECF4FF}81.265} &
  \multicolumn{1}{l|}{} &
  \multicolumn{1}{l|}{\textbf{AVG}} &
  \cellcolor[HTML]{ECF4FF}72.56 &
  \cellcolor[HTML]{ECF4FF}77.91 &
  \cellcolor[HTML]{ECF4FF}75.145 \\ \cline{2-5} \cline{7-10} 
\multicolumn{1}{|l|}{} &
  \multicolumn{1}{l|}{} &
  75.36 &
  88.33 &
  \multicolumn{1}{r|}{81.33} &
  \multicolumn{1}{l|}{} &
  \multicolumn{1}{l|}{} &
  75.12 &
  74.84 &
  74.98 \\ \cline{3-5} \cline{8-10} 
\multicolumn{1}{|l|}{} &
  \multicolumn{1}{l|}{} &
  73.13 &
  91.74 &
  \multicolumn{1}{r|}{81.38} &
  \multicolumn{1}{l|}{} &
  \multicolumn{1}{l|}{} &
  72.03 &
  71.88 &
  71.96 \\ \cline{3-5} \cline{8-10} 
\multicolumn{1}{|l|}{} &
  \multicolumn{1}{l|}{} &
  71.84 &
  91.1 &
  \multicolumn{1}{r|}{80.33} &
  \multicolumn{1}{l|}{} &
  \multicolumn{1}{l|}{} &
  76.88 &
  77.20 &
  77.04 \\ \cline{3-5} \cline{8-10} 
\multicolumn{1}{|l|}{} &
  \multicolumn{1}{l|}{\multirow{-4}{*}{\textbf{20-shot}}} &
  77.94 &
  85.54 &
  \multicolumn{1}{r|}{81.57} &
  \multicolumn{1}{l|}{} &
  \multicolumn{1}{l|}{\multirow{-4}{*}{\textbf{20-shot}}} &
  77.64 &
  78.40 &
  78.02 \\ \cline{2-5} \cline{7-10} 
\multicolumn{1}{|l|}{\multirow{-15}{*}{\textbf{GPT 4}}} &
  \multicolumn{1}{l|}{\textbf{AVG}} &
  \cellcolor[HTML]{ECF4FF}74.5675 &
  \cellcolor[HTML]{ECF4FF}89.1775 &
  \multicolumn{1}{r|}{\cellcolor[HTML]{ECF4FF}81.1525} &
  \multicolumn{1}{l|}{\multirow{-15}{*}{\textbf{Llama3}}} &
  \multicolumn{1}{l|}{\textbf{AVG}} &
  \cellcolor[HTML]{ECF4FF}75.4175 &
  \cellcolor[HTML]{ECF4FF}75.58 &
  \cellcolor[HTML]{ECF4FF}75.5 \\ \hline
\end{tabular}%
}
\caption{Averaged performance of the baseline dynamic prompt model on the BC5CDR dataset across different shot settings.}
\label{tab:bc5cdr_avg}
\end{table}
\begin{table}[h]
\centering
\footnotesize
\renewcommand{\arraystretch}{1.2}
\resizebox{\columnwidth}{!}{%
\begin{tabular}{llrrrllrrr}
\hline
 &
   &
  \multicolumn{1}{r}{\textbf{Precision}} &
  \multicolumn{1}{r}{\textbf{Recall}} &
  \multicolumn{1}{r}{\textbf{F$_1$-score}} &
   &
   &
  \multicolumn{1}{r}{\textbf{Precision}} &
  \multicolumn{1}{r}{\textbf{Recall}} &
  \multicolumn{1}{r}{\textbf{F$_1$-score}} \\ \hline
\multicolumn{1}{|l|}{} &
  \multicolumn{1}{l|}{} &
  62.38 &
  63.78 &
  \multicolumn{1}{r|}{63.07} &
  \multicolumn{1}{l|}{} &
  \multicolumn{1}{l|}{} &
  57.69 &
  68.39 &
  62.58 \\ \cline{3-5} \cline{8-10} 
\multicolumn{1}{|l|}{} &
  \multicolumn{1}{l|}{} &
  61.88 &
  62.19 &
  \multicolumn{1}{r|}{62.03} &
  \multicolumn{1}{l|}{} &
  \multicolumn{1}{l|}{} &
  61.83 &
  61.33 &
  61.58 \\ \cline{3-5} \cline{8-10} 
\multicolumn{1}{|l|}{} &
  \multicolumn{1}{l|}{} &
  65.26 &
  68.14 &
  \multicolumn{1}{r|}{66.67} &
  \multicolumn{1}{l|}{} &
  \multicolumn{1}{l|}{} &
  58.24 &
  68.06 &
  62.77 \\ \cline{3-5} \cline{8-10} 
\multicolumn{1}{|l|}{} &
  \multicolumn{1}{l|}{\multirow{-4}{*}{\textbf{5-shot}}} &
  62.71 &
  62.36 &
  \multicolumn{1}{r|}{62.54} &
  \multicolumn{1}{l|}{} &
  \multicolumn{1}{l|}{\multirow{-4}{*}{\textbf{5-shot}}} &
  59.44 &
  71.29 &
  64.82 \\ \cline{2-5} \cline{7-10} 
\multicolumn{1}{|l|}{} &
  \multicolumn{1}{l|}{\textbf{AVG}} &
  \cellcolor[HTML]{ECF4FF}63.0575 &
  \cellcolor[HTML]{ECF4FF}64.1175 &
  \multicolumn{1}{r|}{\cellcolor[HTML]{ECF4FF}63.5775} &
  \multicolumn{1}{l|}{} &
  \multicolumn{1}{l|}{\textbf{AVG}} &
  \cellcolor[HTML]{ECF4FF}59.3 &
  \cellcolor[HTML]{ECF4FF}67.2675 &
  \cellcolor[HTML]{ECF4FF}62.9375 \\ \cline{2-5} \cline{7-10} 
\multicolumn{1}{|l|}{} &
  \multicolumn{1}{l|}{} &
  66.38 &
  74.24 &
  \multicolumn{1}{r|}{70.09} &
  \multicolumn{1}{l|}{} &
  \multicolumn{1}{l|}{} &
  59.13 &
  71.63 &
  64.78 \\ \cline{3-5} \cline{8-10} 
\multicolumn{1}{|l|}{} &
  \multicolumn{1}{l|}{} &
  68.37 &
  74.86 &
  \multicolumn{1}{r|}{71.47} &
  \multicolumn{1}{l|}{} &
  \multicolumn{1}{l|}{} &
  62.68 &
  63.8 &
  63.23 \\ \cline{3-5} \cline{8-10} 
\multicolumn{1}{|l|}{} &
  \multicolumn{1}{l|}{} &
  66.72 &
  75.33 &
  \multicolumn{1}{r|}{70.77} &
  \multicolumn{1}{l|}{} &
  \multicolumn{1}{l|}{} &
  61.71 &
  67.58 &
  64.51 \\ \cline{3-5} \cline{8-10} 
\multicolumn{1}{|l|}{} &
  \multicolumn{1}{l|}{\multirow{-4}{*}{\textbf{10-shot}}} &
  66.46 &
  73.34 &
  \multicolumn{1}{r|}{69.73} &
  \multicolumn{1}{l|}{} &
  \multicolumn{1}{l|}{\multirow{-4}{*}{\textbf{10-shot}}} &
  62.24 &
  62.84 &
  62.54 \\ \cline{2-5} \cline{7-10} 
\multicolumn{1}{|l|}{} &
  \multicolumn{1}{l|}{\textbf{AVG}} &
  \cellcolor[HTML]{ECF4FF}66.9825 &
  \cellcolor[HTML]{ECF4FF}74.4425 &
  \multicolumn{1}{r|}{\cellcolor[HTML]{ECF4FF}70.515} &
  \multicolumn{1}{l|}{} &
  \multicolumn{1}{l|}{\textbf{AVG}} &
  \cellcolor[HTML]{ECF4FF}61.44 &
  \cellcolor[HTML]{ECF4FF}66.4625 &
  \cellcolor[HTML]{ECF4FF}63.765 \\ \cline{2-5} \cline{7-10} 
\multicolumn{1}{|l|}{} &
  \multicolumn{1}{l|}{} &
  70.41 &
  69.84 &
  \multicolumn{1}{r|}{70.12} &
  \multicolumn{1}{l|}{} &
  \multicolumn{1}{l|}{} &
  63.89 &
  62.04 &
  62.95 \\ \cline{3-5} \cline{8-10} 
\multicolumn{1}{|l|}{} &
  \multicolumn{1}{l|}{} &
  70.11 &
  71.24 &
  \multicolumn{1}{r|}{70.67} &
  \multicolumn{1}{l|}{} &
  \multicolumn{1}{l|}{} &
  63.96 &
  62.83 &
  63.39 \\ \cline{3-5} \cline{8-10} 
\multicolumn{1}{|l|}{} &
  \multicolumn{1}{l|}{} &
  71.45 &
  73.39 &
  \multicolumn{1}{r|}{72.41} &
  \multicolumn{1}{l|}{} &
  \multicolumn{1}{l|}{} &
  59.22 &
  61.71 &
  60.44 \\ \cline{3-5} \cline{8-10} 
\multicolumn{1}{|l|}{} &
  \multicolumn{1}{l|}{\multirow{-4}{*}{\textbf{20-shot}}} &
  70.64 &
  70.80 &
  \multicolumn{1}{r|}{70.72} &
  \multicolumn{1}{l|}{} &
  \multicolumn{1}{l|}{\multirow{-4}{*}{\textbf{20-shot}}} &
  60.96 &
  61.88 &
  61.42 \\ \cline{2-5} \cline{7-10} 
\multicolumn{1}{|l|}{\multirow{-15}{*}{\textbf{GPT 4}}} &
  \multicolumn{1}{l|}{\textbf{AVG}} &
  \cellcolor[HTML]{ECF4FF}70.6525 &
  \cellcolor[HTML]{ECF4FF}71.3175 &
  \multicolumn{1}{r|}{\cellcolor[HTML]{ECF4FF}70.98} &
  \multicolumn{1}{l|}{\multirow{-15}{*}{\textbf{Llama3}}} &
  \multicolumn{1}{l|}{\textbf{AVG}} &
  \cellcolor[HTML]{ECF4FF}62.0075 &
  \cellcolor[HTML]{ECF4FF}62.115 &
  \cellcolor[HTML]{ECF4FF}62.05 \\ \hline
\end{tabular}%
}
\caption{Averaged performance of the baseline dynamic prompt model on the MIMIC III dataset across different shot settings.}
\label{tab:mimic_avg}
\end{table}
\begin{table}[h]
\centering
\footnotesize
\renewcommand{\arraystretch}{1.2}
\resizebox{\columnwidth}{!}{%
\begin{tabular}{llrrrllrrr}
\hline
 &
   &
  \multicolumn{1}{r}{\textbf{Precision}} &
  \multicolumn{1}{r}{\textbf{Recall}} &
  \multicolumn{1}{r}{\textbf{F$_1$-score}} &
   &
   &
  \multicolumn{1}{r}{\textbf{Precision}} &
  \multicolumn{1}{r}{\textbf{Recall}} &
  \multicolumn{1}{r}{\textbf{F$_1$-score}} \\ \hline
\multicolumn{1}{|l|}{} &
  \multicolumn{1}{l|}{} &
  46.17 &
  50.52 &
  \multicolumn{1}{r|}{48.25} &
  \multicolumn{1}{l|}{} &
  \multicolumn{1}{l|}{} &
  33.88 &
  40.67 &
  36.97 \\ \cline{3-5} \cline{8-10} 
\multicolumn{1}{|l|}{} &
  \multicolumn{1}{l|}{} &
  45.69 &
  49.07 &
  \multicolumn{1}{r|}{47.32} &
  \multicolumn{1}{l|}{} &
  \multicolumn{1}{l|}{} &
  32.26 &
  39.14 &
  35.37 \\ \cline{3-5} \cline{8-10} 
\multicolumn{1}{|l|}{} &
  \multicolumn{1}{l|}{} &
  40.24 &
  43.65 &
  \multicolumn{1}{r|}{41.88} &
  \multicolumn{1}{l|}{} &
  \multicolumn{1}{l|}{} &
  36.38 &
  28.49 &
  31.95 \\ \cline{3-5} \cline{8-10} 
\multicolumn{1}{|l|}{} &
  \multicolumn{1}{l|}{\multirow{-4}{*}{\textbf{5-shot}}} &
  47.96 &
  52.83 &
  \multicolumn{1}{r|}{50.28} &
  \multicolumn{1}{l|}{} &
  \multicolumn{1}{l|}{\multirow{-4}{*}{\textbf{5-shot}}} &
  40.73 &
  30.53 &
  34.9 \\ \cline{2-5} \cline{7-10} 
\multicolumn{1}{|l|}{} &
  \multicolumn{1}{l|}{\textbf{AVG}} &
  \cellcolor[HTML]{ECF4FF}45.015 &
  \cellcolor[HTML]{ECF4FF}49.0175 &
  \multicolumn{1}{r|}{\cellcolor[HTML]{ECF4FF}46.9325} &
  \multicolumn{1}{l|}{} &
  \multicolumn{1}{l|}{\textbf{AVG}} &
  \cellcolor[HTML]{ECF4FF}35.8125 &
  \cellcolor[HTML]{ECF4FF}34.7075 &
  \cellcolor[HTML]{ECF4FF}34.7975 \\ \cline{2-5} \cline{7-10} 
\multicolumn{1}{|l|}{} &
  \multicolumn{1}{l|}{} &
  52.98 &
  52.67 &
  \multicolumn{1}{r|}{52.82} &
  \multicolumn{1}{l|}{} &
  \multicolumn{1}{l|}{} &
  42.79 &
  31.08 &
  35.66 \\ \cline{3-5} \cline{8-10} 
\multicolumn{1}{|l|}{} &
  \multicolumn{1}{l|}{} &
  53.71 &
  53.36 &
  \multicolumn{1}{r|}{53.54} &
  \multicolumn{1}{l|}{} &
  \multicolumn{1}{l|}{} &
  40.18 &
  28.73 &
  36.27 \\ \cline{3-5} \cline{8-10} 
\multicolumn{1}{|l|}{} &
  \multicolumn{1}{l|}{} &
  52.99 &
  50.35 &
  \multicolumn{1}{r|}{51.64} &
  \multicolumn{1}{l|}{} &
  \multicolumn{1}{l|}{} &
  35.75 &
  32.05 &
  33.76 \\ \cline{3-5} \cline{8-10} 
\multicolumn{1}{|l|}{} &
  \multicolumn{1}{l|}{\multirow{-4}{*}{\textbf{10-shot}}} &
  53.85 &
  52.3 &
  \multicolumn{1}{r|}{53.06} &
  \multicolumn{1}{l|}{} &
  \multicolumn{1}{l|}{\multirow{-4}{*}{\textbf{10-shot}}} &
  39.95 &
  34.11 &
  36.71 \\ \cline{2-5} \cline{7-10} 
\multicolumn{1}{|l|}{} &
  \multicolumn{1}{l|}{\textbf{AVG}} &
  \cellcolor[HTML]{ECF4FF}53.3825 &
  \cellcolor[HTML]{ECF4FF}52.17 &
  \multicolumn{1}{r|}{\cellcolor[HTML]{ECF4FF}52.765} &
  \multicolumn{1}{l|}{} &
  \multicolumn{1}{l|}{\textbf{AVG}} &
  \cellcolor[HTML]{ECF4FF}39.6675 &
  \cellcolor[HTML]{ECF4FF}31.4925 &
  \cellcolor[HTML]{ECF4FF}35.6 \\ \cline{2-5} \cline{7-10} 
\multicolumn{1}{|l|}{} &
  \multicolumn{1}{l|}{} &
  54.57 &
  54.73 &
  \multicolumn{1}{r|}{54.65} &
  \multicolumn{1}{l|}{} &
  \multicolumn{1}{l|}{} &
  40.59 &
  42.07 &
  41.32 \\ \cline{3-5} \cline{8-10} 
\multicolumn{1}{|l|}{} &
  \multicolumn{1}{l|}{} &
  44.86 &
  45.43 &
  \multicolumn{1}{r|}{45.14} &
  \multicolumn{1}{l|}{} &
  \multicolumn{1}{l|}{} &
  40.87 &
  42.8 &
  41.81 \\ \cline{3-5} \cline{8-10} 
\multicolumn{1}{|l|}{} &
  \multicolumn{1}{l|}{} &
  51.6 &
  51.75 &
  \multicolumn{1}{r|}{51.68} &
  \multicolumn{1}{l|}{} &
  \multicolumn{1}{l|}{} &
  41.48 &
  43.02 &
  42.24 \\ \cline{3-5} \cline{8-10} 
\multicolumn{1}{|l|}{} &
  \multicolumn{1}{l|}{\multirow{-4}{*}{\textbf{20-shot}}} &
  55.7 &
  57.24 &
  \multicolumn{1}{r|}{56.46} &
  \multicolumn{1}{l|}{} &
  \multicolumn{1}{l|}{\multirow{-4}{*}{\textbf{20-shot}}} &
  39.88 &
  42.41 &
  41.1 \\ \cline{2-5} \cline{7-10} 
\multicolumn{1}{|l|}{\multirow{-15}{*}{\textbf{GPT 4}}} &
  \multicolumn{1}{l|}{\textbf{AVG}} &
  \cellcolor[HTML]{ECF4FF}51.6825 &
  \cellcolor[HTML]{ECF4FF}52.2875 &
  \multicolumn{1}{r|}{\cellcolor[HTML]{ECF4FF}51.9825} &
  \multicolumn{1}{l|}{\multirow{-15}{*}{\textbf{Llama3}}} &
  \multicolumn{1}{l|}{\textbf{AVG}} &
  \cellcolor[HTML]{ECF4FF}40.705 &
  \cellcolor[HTML]{ECF4FF}42.575 &
  \cellcolor[HTML]{ECF4FF}41.6175 \\ \hline
\end{tabular}%
}
\caption{Averaged performance of the baseline dynamic prompt model on the NCBI dataset across different shot settings.}
\label{tab:ncbi_avg}
\end{table}
\begin{table}[h]
\centering
\footnotesize
\renewcommand{\arraystretch}{1.2}
\resizebox{\columnwidth}{!}{%
\begin{tabular}{llrrrllrrr}
\hline
 &
   &
  \multicolumn{1}{r}{\textbf{Precision}} &
  \multicolumn{1}{r}{\textbf{Recall}} &
  \multicolumn{1}{r}{\textbf{F$_1$-score}} &
   &
   &
  \multicolumn{1}{r}{\textbf{Precision}} &
  \multicolumn{1}{r}{\textbf{Recall}} &
  \multicolumn{1}{r}{\textbf{F$_1$-score}} \\ \hline
\multicolumn{1}{|l|}{} &
  \multicolumn{1}{l|}{} &
  27.24 &
  62.73 &
  \multicolumn{1}{r|}{37.99} &
  \multicolumn{1}{l|}{} &
  \multicolumn{1}{l|}{} &
  24.94 &
  73.03 &
  37.18 \\ \cline{3-5} \cline{8-10} 
\multicolumn{1}{|l|}{} &
  \multicolumn{1}{l|}{} &
  26.71 &
  58.58 &
  \multicolumn{1}{r|}{36.69} &
  \multicolumn{1}{l|}{} &
  \multicolumn{1}{l|}{} &
  25.95 &
  61.96 &
  36.58 \\ \cline{3-5} \cline{8-10} 
\multicolumn{1}{|l|}{} &
  \multicolumn{1}{l|}{} &
  29.23 &
  60.47 &
  \multicolumn{1}{r|}{39.41} &
  \multicolumn{1}{l|}{} &
  \multicolumn{1}{l|}{} &
  26.11 &
  73.57 &
  38.54 \\ \cline{3-5} \cline{8-10} 
\multicolumn{1}{|l|}{} &
  \multicolumn{1}{l|}{\multirow{-4}{*}{\textbf{5-shot}}} &
  25.86 &
  58.41 &
  \multicolumn{1}{r|}{35.85} &
  \multicolumn{1}{l|}{} &
  \multicolumn{1}{l|}{\multirow{-4}{*}{\textbf{5-shot}}} &
  26.55 &
  59.62 &
  36.74 \\ \cline{2-5} \cline{7-10} 
\multicolumn{1}{|l|}{} &
  \multicolumn{1}{l|}{\textbf{AVG}} &
  \cellcolor[HTML]{ECF4FF}27.26 &
  \cellcolor[HTML]{ECF4FF}60.0475 &
  \multicolumn{1}{r|}{\cellcolor[HTML]{ECF4FF}37.485} &
  \multicolumn{1}{l|}{} &
  \multicolumn{1}{l|}{\textbf{AVG}} &
  \cellcolor[HTML]{ECF4FF}25.8875 &
  \cellcolor[HTML]{ECF4FF}67.045 &
  \cellcolor[HTML]{ECF4FF}37.26 \\ \cline{2-5} \cline{7-10} 
\multicolumn{1}{|l|}{} &
  \multicolumn{1}{l|}{} &
  26.92 &
  58.26 &
  \multicolumn{1}{r|}{36.82} &
  \multicolumn{1}{l|}{} &
  \multicolumn{1}{l|}{} &
  25.67 &
  70.61 &
  37.65 \\ \cline{3-5} \cline{8-10} 
\multicolumn{1}{|l|}{} &
  \multicolumn{1}{l|}{} &
  26.1 &
  59.15 &
  \multicolumn{1}{r|}{36.22} &
  \multicolumn{1}{l|}{} &
  \multicolumn{1}{l|}{} &
  25.1 &
  59.19 &
  35.15 \\ \cline{3-5} \cline{8-10} 
\multicolumn{1}{|l|}{} &
  \multicolumn{1}{l|}{} &
  24.41 &
  58.56 &
  \multicolumn{1}{r|}{34.46} &
  \multicolumn{1}{l|}{} &
  \multicolumn{1}{l|}{} &
  25.76 &
  70.06 &
  37.67 \\ \cline{3-5} \cline{8-10} 
\multicolumn{1}{|l|}{} &
  \multicolumn{1}{l|}{\multirow{-4}{*}{\textbf{10-shot}}} &
  29.19 &
  60.82 &
  \multicolumn{1}{r|}{39.45} &
  \multicolumn{1}{l|}{} &
  \multicolumn{1}{l|}{\multirow{-4}{*}{\textbf{10-shot}}} &
  25.73 &
  57.46 &
  35.54 \\ \cline{2-5} \cline{7-10} 
\multicolumn{1}{|l|}{} &
  \multicolumn{1}{l|}{\textbf{AVG}} &
  \cellcolor[HTML]{ECF4FF}26.655 &
  \cellcolor[HTML]{ECF4FF}59.1975 &
  \multicolumn{1}{r|}{\cellcolor[HTML]{ECF4FF}36.7375} &
  \multicolumn{1}{l|}{} &
  \multicolumn{1}{l|}{\textbf{AVG}} &
  \cellcolor[HTML]{ECF4FF}25.565 &
  \cellcolor[HTML]{ECF4FF}64.33 &
  \cellcolor[HTML]{ECF4FF}36.5025 \\ \cline{2-5} \cline{7-10} 
\multicolumn{1}{|l|}{} &
  \multicolumn{1}{l|}{} &
  27.86 &
  59.12 &
  \multicolumn{1}{r|}{37.87} &
  \multicolumn{1}{l|}{} &
  \multicolumn{1}{l|}{} &
  26.18 &
  63.59 &
  37.09 \\ \cline{3-5} \cline{8-10} 
\multicolumn{1}{|l|}{} &
  \multicolumn{1}{l|}{} &
  25.02 &
  60.64 &
  \multicolumn{1}{r|}{35.42} &
  \multicolumn{1}{l|}{} &
  \multicolumn{1}{l|}{} &
  26.63 &
  61.14 &
  37.1 \\ \cline{3-5} \cline{8-10} 
\multicolumn{1}{|l|}{} &
  \multicolumn{1}{l|}{} &
  29.33 &
  61.71 &
  \multicolumn{1}{r|}{39.76} &
  \multicolumn{1}{l|}{} &
  \multicolumn{1}{l|}{} &
  25.93 &
  63.57 &
  36.84 \\ \cline{3-5} \cline{8-10} 
\multicolumn{1}{|l|}{} &
  \multicolumn{1}{l|}{\multirow{-4}{*}{\textbf{20-shot}}} &
  30.18 &
  61.66 &
  \multicolumn{1}{r|}{40.52} &
  \multicolumn{1}{l|}{} &
  \multicolumn{1}{l|}{\multirow{-4}{*}{\textbf{20-shot}}} &
  27.54 &
  70.85 &
  39.66 \\ \cline{2-5} \cline{7-10} 
\multicolumn{1}{|l|}{\multirow{-15}{*}{\textbf{GPT 4}}} &
  \multicolumn{1}{l|}{\textbf{AVG}} &
  \cellcolor[HTML]{ECF4FF}28.0975 &
  \cellcolor[HTML]{ECF4FF}60.7825 &
  \multicolumn{1}{r|}{\cellcolor[HTML]{ECF4FF}38.3925} &
  \multicolumn{1}{l|}{\multirow{-15}{*}{\textbf{Llama3}}} &
  \multicolumn{1}{l|}{\textbf{AVG}} &
  \cellcolor[HTML]{ECF4FF}26.57 &
  \cellcolor[HTML]{ECF4FF}64.7875 &
  \cellcolor[HTML]{ECF4FF}37.6725 \\ \hline
\end{tabular}%
}
\caption{Averaged performance of the baseline dynamic prompt model on the Med-Mentions dataset across different shot settings.}
\label{tab:Medmentions_avg}
\end{table}

\clearpage
\section{Results of 95\% CIs for Each Metric}\label{app:B}

\begin{table}[h]
\centering
\renewcommand{\arraystretch}{1.2}
\resizebox{\columnwidth}{!}{%
\begin{tabular}{llllll}
\hline
 &
  \multicolumn{1}{r}{\cellcolor[HTML]{D9EAD3}Reddit\_Impacts} &
  \multicolumn{1}{r}{\cellcolor[HTML]{D9EAD3}BC5CDR} &
  \multicolumn{1}{r}{\cellcolor[HTML]{D9EAD3}MIMIC III} &
  \multicolumn{1}{c}{\cellcolor[HTML]{D9EAD3}NCBI} &
  \multicolumn{1}{c}{\cellcolor[HTML]{D9EAD3}Med-Mentions} \\ \hline
\rowcolor[HTML]{ECF4FF} 
\multicolumn{6}{l}{\cellcolor[HTML]{ECF4FF}\textbf{GPT-3.5}} \\ \hline
\rowcolor[HTML]{FCEBEB} 
\cellcolor[HTML]{FCEBEB}\textbf{Basic Prompt (BP)} &
  {\color[HTML]{333333} 16.73 {[}11.53, 22.83{]}} &
  {\color[HTML]{333333} 64.56 {[}61.55, 67.73{]}} &
  {\color[HTML]{333333} 54.70 {[}49.60, 58.73{]}} &
  {\color[HTML]{333333} 26.96 {[}24.43, 30.98{]}} &
  {\color[HTML]{333333} 9.27 {[}7.81, 12.22{]}} \\ \hline
\textbf{BP + Description of datasets} &
  {\color[HTML]{333333} 21.15 {[}14.88, 26.64{]}} &
  {\color[HTML]{333333} 68.61 {[}66.74, 70.72{]}} &
  {\color[HTML]{333333} 56.73 {[}52.58, 61.22{]}} &
  {\color[HTML]{333333} 34.48 {[}31.08, 39.25{]}} &
  {\color[HTML]{333333} 12.71 {[}10.93, 15.65{]}} \\ \hline
\textbf{BP + High-frequency instances} &
  {\color[HTML]{333333} 21.15 {[}15.75, 27.40{]}} &
  {\color[HTML]{333333} 69.01 {[}66.24, 70.98{]}} &
  {\color[HTML]{333333} 57.72 {[}52.75, 62.26{]}} &
  {\color[HTML]{333333} 35.95 {[}33.36, 38.44{]}} &
  {\color[HTML]{333333} 17.22 {[}14.47, 19.80{]}} \\ \hline
\textbf{BP + UMLS knowledge} &
  {\color[HTML]{333333} 16.44 {[}8.43, 23.07{]}} &
  {\color[HTML]{333333} 64.83 {[}61.83, 66.41{]}} &
  {\color[HTML]{333333} 50.57 {[}46.17, 55.04{]}} &
  {\color[HTML]{333333} 30.75 {[}27.73, 33.26{]}} &
  {\color[HTML]{333333} 10.88 {[}8.81, 12.29{]}} \\ \hline
\textbf{BP + Error analysis} &
  {\color[HTML]{333333} 19.24 {[}12.91, 26.17{]}} &
  {\color[HTML]{333333} 67.67 {[}65.53, 70.32{]}} &
  {\color[HTML]{333333} 59.52 {[}54.96, 64.47{]}} &
  {\color[HTML]{333333} 33.15 {[}31.24, 38.87{]}} &
  {\color[HTML]{333333} 15.52 {[}13.14, 17.20{]}} \\ \hline
\textbf{BP + 5-shot learning with sentences} &
  {\color[HTML]{333333} 19.30 {[}12.26, 25.78{]}} &
  {\color[HTML]{333333} 68.84 {[}67.25, 70.49{]}} &
  {\color[HTML]{333333} 57.03 {[}53.06, 62.85{]}} &
  {\color[HTML]{333333} 40.16 {[}38.78, 46.45{]}} &
  {\color[HTML]{333333} 20.61 {[}17.58, 22.29{]}} \\ \hline
\textbf{BP + 5-shot learning with tokens} &
  {\color[HTML]{333333} \textbf{21.69 {[}15.92, 28.89{]}}} &
  {\color[HTML]{333333} \textbf{70.79 {[}68.87, 73.15{]}}} &
  {\color[HTML]{333333} \textbf{61.21 {[}56.81, 66.05{]}}} &
  {\color[HTML]{333333} \textbf{43.01 {[}41.43, 48.21{]}}} &
  {\color[HTML]{333333} \textbf{24.57 {[}22.88, 26.64{]}}} \\ \hline
\textbf{BP + All above} &
  \cellcolor[HTML]{FCE5CD}{\color[HTML]{CC0000} \textbf{23.91 {[}15.87, 30.97{]}}} &
  \cellcolor[HTML]{FCE5CD}{\color[HTML]{CC0000} \textbf{72.73 {[}70.32, 74.86{]}}} &
  \cellcolor[HTML]{FCE5CD}{\color[HTML]{CC0000} \textbf{61.99 {[}57.24, 66.38{]}}} &
  \cellcolor[HTML]{FCE5CD}{\color[HTML]{CC0000} \textbf{45.24 {[}42.64, 50.58{]}}} &
  \cellcolor[HTML]{FCE5CD}{\color[HTML]{CC0000} \textbf{31.63 {[}29.36, 34.74{]}}} \\ \hline
\rowcolor[HTML]{E6F1FA} 
\multicolumn{6}{l}{\cellcolor[HTML]{E6F1FA}\textbf{GPT-4}} \\ \hline
\rowcolor[HTML]{FCEBEB} 
\textbf{Basic Prompt (BP)} &
  20.16 {[}13.29, 26.54{]} &
  69.43 {[}66.28, 72.44{]} &
  56.63 {[}51.27, 60.83{]} &
  33.56 {[}31.59, 37.25{]} &
  13.83 {[}11.85, 15.09{]} \\ \hline
\textbf{BP + Description of datasets} &
  23.52 {[}16.46, 30.84{]} &
  70.65 {[}67.47, 72.72{]} &
  59.68 {[}55.18, 64.09{]} &
  35.75 {[}33.54, 40.58{]} &
  15.30 {[}13.61, 17.15{]} \\ \hline
\textbf{BP + High-frequency instances} &
  24.64 {[}17.72, 31.11{]} &
  72.60 {[}71.17, 74.28{]} &
  60.08 {[}56.33, 65.37{]} &
  37.96 {[}36.95, 41.73{]} &
  19.50 {[}17.11, 22.97{]} \\ \hline
\textbf{BP + UMLS knowledge} &
  20.46 {[}13.84, 27.07{]} &
  69.86 {[}66.05, 72.62{]} &
  {\color[HTML]{333333} 55.13 {[}50.20, 60.29{]}} &
  {\color[HTML]{333333} 30.90 {[}28.68, 34.30{]}} &
  {\color[HTML]{333333} 14.50 {[}12.57, 16.46{]}} \\ \hline
\textbf{BP + Error analysis} &
  23.13 {[}16.65, 30.69{]} &
  {\color[HTML]{333333} 74.61 {[}71.44, 77.29{]}} &
  {\color[HTML]{333333} 60.11 {[}55.44, 64.72{]}} &
  {\color[HTML]{333333} 37.84 {[}34.13, 42.71{]}} &
  {\color[HTML]{333333} 18.25 {[}15.06, 20.43{]}} \\ \hline
\textbf{BP + 5-shot learning with sentences} &
  {\color[HTML]{333333} 22.88 {[}16.23, 30.59{]}} &
  {\color[HTML]{333333} 73.00 {[}71.26, 76.22{]}} &
  {\color[HTML]{333333} 58.25 {[}53.28, 63.95{]}} &
  {\color[HTML]{333333} 40.86 {[}39.37, 45.36{]}} &
  {\color[HTML]{333333} 28.80 {[}26.71, 30.20{]}} \\ \hline
\textbf{BP + 5-shot learning with tokens} &
  {\color[HTML]{333333} \textbf{25.95 {[}18.50, 32.07{]}}} &
  {\color[HTML]{333333} \textbf{76.65 {[}74.15, 77.92{]}}} &
  {\color[HTML]{333333} \textbf{62.94 {[}57.56, 66.87{]}}} &
  {\color[HTML]{333333} \textbf{44.24 {[}42.93, 48.28{]}}} &
  {\color[HTML]{333333} \textbf{33.20 {[}31.64, 35.70{]}}} \\ \hline
\textbf{BP + All above} &
  \cellcolor[HTML]{FCE5CD}{\color[HTML]{CC0000} \textbf{27.60 {[}19.43, 33.80{]}}} &
  \cellcolor[HTML]{FCE5CD}{\color[HTML]{CC0000} \textbf{78.03 {[}75.51, 80.02{]}}} &
  \cellcolor[HTML]{FCE5CD}{\color[HTML]{CC0000} \textbf{63.58 {[}58.73, 67.18{]}}} &
  \cellcolor[HTML]{FCE5CD}{\color[HTML]{CC0000} \textbf{46.93 {[}44.85, 51.58{]}}} &
  \cellcolor[HTML]{FCE5CD}{\color[HTML]{CC0000} \textbf{37.95 {[}35.88, 39.90{]}}} \\ \hline
\rowcolor[HTML]{E6F1FA} 
\multicolumn{6}{l}{\cellcolor[HTML]{E6F1FA}\textbf{Llama3}} \\ \hline
\rowcolor[HTML]{FCEBEB} 
\textbf{Basic Prompt (BP)} &
  15.61 {[}8.20, 22.12{]} &
  62.13 {[}59.24, 63.58{]} &
  50.70 {[}45.93, 54.19{]} &
  19.15 {[}15.21, 21.38{]} &
  21.23 {[}19.24, 23.42{]} \\ \hline
\textbf{BP + Description of datasets} &
  19.28 {[}11.71, 25.96{]} &
  67.68 {[}64.86, 69.10{]} &
  56.22 {[}52.77, 60.25{]} &
  21.44 {[}20.80, 24.65{]} &
  21.57 {[}19.30, 24.76{]} \\ \hline
\textbf{BP + High-frequency instances} &
  {\color[HTML]{333333} \textbf{20.44 {[}13.79, 27.51{]}}} &
  {\color[HTML]{333333} 68.39 {[}66.48, 70.35{]}} &
  {\color[HTML]{333333} 56.06 {[}52.62, 61.42{]}} &
  {\color[HTML]{333333} 26.62 {[}22.16, 28.31{]}} &
  {\color[HTML]{333333} 27.12 {[}26.37, 29.35{]}} \\ \hline
\textbf{BP + UMLS knowledge} &
  {\color[HTML]{333333} 12.91 {[}7.40, 18.71{]}} &
  {\color[HTML]{333333} 64.71 {[}61.44, 67.01{]}} &
  {\color[HTML]{333333} 48.92 {[}44.75, 53.37{]}} &
  {\color[HTML]{333333} 20.91 {[}17.07, 22.61{]}} &
  {\color[HTML]{333333} 23.68 {[}20.59, 25.17{]}} \\ \hline
\textbf{BP + Error analysis} &
  18.87 {[}13.34, 25.13{]} &
  68.07 {[}65.41, 70.58{]} &
  58.92 {[}53.90, 63.84{]} &
  24.46 {[}20.97, 25.20{]} &
  25.78 {[}23.48, 27.56{]} \\ \hline
\textbf{BP + 5-shot learning with sentences} &
  17.65 {[}13.62, 24.69{]} &
  70.70 {[}69.36, 72.83{]} &
  56.85 {[}52.32, 61.33{]} &
  30.52 {[}26.50, 33.96{]} &
  34.87 {[}32.18, 37.25{]} \\ \hline
\textbf{BP + 5-shot learning with tokens} &
  20.04 {[}14.81, 27.29{]} &
  {\color[HTML]{333333} \textbf{71.76 {[}69.58, 73.51{]}}} &
  {\color[HTML]{333333} \textbf{61.98 {[}56.59, 65.18{]}}} &
  {\color[HTML]{333333} \textbf{33.42 {[}28.72, 35.12{]}}} &
  {\color[HTML]{333333} \textbf{35.23 {[}33.17, 37.08{]}}} \\ \hline
\textbf{BP + All above} &
  \cellcolor[HTML]{FCE5CD}{\color[HTML]{CC0000} \textbf{21.43 {[}14.24, 28.80{]}}} &
  \cellcolor[HTML]{FCE5CD}{\color[HTML]{CC0000} \textbf{73.32 {[}72.27, 74.26{]}}} &
  \cellcolor[HTML]{FCE5CD}{\color[HTML]{CC0000} \textbf{62.94 {[}57.07, 65.79{]}}} &
  \cellcolor[HTML]{FCE5CD}{\color[HTML]{CC0000} \textbf{34.80 {[}28.57, 35.44{]}}} &
  \cellcolor[HTML]{FCE5CD}{\color[HTML]{CC0000} \textbf{37.26 {[}35.45, 39.08{]}}} \\ \hline
\end{tabular}%
}
\caption[95\% CIs of static prompting strategies]{Evaluation of static prompting strategies using GPT-3.5, GPT-4 and Llama 3 across five biomedical datasets. The table presents F$_1$-score with 95\% confidence intervals reported for each metric to indicate the statistical reliability of the results.}
\label{tab:CI_F1_static}
\end{table}

\begin{table}[h]
\centering
\renewcommand{\arraystretch}{1.2}
\resizebox{\columnwidth}{!}{%
\begin{tabular}{lllllll}
\hline
 &
   &
  \multicolumn{1}{c}{\cellcolor[HTML]{D9EAD3}\textit{\textbf{Reddit\_Impacts}}} &
  \multicolumn{1}{c}{\cellcolor[HTML]{D9EAD3}\textit{\textbf{BC5CDR}}} &
  \multicolumn{1}{c}{\cellcolor[HTML]{D9EAD3}\textit{\textbf{MIMIC III}}} &
  \multicolumn{1}{c}{\cellcolor[HTML]{D9EAD3}\textit{\textbf{NCBI}}} &
  \multicolumn{1}{c}{\cellcolor[HTML]{D9EAD3}\textit{\textbf{Med-Mentions}}} \\ \hline
\rowcolor[HTML]{ECF4FF} 
\textbf{GPT-4} &
   &
  \multicolumn{1}{r}{\cellcolor[HTML]{ECF4FF}} &
  \multicolumn{1}{r}{\cellcolor[HTML]{ECF4FF}} &
  \multicolumn{1}{r}{\cellcolor[HTML]{ECF4FF}} &
   &
   \\ \hline
 &
  \cellcolor[HTML]{FCEBEB}\textbf{Base} &
  \cellcolor[HTML]{FCEBEB}27.60 {[}19.43, 33.80{]} &
  \cellcolor[HTML]{FCEBEB}78.03 {[}75.51, 80.02{]} &
  \cellcolor[HTML]{FCEBEB}63.58 {[}58.73, 67.18{]} &
  \cellcolor[HTML]{FCEBEB}46.93 {[}44.85, 51.58{]} &
  \cellcolor[HTML]{FCEBEB}37.95 {[}35.88, 39.90{]} \\ \cline{2-7} 
 &
  \textbf{TF-IDF} &
  28.47 {[}21.78, 35.47{]} &
  {\color[HTML]{CB0000} \textbf{85.88 {[}84.53, 86.42{]}}} &
  {\color[HTML]{CB0000} \textbf{76.24 {[}72.98, 79.63{]}}} &
  {\color[HTML]{CB0000} \textbf{60.08 {[}56.70, 63.32{]}}} &
  37.96 {[}35.90, 39.84{]} \\ \cline{2-7} 
 &
  \textbf{SBERT} &
  {\color[HTML]{CB0000} \textbf{33.72 {[}26.28, 42.20{]}}} &
  83.37 {[}82.51, 84.22{]} &
  73.44 {[}69.91, 76.81{]} &
  57.56 {[}54.05, 60.73{]} &
  39.12 {[}36.84, 41.34{]} \\ \cline{2-7} 
 &
  \textbf{ColBERT} &
  32.39 {[}25.10, 39.85{]} &
  79.82 {[}78.24, 80.98{]} &
  75.56 {[}72.06, 78.94{]} &
  52.38 {[}49.06, 55.55{]} &
  {\color[HTML]{CB0000} \textbf{39.93 {[}37.93, 41.73{]}}} \\ \cline{2-7} 
\multirow{-5}{*}{5-shot} &
  \textbf{DPR} &
  32.64 {[}25.42, 40.17{]} &
  83.58 {[}82.30, 84.88{]} &
  69.89 {[}65.75, 73.63{]} &
  49.37 {[}45.37, 52.94{]} &
  39.13 {[}34.44, 41.35{]} \\ \hline
 &
  \cellcolor[HTML]{FCEBEB}\textbf{Base} &
  \cellcolor[HTML]{FCEBEB}31.92 {[}23.77, 38.44{]} &
  \cellcolor[HTML]{FCEBEB}81.27 {[}80.81, 82.37{]} &
  \cellcolor[HTML]{FCEBEB}70.52 {[}66.10, 73.81{]} &
  \cellcolor[HTML]{FCEBEB}52.67 {[}49.36, 56.76{]} &
  \cellcolor[HTML]{FCEBEB}36.74 {[}32.29, 38.83{]} \\ \cline{2-7} 
 &
  \textbf{TF-IDF} &
  31.14 {[}24.33, 38.13{]} &
  {\color[HTML]{CB0000} \textbf{86.64 {[}85.15, 88.09{]}}} &
  75.53 {[}72.18, 79.10{]} &
  {\color[HTML]{CB0000} \textbf{62.05 {[}58.79, 65.11{]}}} &
  40.37 {[}38.23, 42.43{]} \\ \cline{2-7} 
 &
  \textbf{SBERT} &
  {\color[HTML]{CB0000} \textbf{35.47 {[}27.17, 43.21{]}}} &
  85.92 {[}83.09, 87.27{]} &
  73.89 {[}70.22, 77.80{]} &
  60.83 {[}57.47, 64.03{]} &
  40.37 {[}38.23, 42.39{]} \\ \cline{2-7} 
 &
  \textbf{ColBERT} &
  33.81 {[}26.24, 41.55{]} &
  85.71 {[}84.42, 86.07{]} &
  {\color[HTML]{CB0000} \textbf{76.34 {[}73.01, 79.68{]}}} &
  57.25 {[}53.75, 60.72{]} &
  {\color[HTML]{CB0000} \textbf{40.48 {[}38.13, 42.54{]}}} \\ \cline{2-7} 
\multirow{-5}{*}{10-shot} &
  \textbf{DPR} &
  32.61 {[}24.50, 40.33{]} &
  84.79 {[}82.96, 86.78{]} &
  72.13 {[}68.06, 75.85{]} &
  58.70 {[}54.99, 61.91{]} &
  40.25 {[}30.83, 50.75{]} \\ \hline
 &
  \cellcolor[HTML]{FCEBEB}\textbf{Base} &
  \cellcolor[HTML]{FCEBEB}37.67 {[}30.04, 43.44{]} &
  \cellcolor[HTML]{FCEBEB}81.15 {[}80.40, 82.24{]} &
  \cellcolor[HTML]{FCEBEB}70.98 {[}65.77, 73.82{]} &
  \cellcolor[HTML]{FCEBEB}51.98 {[}50.33, 58.84{]} &
  \cellcolor[HTML]{FCEBEB}38.39 {[}35.26, 40.29{]} \\ \cline{2-7} 
 &
  \textbf{TF-IDF} &
  38.35 {[}30.77, 46.28{]} &
  87.16 {[}85.77, 88.62{]} &
  {\color[HTML]{CB0000} \textbf{77.66 {[}71.91, 78.88{]}}} &
  {\color[HTML]{CB0000} \textbf{64.36 {[}61.18, 67.87{]}}} &
  {\color[HTML]{CB0000} \textbf{41.32 {[}39.21, 43.26{]}}} \\ \cline{2-7} 
 &
  \textbf{SBERT} &
  38.22 {[}28.57, 44.90{]} &
  {\color[HTML]{CB0000} \textbf{87.42 {[}85.26, 89.12{]}}} &
  75.14 {[}71.77, 78.75{]} &
  62.21 {[}59.01, 65.18{]} &
  39.37 {[}35.05, 40.39{]} \\ \cline{2-7} 
 &
  \textbf{ColBERT} &
  {\color[HTML]{CB0000} \textbf{42.49 {[}32.52, 48.33{]}}} &
  83.00 {[}81.39, 84.40{]} &
  76.70 {[}73.11, 79.89{]} &
  57.69 {[}54.22, 61.18{]} &
  40.53 {[}37.61, 43.26{]} \\ \cline{2-7} 
\multirow{-5}{*}{20-shot} &
  \textbf{DPR} &
  38.84 {[}29.01, 44.44{]} &
  85.60 {[}84.28, 86.93{]} &
  72.28 {[}68.56, 75.95{]} &
  60.34 {[}56.54, 63.72{]} &
  39.23 {[}34.22, 41.56{]} \\ \hline
\rowcolor[HTML]{ECF4FF} 
\textbf{Llama3} &
  \textbf{} &
  \textbf{} &
  \textbf{} &
  \textbf{} &
   &
  \multicolumn{1}{r}{\cellcolor[HTML]{ECF4FF}} \\ \hline
 &
  \cellcolor[HTML]{FCEBEB}\textbf{Base} &
  \cellcolor[HTML]{FCEBEB}21.43 {[}14.24, 28.80{]} &
  \cellcolor[HTML]{FCEBEB}73.32 {[}72.27, 74.26{]} &
  \cellcolor[HTML]{FCEBEB}62.94 {[}57.07, 65.79{]} &
  \cellcolor[HTML]{FCEBEB}34.80 {[}28.57, 35.44{]} &
  \cellcolor[HTML]{FCEBEB}37.26 {[}35.45, 39.08{]} \\ \cline{2-7} 
 &
  \textbf{TF-IDF} &
  28.57 {[}21.74, 36.06{]} &
  80.11 {[}79.25, 81.00{]} &
  70.41 {[}66.87, 73.76{]} &
  49.80 {[}46.38, 53.03{]} &
  38.68 {[}35.67, 40.81{]} \\ \cline{2-7} 
 &
  \textbf{SBERT} &
  {\color[HTML]{CB0000} \textbf{34.42 {[}26.28, 41.52{]}}} &
  {\color[HTML]{CB0000} \textbf{80.39 {[}79.50, 81.33{]}}} &
  67.88 {[}64.09, 71.69{]} &
  {\color[HTML]{CB0000} \textbf{50.12 {[}46.89, 53.66{]}}} &
  37.91 {[}36.02, 39.81{]} \\ \cline{2-7} 
 &
  \textbf{ColBERT} &
  32.94 {[}25.00, 39.84{]} &
  71.76 {[}70.75, 72.69{]} &
  {\color[HTML]{CB0000} \textbf{71.68 {[}68.08, 75.21{]}}} &
  45.50 {[}41.95, 49.49{]} &
  {\color[HTML]{CB0000} \textbf{38.99 {[}36.15, 41.34{]}}} \\ \cline{2-7} 
\multirow{-5}{*}{5-shot} &
  \textbf{DPR} &
  29.00 {[}22.86, 36.36{]} &
  75.67 {[}74.67, 76.70{]} &
  68.97 {[}65.05, 72.70{]} &
  44.54 {[}41.24, 48.25{]} &
  38.66 {[}36.78, 40.50{]} \\ \hline
 &
  \cellcolor[HTML]{FCEBEB}\textbf{Base} &
  \cellcolor[HTML]{FCEBEB}32.50 {[}26.94, 42.26{]} &
  \cellcolor[HTML]{FCEBEB}75.15 {[}74.65, 76.67{]} &
  \cellcolor[HTML]{FCEBEB}63.77 {[}58.59, 67.75{]} &
  \cellcolor[HTML]{FCEBEB}35.60 {[}32.17, 39.12{]} &
  \cellcolor[HTML]{FCEBEB}36.50 {[}35.73, 39.57{]} \\ \cline{2-7} 
 &
  \textbf{TF-IDF} &
  34.21 {[}27.24, 42.03{]} &
  {\color[HTML]{CB0000} \textbf{80.57 {[}79.65, 81.47{]}}} &
  55.56 {[}53.11, 60.44{]} &
  49.50 {[}46.05, 52.92{]} &
  35.51 {[}34.75, 37.45{]} \\ \cline{2-7} 
 &
  \textbf{SBERT} &
  32.45 {[}25.33, 39.63{]} &
  81.17 {[}80.26, 82.03{]} &
  71.63 {[}67.75, 75.15{]} &
  {\color[HTML]{CB0000} \textbf{51.35 {[}47.49, 55.16{]}}} &
  {\color[HTML]{CC0000} \textbf{39.08 {[}36.39, 41.38{]}}} \\ \cline{2-7} 
 &
  \textbf{ColBERT} &
  32.89 {[}20.35, 35.05{]} &
  80.34 {[}79.53, 81.24{]} &
  {\color[HTML]{CB0000} \textbf{72.85 {[}69.46, 76.49{]}}} &
  38.77 {[}34.91, 42.29{]} &
  38.06 {[}35.52, 40.71{]} \\ \cline{2-7} 
\multirow{-5}{*}{10-shot} &
  \textbf{DPR} &
  {\color[HTML]{CB0000} \textbf{34.29 {[}26.11, 41.98{]}}} &
  74.72 {[}73.77, 75.73{]} &
  69.54 {[}65.61, 73.17{]} &
  46.28 {[}42.77, 49.65{]} &
  37.85 {[}36.06, 39.69{]} \\ \hline
 &
  \cellcolor[HTML]{FCEBEB}\textbf{Base} &
  \cellcolor[HTML]{FCEBEB}33.67 {[}24.09, 40.88{]} &
  \cellcolor[HTML]{FCEBEB}75.50 {[}73.57, 76.36{]} &
  \cellcolor[HTML]{FCEBEB}62.05 {[}58.23, 67.15{]} &
  \cellcolor[HTML]{FCEBEB}41.62 {[}38.83, 45.71{]} &
  \cellcolor[HTML]{FCEBEB}37.67 {[}35.22, 40.57{]} \\ \cline{2-7} 
 &
  \textbf{TF-IDF} &
  39.11 {[}31.34, 47.70{]} &
  {\color[HTML]{CB0000} \textbf{78.36 {[}77.42, 79.30{]}}} &
  57.66 {[}51.19, 59.80{]} &
  47.50 {[}43.87, 50.84{]} &
  {\color[HTML]{CB0000} \textbf{38.83 {[}37.54, 39.11{]}}} \\ \cline{2-7} 
 &
  \textbf{SBERT} &
  {\color[HTML]{CB0000} \textbf{41.43 {[}31.58, 48.98{]}}} &
  76.85 {[}74.86, 78.96{]} &
  65.35 {[}60.44, 70.40{]} &
  44.14 {[}40.57, 47.86{]} &
  36.01 {[}34.16, 37.75{]} \\ \cline{2-7} 
 &
  \textbf{ColBERT} &
  34.66 {[}24.07, 36.31{]} &
  72.19 {[}71.17, 73.20{]} &
  57.63 {[}53.19, 61.93{]} &
  {\color[HTML]{CB0000} \textbf{48.44 {[}45.07, 51.78{]}}} &
  36.85 {[}34.10, 39.29{]} \\ \cline{2-7} 
\multirow{-5}{*}{20-shot} &
  \textbf{DPR} &
  37.30 {[}27.13, 44.76{]} &
  74.80 {[}72.49, 76.36{]} &
  {\color[HTML]{CB0000} \textbf{65.80 {[}61.82, 69.69{]}}} &
  40.36 {[}36.96, 43.96{]} &
  36.89 {[}34.46, 38.84{]} \\ \hline
\end{tabular}%
}
\caption[95\% CIs of dynamic prompting strategies]{Evaluation of dynamic prompting strategies (5-shot, 10-shot, and 20-shot) using GPT-4 and Llama 3 across five biomedical datasets. The table presents F$_1$-score for each retrieval method: Base Prompt, TF-IDF, SBERT, ColBERT, and DPR, with 95\% confidence intervals reported for each metric to indicate the statistical reliability of the results.}
\label{tab:CI_F1_dynamic}
\end{table}

\clearpage
\section{Detailed Task-specific Static Prompts}\label{app:C}

\begin{table}[h]
\centering
\renewcommand{\arraystretch}{1.2}
\resizebox{\columnwidth}{!}{%
\begin{tabular}{ll}
\hline
\rowcolor[HTML]{FCEBEB} 
\textbf{Prompt Strategies} &
  \multicolumn{1}{c}{\cellcolor[HTML]{D9EAD3}\textit{\textbf{Reddit-Impacts}}} \\ \hline
 &
  \begin{tabular}[c]{@{}l@{}}\textbf{[Task Description]:} You are a medical AI trained to identify and classify tokens into \textcolor{blue}{three categories: Clinical Impacts, Social Impacts, and Outside ('O').} \\ Your task is to extract and classify the clinical and social impacts from this dataset, considering your knowledge of the lifestyle of this population and the \\ potential clinical and social impacts they might experience.\end{tabular} \\ \cline{2-2} 
 &
  \begin{tabular}[c]{@{}l@{}}\textbf{[Entity Types with Definitions]:} \textcolor{blue}{'Clinical Impacts' refer to} tokens describing the effects, consequences, or impacts of substance use on individual health or \\ well-being, as defined in UMLS. \textcolor{blue}{'Social Impacts' describe} the societal, interpersonal, or community-level effects, also based on UMLS definitions. Any token \\ not falling into these categories should be labeled as 'O'.\end{tabular} \\ \cline{2-2} 
\multirow{-3}{*}{\textbf{Basic Prompt}} &
  \begin{tabular}[c]{@{}l@{}}\textbf{[Format Specification]:} For example, \textcolor{blue}{the sentence} 'I was a codeine addict.' \textcolor{blue}{is tokenized and labeled} as follows: {[}'I', 'was', 'a', 'codeine', 'addict', '.'{]} \\ with labels {[}'O', 'O', 'O', 'Clinical Impacts', 'Clinical Impacts', 'O'{]}. Your task is to predict and return the label for each provided token, ensuring the number of \\ output labels matches the number of input tokens exactly. \textcolor{blue}{The output format should be tokens with their labels:} {[}'I-O', 'was-O', 'a-O', 'codeine-Clinical Impacts', \\ 'addict-Clinical Impacts', '.-O'{]}.\end{tabular} \\ \hline
\textbf{Description of datasets} &
  \begin{tabular}[c]{@{}l@{}}The data you are working with has been collected from 14 forums on Reddit (subreddits) that focused on prescription and illicit opioids, and medications \\ for opioid use disorder. This dataset represents a social media context, coming from individuals who may use prescription and illicit opioids and stimulants.\end{tabular} \\ \hline
\textbf{High-frequency instances} &
  \begin{tabular}[c]{@{}l@{}}In this dataset, \textcolor{blue}{high-frequency clinical impacts} include 'withdrawal', 'rehab', 'addicted', 'detox', 'overdosed', and 'rehabs'. \textcolor{blue}{High-frequency social impacts} include \\ 'lost', 'homeless', 'charged', 'streets', 'jail', and 'disorderly'.\end{tabular} \\ \hline
\textbf{UMLS knowledge} &
  \begin{tabular}[c]{@{}l@{}}\textcolor{blue}{The Unified Medical Language System (UMLS) }is developed by the U.S. National Library of Medicine (NLM) to integrate and standardize diverse medical \\ terminologies and coding systems. It consists of three main components: the Metathesaurus, Semantic Network, and SPECIALIST Lexicon, supporting medical \\ information retrieval and semantic analysis. \textcolor{blue}{You understand medical terminology and concepts from UMLS.}\end{tabular} \\ \hline
\textbf{Error analysis} &
  \begin{tabular}[c]{@{}l@{}}\textcolor{blue}{Possible analysis of prediction errors:} If a sentence describes the background information of an event, facility, or project, then even if it mentions keywords \\ related to social impact like 'at jail', it still cannot be determined as describing a patient being in jail. It is essential to clearly determine whether the sentence is \\ describing the patient's condition. Second, if the sentence is about the usage, operation, or introduction of a drug or medicine, it does not belong to the patient's \\ clinical impacts, even if it mentions some symptoms. Pay attention to whether the sentence contains words like 'if' that indicate conditions.\end{tabular} \\ \hline
\end{tabular}%
}
\caption{Specific static prompts for each component we used for the \textsc{Reddit-Impacts} dataset.}
\label{tab:reddit_impacts_prompt}
\end{table}
\begin{table}[h]
\centering
\renewcommand{\arraystretch}{1.2}
\resizebox{\columnwidth}{!}{%
\begin{tabular}{ll}
\hline
\rowcolor[HTML]{FCEBEB} 
\textbf{Prompt Strategies} &
  \multicolumn{1}{c}{\cellcolor[HTML]{D9EAD3}\textit{\textbf{BC5CDR}}} \\ \hline
 &
  \begin{tabular}[c]{@{}l@{}}\textbf{[Task Description]:} You are a medical AI trained to identify and classify tokens into \textcolor{blue}{three categories: 'Disease', 'Chemical' and Outside ('O').} Your task is to extract \\ and classify the Disease and Chemical related concepts from this dataset.\end{tabular} \\ \cline{2-2} 
 &
  \begin{tabular}[c]{@{}l@{}}\textbf{[Entity Types with Definitions]:} \textcolor{blue}{'Disease' is a} particular abnormal condition that adversely affects the structure or function of all or part of an organism and is \\ not immediately due to any external injury. Diseases are often known to be medical conditions that are associated with specific signs and symptoms. A disease \\ may be caused by external factors such as pathogens or by internal dysfunctions. For example, internal dysfunctions of the immune system can produce a variety \\ of different diseases, including various forms of immunodeficiency, hypersensitivity, allergies, and autoimmune disorders. \textcolor{blue}{'Chemical' in this context refers to} \\ substances or compounds with specific chemical properties and structures. These can include drugs, neurotransmitters, elements or ions, vitamins, and other \\ medically relevant chemicals. Any token not falling into Disease categories should be labeled as 'O'.\end{tabular} \\ \cline{2-2} 
\multirow{-3}{*}{\textbf{Basic Prompt}} &
  \begin{tabular}[c]{@{}l@{}}\textbf{[Format Specification]:} For example, \textcolor{blue}{the sentence} 'The hypotensive effect of 100 mg / kg alpha-methyldopa was also partially reversed by naloxone .' \textcolor{blue}{is} \\ \textcolor{blue}{tokenized and labeled} as follows: {[}'The', 'hypotensive', 'effect', 'of', '100', 'mg', '/', 'kg', 'alpha-methyldopa', 'was', 'also', 'partially', 'reversed', 'by', 'naloxone', '.'{]} with\\ labels {[}'O', 'Disease', 'O', 'O', 'O', 'O', 'O', 'O', 'Chemical', 'O', 'O', 'O', 'O', 'O', 'Chemical', 'O'{]}. Your task is to predict and return the label for each provided token, \\ ensuring the number of output labels matches the number of input tokens exactly. \textcolor{blue}{The output format should include tokens with their labels:} {[}'The-O', 'hypotensive-\\ Disease', 'effect-O', 'of-O', '100-O', 'mg-O', '/-O', 'kg-O', 'alpha-methyldopa-Chemical', 'was-O', 'also-O', 'partially-O', 'reversed-O', 'by-O', 'naloxone-Chemical', '.-O'{]}.\end{tabular} \\ \hline
\textbf{Description of datasets} &
  \begin{tabular}[c]{@{}l@{}}The data you are working with is \textcolor{blue}{BC5CDR dataset}, a benchmark dataset for biomedical natural language processing, created from PubMed abstracts. It includes \\ annotations for two entity types—chemicals and diseases—and their relationships, specifically chemical-induced disease interactions. The dataset is widely used \\ for tasks such as named entity recognition and relation extraction, supporting research in biomedical text mining and information extraction.\end{tabular} \\ \hline
\textbf{High-frequency instances} &
  \begin{tabular}[c]{@{}l@{}}In this dataset, \textcolor{blue}{high frequency of 'Disease'} include 'pain', 'toxicity', 'renal', 'failure', 'disease', 'hypotension'; \textcolor{blue}{high frequency of 'Chemical'} include 'cocaine', 'acid', \\ 'dopamine', 'nicotine', 'morphine', 'lithium'.\end{tabular} \\ \hline
\textbf{UMLS knowledge} &
  \begin{tabular}[c]{@{}l@{}}\textcolor{blue}{The Unified Medical Language System (UMLS) }is developed by the U.S. National Library of Medicine (NLM) to integrate and standardize diverse medical \\ terminologies and coding systems. It consists of three main components: the Metathesaurus, Semantic Network, and SPECIALIST Lexicon, supporting medical \\ information retrieval and semantic analysis. \textcolor{blue}{You understand medical terminology and concepts from UMLS.}\end{tabular} \\ \hline
\textbf{Error analysis} &
  \begin{tabular}[c]{@{}l@{}}\textcolor{blue}{Possible analysis of prediction errors:} The prediction errors mainly stem from challenges in distinguishing between entity boundaries and contextual usage. For \\ instance, multi-token entities were partially labeled, causing boundary mismatches. Additionally, certain terms such as "receptor" or "antagonist" were incorrectly \\ labeled as 'O,' despite being part of chemical-related entities. Misclassification also occurred in sentences with conditional phrases or background information, \\ where the relation between entities was not accurately captured. Furthermore, entities mentioned in descriptive or abstract contexts, were sometimes overlooked. \\ These errors highlight difficulties in handling complex sentence structures, context-specific classification, and multi-token entity recognition.\end{tabular} \\ \hline
\end{tabular}%
}
\caption{Specific static prompts for each component we used for the BC5CDR dataset.}
\label{tab:bc5cdr_prompt}
\end{table}
\begin{table}[h]
\centering
\renewcommand{\arraystretch}{1.2}
\resizebox{\columnwidth}{!}{%
\begin{tabular}{ll}
\hline
\rowcolor[HTML]{FCEBEB} 
\textbf{Prompt Strategies} &
  \multicolumn{1}{c}{\cellcolor[HTML]{D9EAD3}\textit{\textbf{MIMIC III}}} \\ \hline
 &
  \begin{tabular}[c]{@{}l@{}}\textbf{[Task Description]:} You are a medical AI trained to identify and classify tokens into \textcolor{blue}{13 categories: 'CONDITION/SYMPTOM', 'DRUG', 'AMOUNT', 'TIME',} \\ \textcolor{blue}{'MEASUREMENT', 'LOCATION', 'EVENT', 'FREQUENCY', 'ORGANIZATION', 'DATE', 'AGE', 'GENDER' and Outside ('O').} Your task is to extract and \\ classify the concepts from this dataset.\end{tabular} \\ \cline{2-2} 
 &
  \begin{tabular}[c]{@{}l@{}}\textbf{[Entity Types with Definitions]:} \textcolor{blue}{'ORGANIZATION' refers to} entities or groups associated with healthcare or emergency medical services. These could be specific \\ departments, teams, or services within a medical or emergency response organization. \textcolor{blue}{'DATE' in this context refers to} specific calendar dates. These dates are \\ typically used to mark particular events, appointments, or deadlines. \textcolor{blue}{'AGE' in this context refers to} the length of time that a person has lived or the number of years \\ since their birth. It can be expressed in various formats, including numerical values, abbreviated forms, or written out in words. \textcolor{blue}{'GENDER' in this context refers to} \\ the socially constructed roles, behaviors, activities, and attributes that a given society considers appropriate for men and women. It encompasses the identities of \\ 'male' and 'female,' which are often associated with biological sex but are also shaped by cultural and social factors. \textcolor{blue}{'FREQUENCY' in this context refers to} the rate \\ or regularity at which an event or phenomenon occurs. It can describe how often something happens, ranging from sporadic or irregular occurrences to more regular \\ or constant patterns. \textcolor{blue}{'EVENT' in this context refers to} specific occurrences or actions that take place, particularly in a medical or clinical setting. These can include \\ procedures, assessments, or other significant incidents. \textcolor{blue}{'LOCATION' in this context refers to} specific places or areas, particularly within a healthcare or medical \\ setting. These can include types of facilities, specific locations within a facility, or other relevant places. \textcolor{blue}{'MEASUREMENT' in this context refers to} quantitative \\ assessments or values used to evaluate specific physiological or medical parameters. These can include vital signs, laboratory test results, numerical values, or \\ other metrics related to patient health. \textcolor{blue}{'TIME' in this context refers to} specific points or periods in the temporal continuum, particularly as they relate to healthcare or \\ medical events. These can include general time references, specific durations, or events tied to time. \textcolor{blue}{'AMOUNT' in this context refers to} specific quantities or \\ dosages, particularly in a medical or pharmaceutical setting. These can include measurements of medication, frequency or number of administrations, and methods of\\ delivery. \textcolor{blue}{'DRUG' in this context refers to} specific medications or pharmaceutical substances used in the treatment, prevention, or diagnosis of diseases. These can \\ include brand names, generic names, or forms of administration. \textcolor{blue}{'CONDITION/SYMPTOM' in this context refers to} physical or subjective signs that indicate a medical \\ condition or disease. These can include sensations of discomfort, specific types of pain or discomfort, respiratory issues, or gastrointestinal symptoms. Any token not \\ falling into categories above should be labeled as 'O'.\end{tabular} \\ \cline{2-2} 
\multirow{-3}{*}{\textbf{Basic Prompt}} &
  \begin{tabular}[c]{@{}l@{}}\textbf{[Format Specification]:} For example, \textcolor{blue}{the sentence} 'The patient was readmitted to the hospital on 2195-6-6 due to fevers to 103 at the rehabilitation facility despite \\ being on intravenous antibiotics HISTORY OF PRESENT ILLNESS 55 year-old female presents with 2/5 week history of non-bloody diarrhea' \textcolor{blue}{is tokenized and} \\ \textcolor{blue}{labeled} as follows: {[}'The', 'patient', 'was', 'readmitted', 'to', 'the', 'hospital', 'on', '2195-6-6', 'due', 'to', 'fevers', 'to', '103', 'at', 'the', 'rehabilitation', 'facility', 'despite', \\ 'being', 'on', 'intravenous', 'antibiotics', 'HISTORY', 'OF', 'PRESENT', 'ILLNESS', '55', 'year-old', 'female', 'presents', 'with', '2/5', 'week', 'history', 'of', 'non-bloody', \\ 'diarrhea'{]} with labels {[}'O', 'O', 'O', 'EVENT', 'O', 'LOCATION', 'LOCATION', 'O', 'O', 'O', 'O', 'MEASUREMENT', 'MEASUREMENT', 'MEASUREMENT', 'O', \\ 'LOCATION', 'LOCATION', 'LOCATION', 'O', 'O', 'O', 'DRUG', 'DRUG', 'O', 'O', 'O', 'O', 'AGE', 'O', 'GENDER', 'O', 'O', 'AMOUNT', 'AMOUNT', 'O', 'O', \\ 'CONDITION/SYMPTOM', 'CONDITION/SYMPTOM'{]}. Your task is to predict and return the label for each provided token, ensuring the number of output labels \\ matches the number of input tokens exactly. \textcolor{blue}{The output format should include tokens with their labels:} {[}'The-O', 'patient-O', 'was-O', 'readmitted-EVENT', 'to-O', \\ 'the-LOCATION', 'hospital-LOCATION', 'on-O', '2195-6-6-O', 'due-O', 'to-O', 'fevers-MEASUREMENT', 'to-MEASUREMENT', '103-MEASUREMENT', 'at-O', \\ 'the-LOCATION', 'rehabilitation-LOCATION', 'facility-LOCATION', 'despite-O', 'being-O', 'on-O', 'intravenous-DRUG', 'antibiotics-DRUG', 'HISTORY-O', 'OF-O', \\ 'PRESENT-O', 'ILLNESS-O', '55-AGE', 'year-old-O', 'female-GENDER', 'presents-O', 'with-O', '2/5-AMOUNT', 'week-AMOUNT', 'history-O', 'of-O', 'non-bloody-\\ CONDITION/SYMPTOM', 'diarrhea-CONDITION/SYMPTOM'{]}.\end{tabular} \\ \hline
\textbf{Description of datasets} &
  \begin{tabular}[c]{@{}l@{}}The data you are working with is MIMIC-III (Medical Information Mart for Intensive Care) dataset, a large, publicly available database containing de-identified \\ health data from critical care patients at the Beth Israel Deaconess Medical Center. It includes structured data, such as demographics, lab results, and vital signs, as \\ well as unstructured data, such as clinical notes and discharge summaries. The dataset is widely used for research in machine learning, natural language processing, \\ and clinical decision support to improve healthcare outcomes.\end{tabular} \\ \hline
\textbf{High-frequency instances} &
  \begin{tabular}[c]{@{}l@{}}In this dataset, \textcolor{blue}{high frequency of 'CONDITION/SYMPTOM'} include 'pain', 'chest', 'cough', 'breath', 'nausea', 'abdominal'; \textcolor{blue}{high frequency of 'DRUG'} include 'iv', \\ 'lasix', 'ceftriaxone', 'oxygen', 'ns', 'coumadin'; high frequency of 'AMOUNT' include 'iv', '2', '1', 'mg', 'days', 'one'; \textcolor{blue}{high frequency of 'TIME'} include 'day', 'admission', \\ 'prior', 'last', 'ago', 'morning'; \textcolor{blue}{high frequency of 'MEASUREMENT'} include 'bp', 'hr', 'pressure', 'blood', 'rr', 'rate', 'heart'; \textcolor{blue}{high frequency of 'LOCATION'} include \\ 'hospital', 'right', 'home', 'floor', 'emergency', 'micu'; \textcolor{blue}{high frequency of 'EVENT'} include 'ct', 'placed', 'cxr', 'intubated', 'exam', 'review'; \textcolor{blue}{high frequency of 'FREQUENCY'} \\ include 'chronic', 'intermittent', 'daily', 'occasionally', 'frequent', 'intermittently'; \textcolor{blue}{high frequency of 'ORGANIZATION'} include 'ems', 'service', 'surgery', 'pcp', \\ 'emergency', 'neuro', 'medicine'; \textcolor{blue}{high frequency of 'DATE'} include '2171114', '21491117'; high frequency of 'AGE' include '60', '80yo', '78', '61', 'seventyeightyearold', \\ '69'; \textcolor{blue}{high frequency of 'GENDER'} include 'man', 'woman', 'f', 'male', 'female', 'm'.\end{tabular} \\ \hline
\textbf{UMLS knowledge} &
  \begin{tabular}[c]{@{}l@{}}\textcolor{blue}{The Unified Medical Language System (UMLS) }is developed by the U.S. National Library of Medicine (NLM) to integrate and standardize diverse medical \\ terminologies and coding systems. It consists of three main components: the Metathesaurus, Semantic Network, and SPECIALIST Lexicon, supporting medical \\ information retrieval and semantic analysis. \textcolor{blue}{You understand medical terminology and concepts from UMLS.}\end{tabular} \\ \hline
\textbf{Error analysis} &
  \begin{tabular}[c]{@{}l@{}}\textcolor{blue}{The prediction errors stem from several factors.} Entity boundary recognition issues were common, particularly with multi-token entities like "shortness of breath" \\ or "paroxysmal nocturnal dyspnea," where some tokens were missed or incorrectly labeled as 'O.' Additionally, the model struggled with entity type confusion, \\ such as distinguishing "pain" as a symptom versus its contextual use related to location. Context-dependent misinterpretations also contributed to errors, especially \\ in handling negations like "denies chest pain" or temporal references such as "last few months." Overlapping entities posed further challenges, where closely related \\ terms (e.g., "MI" and "CABG") interfered with accurate classification. Finally, rare or unseen entities in the training data led to occasional misclassifications, \\ highlighting gaps in the model's ability to generalize.\end{tabular} \\ \hline
\end{tabular}%
}
\caption{Specific static prompts for each component we used for the MIMIC III dataset.}
\label{tab:mimic_prompt}
\end{table}
\begin{table}[h]
\centering
\renewcommand{\arraystretch}{1.2}
\resizebox{\columnwidth}{!}{%
\begin{tabular}{ll}
\hline
\rowcolor[HTML]{FCEBEB} 
\textbf{Prompt Strategies} &
  \multicolumn{1}{c}{\cellcolor[HTML]{D9EAD3}\textit{\textbf{NCBI}}} \\ \hline
 &
  \begin{tabular}[c]{@{}l@{}}\textbf{[Task Description]:} You are a medical AI trained to identify and classify tokens into \textcolor{blue}{five categories: DiseaseClass, SpecificDisease, Modifier, CompositeMention} \\ \textcolor{blue}{and Outside ('O').} Your task is to extract and classify the DiseaseClass, SpecificDisease, Modifier and CompositeMention from this dataset.\end{tabular} \\ \cline{2-2} 
 &
  \begin{tabular}[c]{@{}l@{}}\textbf{[Entity Types with Definitions]:} \textcolor{blue}{'DiseaseClass' refers to} a classification system or category used to group various medical conditions or diseases based on certain \\ characteristics, such as their nature, affected biological systems, or underlying causes. \textcolor{blue}{'SpecificDisease' appears to} describe particular diseases that are identified and \\ classified based on their specific clinical features, genetic origins, or biochemical abnormalities. \textcolor{blue}{'Modifier' refers to} specific attributes or variations or conditions that \\ can modify or influence the presentation, progression, or characteristics of a disease, alter the manifestation or course of a disease, potentially affecting its diagnosis, \\ treatment, and prognosis. \textcolor{blue}{'CompositeMention' describes} medical conditions or characteristics that are composed of several elements or features, often involving \\ multiple tissues, organs, or systems. Any token not falling into these categories should be labeled as 'O'.\end{tabular} \\ \cline{2-2} 
\multirow{-3}{*}{\textbf{Basic Prompt}} &
  \begin{tabular}[c]{@{}l@{}}\textbf{[Format Specification]:} For example, \textcolor{blue}{the sentence} 'Histidinemia. Classical and atypical form in siblings.' \textcolor{blue}{is tokenized and labeled} as follows: {[}'Histidinemia.', \\ 'Classical', 'and', 'atypical', 'form', 'in', 'siblings.'{]} with labels {[}'SpecificDisease', 'O', 'O', 'O', 'O', 'O', 'O'{]}. Your task is to predict and return the label for each provided \\ token, ensuring the number of output labels matches the number of input tokens exactly. \textcolor{blue}{The output format should  include tokens with their labels:} {[}'Histidinemia.-\\ SpecificDisease', 'Classical-O', 'and-O', 'atypical-O', 'form-O', 'in-O', 'siblings.-O'{]}.\end{tabular} \\ \hline
\textbf{Description of datasets} &
  \begin{tabular}[c]{@{}l@{}}The data you are working with is \textcolor{blue}{NCBI disease corpus}, a collection of 793 PubMed abstracts fully annotated at the mention and concept level to serve as a research \\ resource for the biomedical natural language processing community. Each PubMed abstract was manually annotated by two annotators with disease mentions and \\ their corresponding concepts in Medical Subject Headings (MeSH) or Online Mendelian Inheritance in Man (OMIM). The public release of the NCBI disease corpus \\ contains 6892 disease mentions, which are mapped to 790 unique disease concepts. Of these, 88 percent link to a MeSH identifier, while the rest contain an OMIM \\ identifier. We were able to link 91 percent of the mentions to a single disease concept, while the rest are described as a combination of concepts.\end{tabular} \\ \hline
\textbf{High-frequency instances} &
  \begin{tabular}[c]{@{}l@{}}In this dataset, \textcolor{blue}{high-frequency 'DiseaseClass' include} 'disorder', 'abnormalities', 'tumors', 'mental', 'disorders', 'retardation'. \textcolor{blue}{High-frequency 'SpecificDisease' include} \\ 'deficiency', 'syndrome', 'dystrophy', 'familial', 'myotonic', 'colorectal'. \textcolor{blue}{High-frequency 'Modifier' include} 'tumor', 'tumour', 'APC', 'choroideremia', 'DM', 'DMD'. \\ \textcolor{blue}{High-frequency 'CompositeMention' include} 'breast', 'ovarian', 'cancer', 'muscular', 'andor', 'becker'.\end{tabular} \\ \hline
\textbf{UMLS knowledge} &
  \begin{tabular}[c]{@{}l@{}}\textcolor{blue}{The Unified Medical Language System (UMLS) }is developed by the U.S. National Library of Medicine (NLM) to integrate and standardize diverse medical \\ terminologies and coding systems. It consists of three main components: the Metathesaurus, Semantic Network, and SPECIALIST Lexicon, supporting medical \\ information retrieval and semantic analysis. \textcolor{blue}{You understand medical terminology and concepts from UMLS.}\end{tabular} \\ \hline
\textbf{Error analysis} &
  \begin{tabular}[c]{@{}l@{}}\textcolor{blue}{The prediction errors in the NCBI dataset primarily stem} from challenges in distinguishing composite mentions and modifiers within complex biomedical contexts. \\ For instance, entities like "BRCA1 gene" were incorrectly segmented, with "BRCA1" labeled as a modifier instead of being part of the composite mention. \\ Additionally, multi-token composite mentions such as "breast and ovarian cancer" were not consistently labeled, with individual tokens occasionally missed or \\ misclassified. Contextual ambiguity, such as distinguishing between mentions of general biological terms (e.g., "tumor") and their specific functional roles (e.g., \\ "tumor suppressor"), also contributed to errors.\end{tabular} \\ \hline
\end{tabular}%
}
\caption{Specific static prompts for each component we used for the NCBI dataset.}
\label{tab:ncbi_prompt}
\end{table}
\begin{table}[h]
\centering
\renewcommand{\arraystretch}{1.2}
\resizebox{\columnwidth}{!}{%
\begin{tabular}{ll}
\hline
\rowcolor[HTML]{FCEBEB} 
\textbf{Prompt Strategies} &
  \multicolumn{1}{c}{\cellcolor[HTML]{D9EAD3}\textit{\textbf{Med-Mentions}}} \\ \hline
 &
  \begin{tabular}[c]{@{}l@{}}\textbf{[Task Description]:} You are a medical AI trained to identify and classify tokens into \textcolor{blue}{two categories: Disease and Outside ('O').} Your task is to extract and classify \\ the Disease related concepts from this dataset.\end{tabular} \\ \cline{2-2} 
 &
  \begin{tabular}[c]{@{}l@{}}\textbf{[Entity Types with Definitions]:} \textcolor{blue}{'Disease' is a particular abnormal condition that} adversely affects the structure or function of all or part of an organism and is not \\ immediately due to any external injury. Diseases are often known to be medical conditions that are associated with specific signs and symptoms. A disease may be \\ caused by external factors such as pathogens or by internal dysfunctions. For example, internal dysfunctions of the immune system can produce a variety of \\ different diseases, including various forms of immunodeficiency, hypersensitivity, allergies, and autoimmune disorders. Any token not falling into Disease categories \\ should be labeled as 'O'.\end{tabular} \\ \cline{2-2} 
\multirow{-3}{*}{\textbf{Basic Prompt}} &
  \begin{tabular}[c]{@{}l@{}}\textbf{[Format Specification]:} For example, \textcolor{blue}{the sentence} 'A total of 200 children and adolescents with type 1 diabetes, ages 9-18 years, completed the DEPS-R Turkish \\ version.' \textcolor{blue}{is tokenized and labeled} as follows: {[}'A', 'total', 'of', '200', 'children', 'and', 'adolescents', 'with', 'type', '1', 'diabetes,', 'ages', '9-18', 'years,', 'completed', 'the', \\ 'DEPS-R', 'Turkish', 'version.'{]} with labels {[}'O', 'O', 'O', 'O', 'Disease', 'O', 'Disease', 'O', 'Disease', 'Disease', 'Disease', 'Disease', 'O', 'Disease', 'O', 'O', 'Disease', \\ 'Disease', 'Disease'{]}. \textcolor{blue}{The output format should  include tokens with their labels:} {[}'A-O', 'total-O', 'of-O', '200-O', 'children-Disease', 'and-O', 'adolescents-Disease', \\ 'with-O', 'type-Disease', '1-Disease', 'diabetes,-Disease', 'ages-Disease', '9-18-O', 'years,-Disease', 'completed-O', 'the-O', 'DEPS-R-Disease', 'Turkish-Disease', \\ 'version.-Disease'{]}.\end{tabular} \\ \hline
\textbf{Description of datasets} &
  \begin{tabular}[c]{@{}l@{}}The data you are working with is \textcolor{blue}{Med-Mentions}, a new manually annotated resource for the recognition of biomedical concepts. What distinguishes Med-Mentions \\ from other annotated biomedical corpora is its size (over 4,000 abstracts and over 350,000 linked mentions), as well as the size of the concept ontology (over 3 \\ million concepts from UMLS 2017) and its broad coverage of biomedical disciplines.\end{tabular} \\ \hline
\textbf{High-frequency instances} &
  In this dataset, \textcolor{blue}{high frequency 'Disease' related entities include} 'patients', 'cells', 'treatment', 'cancer', 'analysis', 'disease', 'clinical'. \\ \hline
\textbf{UMLS knowledge} &
  \begin{tabular}[c]{@{}l@{}}\textcolor{blue}{The Unified Medical Language System (UMLS) }is developed by the U.S. National Library of Medicine (NLM) to integrate and standardize diverse medical \\ terminologies and coding systems. It consists of three main components: the Metathesaurus, Semantic Network, and SPECIALIST Lexicon, supporting medical \\ information retrieval and semantic analysis. \textcolor{blue}{You understand medical terminology and concepts from UMLS.}\end{tabular} \\ \hline
\textbf{Error analysis} &
  \begin{tabular}[c]{@{}l@{}}\textcolor{blue}{The prediction errors in the Med-Mentions dataset are primarily due to} challenges in identifying complex and overlapping disease mentions, as well as distinguishing \\ between general biomedical terms and specific disease entities. Multi-token entities such as "renal pedicle occlusion" or "intention-to-treat analyses" were often \\ partially labeled, with some tokens being misclassified or excluded. Additionally, the presence of nested or overlapping mentions, such as "prostate cancer" and its \\ relationship to broader contexts like "treatment disparities," led to inconsistent labeling. The model also struggled with domain-specific terminology, misclassifying \\ general terms like "maternal genotype" or "outcome" as disease mentions. These errors highlight limitations in handling nuanced biomedical language, especially \\ when entities span multiple tokens or overlap with related terms.\end{tabular} \\ \hline
\end{tabular}%
}
\caption{Specific static prompts for each component we used for the Med-Mentions dataset.}
\label{tab:Med-Mentions_prompt}
\end{table}

\end{appendices}

\end{document}